\documentclass[%
 aip,
 amsmath,amssymb,
 reprint,%
]{revtex4-1}
\usepackage{multirow}
\usepackage{graphicx}% Include figure files
\usepackage{dcolumn}% Align table columns on decimal point
\usepackage{bm}% bold math
\usepackage[utf8]{inputenc}
\usepackage[T1]{fontenc}
\usepackage{etoolbox}
\usepackage{bbm}
\usepackage{soul, color, xcolor}
\usepackage{pifont}

\usepackage{amsmath}
\usepackage{siunitx}
\usepackage{mathptmx}
\usepackage{mathrsfs}
\usepackage{hyperref}
\usepackage{pifont}
\usepackage{booktabs}
\usepackage{multirow}
\usepackage{threeparttable}
\usepackage{diagbox}
\usepackage{tabularx}
\usepackage{mathrsfs}
\usepackage{makecell}
\usepackage[caption=false,font=footnotesize,labelfont=rm,textfont=rm]{subfig}
\definecolor{highlightcolor}{rgb}{0,0,1}
\sethlcolor{highlightcolor}
\DeclareMathAlphabet{\mathcal}{OMS}{cmsy}{m}{n}
\makeatletter
\def\@email#1#2{%
 \endgroup
 \patchcmd{\titleblock@produce}
  {\frontmatter@RRAPformat}
  {\frontmatter@RRAPformat{\produce@RRAP{*#1\href{mailto:#2}{#2}}}\frontmatter@RRAPformat}
  {}{}
}%
\makeatother
\begin{document}

\preprint{AIP/123-QED}

\title{SAMSGL: Series-Aligned Multi-Scale Graph Learning for Spatio-Temporal Forecasting}
\author{Xiaobei Zou}
    \email{Electronic mail: xbeizou@gmail.com}
\author{Luolin Xiong} 
    \email{Electronic mail: xiongluolin@gmail.com}
\author{Yang Tang}
    \email{Electronic mail: yangtang@ecust.edu.cn}
    \homepage{Corresponding author: Yang Tang.}
\affiliation{The Key Laboratory of Smart Manufacturing in Energy Chemical Process, East China University of Science and Technology, Shanghai 200237, China.}
\author{Jürgen Kurths}
    \email{Electronic mail: Juergen.Kurths@pik-potsdam.de}
    \altaffiliation[Also at ]{Institute of Physics, Humboldt University of Berlin, 12489 Berlin, Germany.}
\affiliation{Potsdam Institute for Climate Impact Research, 14473 Potsdam, Germany.}

% Force line breaks with \\

\date{\today}% It is always \today, today,

\begin{abstract}
Spatio-temporal forecasting in various domains, like traffic prediction and weather forecasting, is a challenging endeavor, primarily due to the difficulties in modeling propagation dynamics and capturing high-dimensional interactions among nodes. Despite the significant strides made by graph-based networks in spatio-temporal forecasting, there remain two pivotal factors closely related to forecasting performance that need further consideration: time delays in propagation dynamics and multi-scale high-dimensional interactions. In this work, we present a Series-Aligned Multi-Scale Graph Learning (SAMSGL) framework, aiming to enhance forecasting performance. In order to handle time delays in spatial interactions, we propose a series-aligned graph convolution layer to facilitate the aggregation of non-delayed graph signals, thereby mitigating the influence of time delays for the improvement in accuracy. To understand global and local spatio-temporal interactions, we develop a spatio-temporal architecture via multi-scale graph learning, which encompasses two essential components: multi-scale graph structure learning and graph-fully connected (Graph-FC) blocks. The multi-scale graph structure learning includes a global graph structure to learn both delayed and non-delayed node embeddings, as well as a local one to learn node variations influenced by neighboring factors. The Graph-FC blocks synergistically fuse spatial and temporal information to boost prediction accuracy. To evaluate the performance of SAMSGL, we conduct experiments on meteorological and traffic forecasting datasets, which demonstrate its effectiveness and superiority. 
\end{abstract}

\maketitle

\begin{quotation}
Accurate prediction of spatio-temporal systems is crucial for understanding system conditions and making effective decisions. This paper introduces the Series-Aligned Multi-Scale Graph Learning (SAMSGL) framework, aimed at capturing propagation dynamics and modeling high-dimensional interactions. Firstly, we propose a series-aligned graph convolution layer to address time delays between nodes, thereby enhancing spatial message aggregation. Additionally, to extract spatio-temporal interactions from data, we present a comprehensive spatio-temporal architecture incorporating a multi-scale graph structure learning module and Graph-fully connected (Graph-FC) blocks. Within the multi-scale graph structure learning module, adaptive node embeddings and node locations are utilized to derive global spatial delayed and non-delayed relations, as well as local information. The Graph-FC blocks effectively fuse spatial information from multiple graphs and temporal information. Experimental results demonstrate that our model improves prediction accuracy compared to previous approaches. We validate the effectiveness of our modules through ablation studies. \\
\end{quotation}

\section{Introduction}
Spatio-temporal forecasting plays a pivotal role in predicting forthcoming states within spatio-temporal networks across diverse domains like chaotic prediction \cite{chaotic1,chaotic2, chaotic3}, traffic prediction \cite{traffic1}, weather forecasting \cite{c:2,cheng2023generative} and power grid prediction \cite{power1, xiong2022two}. Despite these achievements, the performance of spatio-temporal forecasting hinges on two crucial factors: \textbf{1) time delays in propagation dynamics:} it is time-consuming to exchange messages between spatial nodes and the propagation speed during spatial interactions acts as a determinant affecting inter-node time delays. It is contingent not only upon spatial distance, but also intricacies like conditions and directions of propagation paths among nodes. For instance, in traffic forecasting, holidays impact road conditions, influencing traffic speed \cite{macioszek2021road} and leading to increased time delays between nodes. Similarly, shifts in global weather patterns, such as monsoons in meteorological forecasting \cite{boers2019complex}, alter propagation direction and speed, affecting time delays. \textbf{2) high-dimensional interactions:} the complexity of high-dimensional networks characterized by a large number of nodes and the complex connections between them \cite{tang2020introduction} presents a formidable challenge in capturing inter-node interactions, thereby intensifying the intricacy of spatio-temporal forecasting \cite{ji2023signal}. In particular, traffic flow forecasting employs thousands of sensors \cite{chen2001freeway}, and high-resolution weather forecasting involves tens of thousands of grids \cite{c:5}. This intricate interplay extends beyond proximate neighbors, incorporating associations between distant and adjoining nodes, thus complicating the capture of spatio-temporal interactions among nodes. Integrating time delays in propagation dynamics and capturing high-dimensional interactions tend to enhance prediction accuracy.\par
While approaches like Graph Convolution Networks (GCN) \cite{kipf2016semi} endeavor to simulate node propagation dynamics by aggregating spatial information from neighboring nodes, time delays among nodes introduce a challenge in accurately capturing propagation dynamics. However, GCN-based methods aggregate messages from nodes within the same time step, without considering the influence of node features from previous time steps that genuinely impact the present features of nodes. To address the influence of interaction time delays and better learn propagation dynamics, some series alignment methods have emerged. For instance, Corrformer \cite{c:2} aligns node series prior to computing cross-correlation using a tree structure. Although Corrformer embeds node locations, it does not fully account for spatial interaction factors, such as distance, propagation conditions, and propagation path direction, which affect time delays. Consequently, there is a pressing need to develop an approach that effectively aggregates messages from nodes' spatially relevant features, while reducing the influence of time delays. %adequately considers both spatial interactions and time delays to improve the modeling of propagation dynamics.

Graph structure learning methods have been developed to depict the high-dimensional intricate interactions between nodes. The structure of a graph is learned based on nodes' similarity \cite{li2021spatial} or generated with trainable node embeddings. For example, Regularized Graph Structure Learning (RGSL) \cite{yu2022regularized} captures node interactions from implicit and prior graphs. Given the complexity of high-dimensional interactions, the representation of implicit interactions with a single graph may be insufficient to successfully address this issue. Therefore, employing multi-scale graphs that portray interactions at both local explicit and global implicit scales holds significant promise to tackle these intricate high-dimensional interactions \cite{c:14}.

Furthermore, to capture spatio-temporal interactions among nodes, aside from the construction of multi-scale graph structures, it is essential to employ efficient spatio-temporal architectures. While there have been notable advancements in this regard, such as the development of Graph Convolution Recurrent Neural Networks (GCRNN) \cite{c:9,c:13} and the integration of graph convolution with attention mechanisms \cite{c:15,li2022dmgan}, there remains potential for developing a spatio-temporal architecture that can fuse spatial and temporal information more effectively and achieve better forecasting performance.

To tackle these challenges, we present the Series-Aligned Multi-Scale Graph Learning (SAMSGL) framework. This approach incorporates a series-aligned module within graph convolution networks to mitigate the impact of time delays during the propagation process. Additionally, we develop a spatio-temporal architecture via multi-scale graph learning, aiming to capture high-dimensional spatio-temporal interactions from both local and global scales. This architecture encompasses two key components: multi-scale graph structure learning and Graph-Fully Connected (Graph-FC) blocks. In summary, our contributions can be outlined as follows:
\begin{itemize}
\item A series-aligned graph convolution layer is designed, which effectively accounts for the impact of time delays in propagation dynamics. This additional layer aligns the time series of nodes by reweighing node features based on time delays.
\item A spatio-temporal architecture, known as multi-scale graph learning, is developed to comprehensively capture intricate high-dimensional spatio-temporal interactions. This architecture comprises multi-scale graph structure learning to understand global and local spatial interactions, as well as Graph-FC blocks to fuse spatial and temporal information more effectively.
\item Numerous experiments on real-world datasets for meteorology forecasting and traffic flow forecasting tasks are conducted to demonstrate the superior performance compared to existing models.
\end{itemize}
The rest of the paper is organized as follows: Section \ref{sec:section2} provides an overview of previous researches in graph convolution, graph structure learning, and spatio-temporal forecasting. In Section \ref{sec:section3}, we present the specific details of our proposed series-aligned graph convolution layer, multi-scale graph structure learning modules, and the graph-FC block. Section \ref{sec:section4} presents the experimental details and results. Section \ref{sec:section5} makes a conclusion of our work.
\section{Related Work}\label{sec:section2}
\subsection{Graph Convolution.}
Graph convolution which aims to aggregate features and propagate messages from nodes, can be categorized into two types: spectral\cite{bruna2013spectral,defferrard2016convolutional,c:7,guo2022orthogonal} and spatial graph convolution\cite{jiang2021gpens,zhang2021learning,bose2023can,sawhney2021exploring,graph}. Spectral graph convolution can be conceptualized as filtering the graph signals, where the filtering operation of graph can be realized as a self-loop Laplacian matrix \cite{bruna2013spectral} or approximated by Chebyshev polynomials \cite{defferrard2016convolutional}. Efforts to enhance graph representation include refining the graph transformation process for graph signals. AKGNN\cite{c:7} is introduced as a data-driven graph kernel learning module, dynamically adjusting the trade-off between all-pass and low-pass filters. To address instability in feature transformation, Guo \textit{et al.}\cite{guo2022orthogonal} propose Ortho-GConv, ensuring robust orthogonality. 
% Additionally, \cite{lim2022sign} introduces a network enforcing invariance in the sign and basis of eigenvectors, thereby enhancing the expression of spectral graph convolution.

Spatial graph convolution aims to replicate the message propagation process \cite{jiang2021gpens} and aggregates information from neighboring nodes \cite{zhang2021learning}.
Bose \textit{et al.}\cite{bose2023can} propose an aggregation strategy that combines features from lower to higher-order neighborhoods in a non-recursive way by employing a randomized path exploration approach. For tasks unsuitable for Euclidean space, some methods map the graph to alternative spaces. Sawhney \textit{et al.} \cite{sawhney2021exploring} utilize Riemannian manifolds to represent spatial correlations between stocks. These methods primarily focus on information aggregation at a single time step, with less emphasis on considering time delays in message propagation.
Our series-aligned graph convolution also aims to achieve better aggregation of node features by performing convolutions on aligned time series, thereby reducing the influence of time delays.
\subsection{Graph Structure Learning.}
In spatio-temporal forecasting, graph structure learning is employed to establish suitable relationships between nodes, thereby enhancing spatio-temporal representation. Within graph structure learning, some methods directly create a graph and employ it in downstream tasks, while others concentrate on refining existing structures derived from similarities or physical connections. Some approaches directly generate an adjacency matrix from node sequences with node similarity. For instance, STFGNN \cite{li2021spatial} employs dynamic time warping to measure node sequence similarity. 
% ST-LGSL \cite{tang2022spatio} utilizes latent graph topological information based on MLP-KNN.
To tackle graph structure learning of large scale graph or long sequences, some researches use neural networks to learn a graph. Other approaches involve generating graphs from node embeddings. For instance, RGSL \cite{yu2022regularized} captures implicit graph information by a regularized graph generation module. Additionally, the Meta-Graph Convolutional Recurrent Network (MGCRN) \cite{c:13} uses hyper-network and node embedding banks to disentangle graphs in both space and time. 
% The Possibilistic Neighborhood Graph (PNG) \cite{gao2022possibilistic} adaptively measures the possibility of two nodes being neighbors. 

To refine the original graph structure, AdaSTN \cite{ta2022adaptive} employs temporal convolution to assess sequence similarity, using the learned graph to fine-tune the original structure. Chen \textit{et al.}\cite{chen2020iterative} optimize the graph iteratively by refining the structure with graphs generated from similarity metrics. Establishing better relations between nodes, graph structure learning has demonstrated its effectiveness in enhancing graph representation. 
Therefore, in this paper, we focus on constructing multi-scale graph structure learning that generates graphs from global delayed and non-delayed scales, as well as local scales to capture remote and neighboring node interactions.
\subsection{Spatio-temporal Architecture.}
% Spatio-temporal forecasting is often treated as a multivariate time series forecasting task. 
Spatio-temporal forecasting tasks can be approached with or without considering spatial information. When treating all node states as variables and disregarding location information, multivariate forecasting techniques \cite{zhang2023ctfnet, earlywarning} are employed to to predict future dynamics. These methods encompass models like the autoregressive integrated moving average model \cite{kothapalli2017real}, Recurrent Neural Networks (RNN) \cite{shao2016traffic}, and transformer-based structures \cite{c:2}. They tend to focus on capturing the seasonal and trend information within time series \cite{rebei2023fsnet, ensemble_air} or decompose time series into various frequency components \cite{wu2022timesnet}. Some approaches emphasize temporal evolution and variable characteristics, aiming to capture relations between variables \cite{zhang2022crossformer} or accommodate variable-specific shifts \cite{cirstea2022triformer}. 
In contrast, efforts have also been made to incorporate location information into multivariate methods. Techniques like embedding spatial positional formation into features have been explored. For instance, by spectral transformation of the graph structure, \cite{jiang2023pdformer} embeds the eigenvalues into features to make the network aware of the structure. Additionally, graph-based methods are utilized for representing spatial relationships. 

Furthermore, graph-based methods \cite{power2, power_graph, zhao2019t, li2018diffusion, c:9} employ adjacency matrices to illustrate the connections between nodes, encompassing factors such as road connectivity and distance between nodes. Some approaches combine graphs with RNN by incorporating graph convolution features into both input and hidden states. For instance, TGCN \cite{zhao2019t} combines combined GCN and the gated recurrent unit to capture the spatio-temporal dependency. Different type of graph convolution layers are applied for better spatial representation. DCRNN \cite{li2018diffusion} applied a diffusion convolution layer model the traffic diffusion process, while AGCRN \cite{c:9} proposed a node and data adaptive graph convolution to achieve the node specific graph convolution. 
Others benefit from alternating graph networks and temporal modules, as well as stacked multi-layer spatio-temporal blocks. ASTGCN \cite{c:15} use a spatial temporal attention to predict traffic flow by alternately learning the spatial and temporal attention modules. Approaches that consider both spatial and temporal relationships typically yield improved performance in spatio-temporal forecasting tasks, emphasizing the significance of effectively combining spatial and temporal interactions. 
In this paper, we propose a spatio-temporal architecture that combines multi-scale graph structure learning and spatio-temporal fusion Graph-FC blocks to learn the complex spatio-temporal dependencies.
\section{Methodologies}\label{sec:section3}
\begin{figure*}[!t]
\centering
\includegraphics[width=\textwidth]{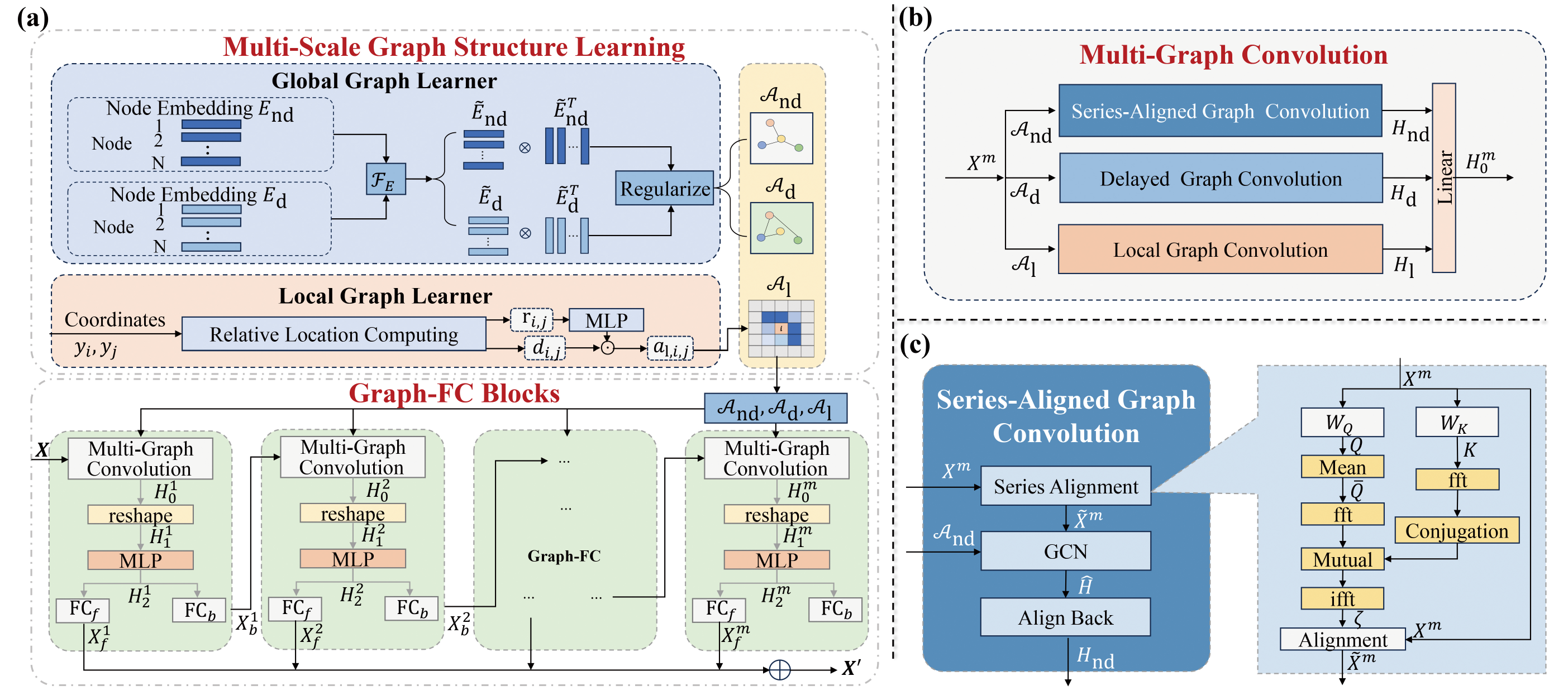} % Reduce the figure size so that it is slightly narrower than the column. Don't use precise values for figure width.This setup will avoid overfull boxes.
\caption{The framework of Series-Aligned Multi-Scale Graph Learning (SAMSGL). (a) The overall structure of SAMSGL. The Multi-scale graph structure learning module is developed to generate spatial relations of different scales, which are denoted as $\mathcal{A}_{\rm nd}$, $\mathcal{A}_{\rm d}$, and $\mathcal{A}_{\rm l}$, respectively. These generated adjacency matrices, along with the node states $X$, are input into stacks of Graph-FC blocks. (b) The structure of Multi-Graph Convolution in the Graph-FC blocks. (c) The workflow of Series-Aligned Graph Convolution where "fft" represents fast Fourier transform and "ifft" is inversed fast Fourier transform.}
\label{fig1}
\end{figure*}

In this section, we present the framework of the proposed SAMSGL, as illustrated in Fig. \ref{fig1}. We begin by defining the problem and introducing the notations in this study. Subsequently, we describe the key components of our model,which includes: the series-aligned Graph Convolution which captures the time delay node features, the multi-scale graph structure learning module that obtains global delayed and non-delayed features and local features, and Graph-FC blocks that combine the spatial and temporal information.
The overall workflow of our SAMSGL is depicted in Fig. \ref{fig1}(a), encompassing the multi-scale graph structure learning and the Graph-FC blocks. The specifics of multi-graph convolution in Graph-FC blocks are outlined in Fig. \ref{fig1}(b), while the series-aligned graph convolution workflow is presented in Fig. \ref{fig1}(c).

% Firstly, the series-aligned graph convolution (depicted in Figure \ref{fig1} c) is designed to capture time-delayed node features. Next, the multi-scale graph structure learning (shown in upper part of Figure \ref{fig1} a) is responsible for acquiring global delayed and non-delayed graphs, along with the local graph. Lastly, we learn the multi-scale graph information together with spatio-temporal relations with Graph-FC (located in the lower part of Figure \ref{fig1} (a)) and the details of the multi-graph convolution in Graph-FC (shown in Figure \ref{fig1} b).
  %   Q, K &= W^{q}X, W^{k}X\\
  %   % s &= \mathcal{F}_{c}(\bar{Q},K,V)\\
  %   %   &= 
  %   % \tau &= L- argmax(s_{i}),i\in{\{1, 2, \ldots, L\}}
  %   % s &= \mathcal{F}_{c}(Q, K) \\
  %   %     &=\mathscr{F}(\mathscr{F}\bar{\mathscr{F}})
  %   %     &=\int_{-\infty}^{+\infty}Q_{t}e^{-2\pi it\omega}dt\bar{\int_{-\infty}^{+\infty}K_{t}e^{-2\pi it\omega}}dt
  %   % Q, K &= W^{q}X, W^{k}X\\
  %   % s &= \mathcal{F}_{c}(Q, K)\\
  %   % a&= \mathscr{F}(\mathscr{F}\mathscr{F}*)\\
  % % b&= \int_{-\infty}^{+\infty} Q_{t} e^{-2\pi i t \omega}\mathrm{d}t \int_{-\infty}^{+\infty}\bar{ K_{t} e^{-2\pi i t \omega}\mathrm{d}t}

\subsection{Preliminaries}
Spatio-temporal forecasting involves predicting future dynamics of spatio-temporal systems. In graph-based models, the interactions between $N$ nodes are represented by the adjacency matrix $\mathcal{A}\in \mathbb{R}^{{N}\times{N}}$, where the elements of the adjacency matrix indicate interaction coefficients among nodes. The state of node $i$ at time step $t$ is denoted as $X_{i,t}\in \mathbb{R}^{C}$, and $C$ is the state's dimensions for each node. For spatio-temporal forecasting tasks, the input comprises historical time series with length $L$ of $N$ nodes, which can be represented as $X \in \mathbb{R}^{{N}\times{L}\times{C}}$. The objective is to train a network $\mathcal{F}_{\theta}$ with parameters $\theta$ to predict future series $X^{'}\in \mathbb{R}^{{N}\times{L^{'}}\times{C}}$ with length $L^{'}$ as follows:
\begin{equation} X^{'} =\mathcal{F}_{\theta}(X,\mathcal{A}).\end{equation}
%do we need to add an object function

\subsection{Series-Aligned Graph Convolution}
Due to the existence of time delays in propagation dynamics, the states of target nodes may be truly influenced by the states of its neighboring nodes a few time steps ago. However, traditional graph convolution aggregates state messages of neighboring nodes at the same time step, without considering time delays in propagation. To illustrate, due to time delays in propagation, the state $X_{i,t}$ of node $i$ at time step $t$ may be influenced not only by its immediate previous neighboring node's state $X_{i-1, t-1}$, but also by the state of its neighboring node $X_{j, t-\tau_{i,j}}$, where $\tau_{i,j}$ represents the delay between node $i$ and node $j$. Therefore, it is necessary to consider time delays by series alignment \cite{c:2} before graph convolution. Our series-aligned graph convolution aims to enhance the aggregation of related node features by conducting series alignment, thereby mitigating the impact of intricate time delays between nodes. The procedure, as shown in Fig. \ref{fig1} (c), involves a sequence of steps: series alignment, followed by graph convolution, and finally, an aligning back operation.

We firstly implement a series alignment module to ascertain time delays between nodes. Recognizing that calculating the delay between each node pair is computationally demanding, we set a reference node $\bar{q}$ and calculate the delay of each node relative to the reference node. By aligning all the nodes with the reference node, we achieve computational efficiency while still obtaining relevant delay information. %we adopt more efficient mean series to calculate the delay for each node in relation to the mean time series across all nodes. 
To accomplish this, inspired by Wu \textit{et al.}\cite{c:2}, we identify the time step with the highest correlation as the initial point of the time series of each node and align the series. The process unfolds as follows:
\begin{subequations}
\begin{align}
    Q&= X{W}_{Q}, \label{eq:2-a} \\
    K&=X{W}_{K}, \label{eq:2-a}
\end{align}
\end{subequations}
where we utilize parameter matrices ${W}_{Q}$ and ${W}_{K}$ to project $X$ to query $Q\in \mathbb{R}^{N\times{L}\times{D}}$ and key $K\in \mathbb{R}^{N\times{L}\times{D}}$. $D$ denotes the number of channels of node features after feature embedding. Then we define a reference node $\bar{q}$ with series $\bar{Q}\in \mathbb{R}^{L\times{D}}$, which is the average series of $N$ nodes. $\bar{Q} = \frac{1}{N}\sum_{i=1}^{N}Q_{i}$, and 
%\begin{align}
%    \label{eq_ave} \bar{Q} &= \frac{1}{N}\sum_{i=1}^{N}Q_{i},
%\end{align}
$Q_{i}\in \mathbb{R}^{L\times{D}}$ is the projected query time series of node $i$. We calculate correlation scores $\zeta$ between the global average series $\bar{Q}$ and all nodes, facilitated through the correlation function $\mathcal{R}$.
\begin{align}
    \label{eq1}    \zeta &= \mathcal{R}(\bar{Q}, K) = \mathscr{F}^{-1}(\mathscr{F}(\bar{Q})\mathscr{F}^*(K)),
\end{align}
where $\mathscr{F}(\bar{Q})$ entails a fast Fourier transform applied to $\bar{Q}$. $\mathscr{F}^*(K)$ corresponds to the conjugation of the fast Fourier transform of $K$, and $\mathscr{F}^{-1}(\cdot)$ refers to the inversed fast Fourier transform. After deriving correlation scores at each time step between nodes in the graph and the reference node, we identify the time step with the highest correlation score as the time of messages arrival, since nodes often exhibit similar patterns within message propagation through a graph. Consequently, we determine the time delay between the reference node and each node $i$ as follows:
\begin{equation}
    \begin{aligned}
      \label{eq2} \tau_{i,\bar{q}} = L- \mathop{\rm argmax}\limits_{t}(\zeta_{i,t}), \quad t\in{\{1, 2, \ldots, L\}}, \\ 
    \end{aligned}
\end{equation}
where $\zeta_{i,t}$ represents the correlation score of node $i$ to the average series at time step $t$. $\tau_{i,\bar{q}}$ represents the time delay between node $i$ and the reference node $\bar{q}$.
% We presume that the time series of nodes with length $L$ share a common pattern implicit in the dynamics of each node, which can be perceived as a result of propagation dynamics. So the time step with highest correlation represents the .
% Therefore, we obtain the time delay among reference
% $\tau$ captures the delay associated with each feature of every node. 
The alignment of the time series of node $i$ is carried out as follows:
\begin{eqnarray}
    \Tilde{X}_{i,t}=\left\{
    \begin{aligned}
        &X_{i,(t+\tau_{i,\bar{q}})}, t+\tau_{i,\bar{q}}\leq L, \\
        &X_{i,(t+\tau_{i,\bar{q}}-L)}, t+\tau_{i,\bar{q}}\textgreater L,
    \end{aligned}
    \right.
\end{eqnarray}
where $\Tilde{X}_{i,t}$ corresponds to the time series after alignment of node $i$ at time step $t$. Then the aligned time series $\Tilde{X}$ is reweighed with correlation scores $\zeta$, and a $1$-dimensional convolution layer denoted as ``Conv1d" is utilized. Subsequently, we proceed to execute the graph convolution operation on the adjacency matrix $\mathcal{A}_{\rm nd}$ of the non-delayed graph.
\begin{alignat}{2}
    % \mathcal{F}_{\rm r}(\Tilde{X}) &= {\rm Conv1d}(\zeta\Tilde{X}), \\
    \hat{H} &= \mathcal{A}_{\rm nd}{\rm Conv1d}(\zeta\Tilde{X}),
\end{alignat}
where $\hat{H}$ denotes the output of the graph convolution. Preparing to fuse with other graph convolution output, we align back the time series of each node and obtain the final output of series aligned graph convolution $H_{\rm nd}$.
\begin{eqnarray}
    H_{{\rm nd},i,t}=\left\{
    \begin{aligned}
        &\hat{H}_{i,(t+\tau_{i,\bar{q}})}, t\geq \tau_{i,\bar{q}}, \\
        &\hat{H}_{i,(t+\tau_{i, \bar{q}}-L)}, t\textless \tau_{i,\bar{q}}.
    \end{aligned}
    \right.
\end{eqnarray}
\subsection{Multi-Scale Graph Structure Learning}
The high-dimensional interactions between nodes can be characterized by connections among a large number of nodes across various spatial scales. 
To learn spatial interactions, we adopt multi-scale graph structure learning for different spatial scales.
We construct graphs in a multi-scale sense, including the global delayed and non-delayed graphs, as well as the local graph. These two global graphs depict global interactions from two pertinent perspectives based on the overall node topology, which are tailored for spatial graph aggregation and aligned node features. Inspired by Jiang \textit{et al.}\cite{c:13} which employ a shared network to ensure the universality of graph structures, we generate the global delayed and non-delayed graphs employing separate trainable node embeddings and a shared network $\mathcal{F}_{E}$. These embeddings are generated randomly to ensure diverse graph perspectives. The procedure unfolds as follows:
% \begin{align}
%      \Tilde{E}_{nd} &= NN_{E}(E_{nd})\\
%      \Tilde{E}_{d} &= NN_{E}(E_{d})\\
%      \mathcal{G}_{nd0},\mathcal{G}_{d0} &= \Tilde{E}_{nd}\Tilde{E}_{nd}^T,\Tilde{E}_{d}\Tilde{E}_{d}^T
% \end{align}
\begin{subequations}
    \begin{align}
     \Tilde{E_c} &= \mathcal{F}_{E}(E_c),\\
     \hat{\mathcal{A}_c} &= \Tilde{E}_{c}\Tilde{E}_{c}^T,
\end{align}
\end{subequations}
where $c$ denotes the delayed or non-delayed classes, and $c \in {\{\rm d, \rm nd\}}$. $E_c$ is the trainable node embedding. We derive the generalized node embedding $\Tilde{E}_{c}$ by feeding $E_{c}$ into the shared network $\mathcal{F}_{E}$, and $\Tilde{E}_{c}^T$ is the transpose of $\Tilde{E}_{c}$. $\hat{\mathcal{A}}_{c}$ is the raw adjacency matrix formed using node embeddings. To prevent over-fitting and maintain graph sparsity, we implement a regularization operation \cite{yu2022regularized} on the raw adjacency matrix.
\begin{equation}
\begin{aligned}
    a_{c,i,j} = &\sigma(\log(\hat{a}_{c,i,j}/(1-\hat{a}_{c,i,j}))+(g_{i,j}-g_{i,j}^{'})\slash{s}), \\
    &\text{s.t.} \quad g_{i,j}, g_{i,j}^{'} \sim \text{Gumbel}(0,1),
\end{aligned}
\end{equation}
where $\sigma$ is the activation function, and $\hat{a}_{c,i,j}$ is an element of $\hat{\mathcal{A}}_{c}$. $g_{i,j}$ and $g_{i,j}^{'}$ 
are drawn from a Gumbel(0,1) distribution, of which the probability density function is  
$f(z;\mu, \beta) = \frac{1}{\beta} \exp\left(-\frac{z - \mu}{\beta} - \exp(-(z - \mu)/\beta)\right)$.  
% $f(z;\mu,\beta) = {1}/{\beta} \exp\left(-\left({(z - \mu)}/{\beta} + e^{-(z - \mu)/\beta}\right)\right)$,
$z$ represents the random variable, $\mu=0$ is the location parameter and $\beta=1$ is the scale parameter. 
% are sampled from a Gumbel(0,1) distribution.
$s$ is a hyper-parameter related to temperature. %In Gumbel distribution, we set the location parameter to $0$, and set the scale parameter to 1. 
$a_{c,i,j}$ is an element of the regularized adjacency matrix $\mathcal{A}_c$. By implementing regularization to $\hat{\mathcal{A}}_{\rm nd}$ and $\hat{\mathcal{A}}_{\rm d}$, we derive the adjacency matrices $\mathcal{A}_{\rm nd}$ and $\mathcal{A}_{\rm d}$ of global non-delayed and delayed graphs.

Subsequently, we proceed with the local graph generation based on the distance between nodes. This module is designed to adaptively learn a graph based on the spatial positions of nodes. We fine-tune the local graph, initially constructed using distance or road connections, through a Multi-Layer Perceptron (MLP) network. When node coordinates, such as longitudes and latitudes, are available, we input the relative coordinates between neighboring nodes into the MLP. In the scenarios, where only distances between nodes are provided, we can also input the distance into the MLP. The process is as follows:
\begin{equation}
    {a}_{{\rm l},i,j} = {\rm exp}(-\frac{d_{i,j}^2}{\alpha})\textsc{MLP}(r_{i,j}),
\end{equation}
where ${a}_{{\rm l},i,j}$ represents an element of the adjacency matrix $\mathcal{A}_{\rm l}$ of local graph. $d_{i,j}$ corresponds to the distance between node $i$ and node $j$. $\alpha$ is a hyper-parameter to scale the distance. 
The value $r_{i,j}$ is derived from the relative location computing module, which captures the relationship between node $i$ and node $j$. This relationship may encompass relative coordinates such as $y_i$ and $y_j$, or the distance $d_{i,j}$ between the nodes, depending on the available spatial information. 

It is worth noting that we remap locations distributed across a spherical surface and incorporate the angles between nodes \cite{c:6} to calculate the adjacency matrix for large-scale global spatio-temporal forecasting tasks. 
% The format of this process is as follows:
% \begin{equation}
%     {a}_{{\rm l},i,j} = \frac{\phi_{i,j}}{2\pi}{\rm exp}(-\frac{d_{i,j}^2}{\mu})\textsc{MLP}(y_{i},y_{j}^{'}),
% \end{equation}
% where $\phi_{i,j}$ represents the angle between node $i$ and node $j$. $y_{i}$ denotes the coordinates of node $i$, while $y_{j}^{'}$ is the relative coordinates of node $j$. 
Through the process of global and local graph structure learning, we acquire $\mathcal{A}_{\rm nd}$ of the global non-delayed graph, $\mathcal{A}_{\rm d}$ of the global delayed graph and $\mathcal{A}_{\rm l}$ of the local graph.

\subsection{Graph-FC Blocks}
To capture spatio-temporal dependencies and fuse node features from multi-scale graphs, we build Graph-FC blocks. These blocks combine multi-graph convolution layers with backward-forward residual structures to iteratively learn spatio-temporal dynamics. As shown in Fig. \ref{fig1} (b), each Graph-FC block utilizes multi-graph convolution to capture features from multi-scale graphs. The outputs of graph convolution, $H_{\rm nd}$, $H_{\rm d}$, and $H_{\rm l}$, are then concatenated and fed into a fully-connected layer. The multi-graph convolution, which captures node features from global delayed, non-delayed, and local scales, can be expressed as follows: 
% \begin{align}
%     &H^{nd} &=& \Gamma^{-1}(\mathcal{G}_{nd}\mathcal{F}_{r}(\Gamma(X^(k))) \\
%     &H^{d} &=& \mathcal{G}_{d}X^(k)W_{d} \\
%     &H^{l} &=& \mathcal{G}_{l}X^(k)W_{l} \\
%     &H^{(k)}_{0} &=& FC([H^{nd},H^{d},H^{l}]) 
% \end{align}
\begin{subequations}
    \begin{alignat}{2}
    % &H^{\rm nd} &=& \Gamma^{-1}(\mathcal{A}^{\rm nd}\mathcal{F}_{r}(\Gamma(X^{(k)}))), \\
    &H_{\rm d} &=& \mathcal{A}_{\rm d}X^{m}W_{\rm d}, \\
    &H_{\rm l} &=& \mathcal{A}_{\rm l}X^{m}W_{\rm l}, \\
    &H^{m}_{0} &=& {\rm FC}([H_{\rm nd},H_{\rm d},H_{\rm l}]),
\end{alignat}
\end{subequations}
\begin{table*}[htpb]
\caption{Baseline methods.}
\centering
% \small
\begin{tabular}{|c|c|c|}%{c|c|c}
\hline \hline
\multicolumn{1}{|c|}{Classes} & Baseline Methods  & Graph structure learning                                                                                                              \\ \midrule
\multirow{8}{*}{RNN-based models} 
& Graph Convolutional Gate Recurrent Unit \textbf{(GCGRU)} \cite{seo2018structured}  &\ding{55} \\ 
& Diffusion Convolutional Recurrent Neural Network \textbf{(DCRNN)} \cite{li2018diffusion} & \ding{55} \\ 
& Temporal Graph Convolutional Network \textbf{(TGCN)} \cite{zhao2019t}    &\ding{55}  \\  
& Adaptive Graph Convolutional Recurrent Network \textbf{(AGCRN)} \cite{c:9}   & \ding{51}             \\  
& Conditional Local Convolution Recurrent Network \textbf{(CLCRN)} \cite{c:6}  &  \ding{55}          \\ 
& Scalable Graph Predictor \textbf{(SGP)} \cite{cini2023scalable}   & \ding{55} \\ 
& Meta-Graph Convolutional Recurrent Network (\textbf{MGCRN}) \cite{c:13}   & \ding{51}    \\ 
& Regularized Graph Structure Learning (\textbf{RGSL}) \cite{yu2022regularized}    & \ding{51}     \\ 
\midrule \multirow{3}{*}{\makecell{Time-space alternation}} 
& Spatio-Temporal Graph Convolutional Network (\textbf{STGCN}) \cite{yu2017spatio}   &\ding{55} \\ 
& Multi-Component Spatial-Temporal Graph Convolution Networks (\textbf{MSTGCN}) \cite{c:15}   & \ding{55}   \\ 
& Attention based Spatio-Temporal Graph Convolutional Networks (\textbf{ASTGCN}) \cite{c:15} & \ding{55}    \\  \midrule
Transformer-based model &\textbf{Corrformer} \cite{c:2} &\ding{55} \\ \hline \hline 
\end{tabular}
\label{tab0}
\end{table*}
where $X^{m}$ represents the input for the $m$-th block. $W_{\rm d}$ and $W_{\rm l}$ are the trainable parameters corresponding to delayed graph convolution and local graph convolution. The subsequent step involves reshaping the hidden state of the $k$-th block $H^{m}_{0}\in \mathbb{R}^{{L}\times{N}\times{D}}$ into $H^{m}_{1}\in \mathbb{R}^{{N}\times{(L\times D)}}$ and applying the backward-forward residual operation. Following this, as shown in Fig. \ref{fig1} (a), we pass the hidden state through a MLP to capture temporal features $H_2^{m}$. Finally, two fully-connected layers ${\rm FC}_b$ and ${\rm FC}_f$ are employed to map $H^{m}_{2}$ to historical and forecasting lengths.
\begin{subequations}
    \begin{align}
    H^{m}_{2} &= \text{MLP}(H^{m}_{1}), \\
    X_{b}^{m} &= {\rm FC}_b(H^{m}_{2}), \\
    X_{f}^{m}&={\rm FC}_f(H^{m}_{2}), \\
    X^{'} &= \sum_{m=1}^{M}X_{f}^{m},
\end{align}
\end{subequations}
where $X_{b}^{m} \in \mathbb{R}^{{N}\times{L\times D}}$ and $X_{f}^{m} \in \mathbb{R}^{{N}\times{T\times D}}$ correspond to the backward and forward vectors, respectively. $M$ is the total number of Graph-FC blocks. The input of the first Graph-FC block is $X^{0}=X $, and the input of $(m+1)$-th Graph-FC is $X^{m}=X_{b}^{m-1}$. The final output is denoted as $X^{'}$.
% The overall workflow is as follows. Firstly, in the multi-scale graph structure learner, the non-delayed global graph $\mathcal{A}_{\rm nd}$ and delayed global graph $\mathcal{A}_{\rm d}$ are generated by global graph learner. The local graph $\mathcal{A}_{\rm l}$ is generate by local graph learner with node location information. Then, the time series of nodes together with the learned graphs are taken into the Graph-FC blocks to capture the spatial and temporal interactions.
\section{Experiments}\label{sec:section4}
In this section, we perform experiments on meteorology and traffic datasets and compare our model with spatio-temporal forecasting methods based on artificial intelligence (AI). Additionally, we conduct ablation studies to assess the effectiveness of each component.
\begin{table*}[htpb]
\caption{Overall forecasting performance of different methods on different meteorology and traffic datasets.}
\centering
\begin{tabular}{@{\hskip0.25cm}c|ccccccc@{}}
\hline \hline
\diagbox[width=3.7cm, height=0.85cm]{Methods \& Metrics}{Datasets}   &      & Wind          & Temperature   & Cloud cover   & Humidity      & PeMSD4         & PeMSD8         \\ \midrule
\multirow{2}{*}{TGCN \cite{zhao2019t} }       & MAE  & 4.1747$\pm$0.0324           & 3.8638$\pm$0.0970         & 2.3934$\pm$0.0216         & 1.4700$\pm$0.0295         & 34.7859$\pm$0.2043          & 38.2618$\pm$1.0967         \\
          & RMSE & 5.6730$\pm$0.0412          & 5.8554$\pm$0.1432          &3.6512$\pm$0.0223           & 2.1066$\pm$0.0551          &52.4670$\pm$0.2708  & 57.0713$\pm$1.5649
         \\ \midrule
\multirow{2}{*}{STGCN \cite{yu2017spatio}}      & MAE  & 3.6477$\pm$0.0000          & 4.3525$\pm$1.0442         & 2.019$\pm$0.0392         & 0.7975$\pm$0.2378       &25.3017$\pm$2.3161  &21.3431$\pm$0.1006
          \\
           & RMSE & 4.8146$\pm$0.0003          & 6.8600$\pm$1.1233   & 2.9542$\pm$0.0542          & 1.1109$\pm$0.2913        &38.9132$\pm$2.2506
        &32.5995$\pm$0.2007          \\ \midrule
\multirow{2}{*}{MSTGCN \cite{c:15} }     & MAE  & 1.9440$\pm$0.0150   & 1.2199$\pm$0.0058         & 1.8732$\pm$0.0010          & 0.6093$\pm$0.0012         &21.9152$\pm$0.6523    &18.5269$\pm$0.2870        \\
          & RMSE & 2.9111$\pm$0.0292           & 1.9203$\pm$0.0093           & 2.8629$\pm$0.0073         & 0.8684$\pm$0.0019       &34.6353$\pm$0.9230
&28.5750$\pm$0.4316    \\ \midrule
\multirow{2}{*}{ASTGCN \cite{c:15} }     & MAE  & 2.0889$\pm$0.0006         & 1.4896$\pm$0.0130     & 1.9936$\pm$0.0002         & 0.7288$\pm$0.0229
         &23.5648$\pm$0.9421     & 18.4513$\pm$0.5240     \\
          & RMSE & 3.1356$\pm$0.0012         & 2.4622$\pm$0.0023           & 2.9576$\pm$0.0007           & 1.0471$\pm$0.0402     &36.7932$\pm$1.3277
         &28.4153$\pm$0.8141    \\ \midrule
\multirow{2}{*}{GCGRU \cite{seo2018structured} }      & MAE  & 1.4116$\pm$0.0057         & 1.3256$\pm$0.1499          & 1.5938$\pm$0.0021        & 0.5007$\pm$0.0002           &25.8336$\pm$0.0399   &20.2753$\pm$0.0572         \\
          & RMSE & 2.2931$\pm$0.0047         & 2.1721$\pm$0.1945         & 2.5576$\pm$0.0116            & 0.7891$\pm$0.0006     &39.8696$\pm$0.1136   &31.8069$\pm$0.0812
 \\ \midrule
\multirow{2}{*}{DCRNN \cite{li2018diffusion} }      & MAE  &1.4321$\pm$0.0019	&1.3232$\pm$0.0864	&1.5938$\pm$0.0021	&0.5046$\pm$0.0011
          & 24.9117$\pm$2.0638         & 20.2303$\pm$0.1051       \\
          & RMSE &2.3364$\pm$0.0055	&2.1874$\pm$0.1227	&2.5412$\pm$0.0044	&0.7956$\pm$0.0033          &38.619$\pm$2.8952 
         & 31.7793$\pm$0.0376    \\ \midrule
\multirow{2}{*}{AGCRN \cite{c:9}  }      & MAE  & 2.4194$\pm$0.1149	&1.2651$\pm$0.0080	&1.7501$\pm$0.1467	&0.5759$\pm$0.1632 &19.3291$\pm$0.3053
&15.8135$\pm$0.0957 \\
         & RMSE & 3.4171$\pm$0.0055	&1.9314$\pm$0.0219	&2.7585$\pm$0.1694	&0.8549$\pm$0.2025  &31.3227$\pm$0.5927 &25.5016$\pm$0.1877
        \\ \midrule
\multirow{2}{*}{CLCRN \cite{c:6}  }      & MAE  & 1.3260$\pm$0.0483	&1.1688$\pm$0.0457	&1.4906$\pm$0.0037	&0.4531$\pm$0.0065 & 127.5582$\pm$0.1709 &120.7720$\pm$0.0037        \\
         & RMSE & 2.129$\pm$0.0733	&1.8825$\pm$0.1509	&2.4559$\pm$0.0037	&0.7078$\pm$0.0146   &159.9446$\pm$0.7968 &146.6032$\pm$0.0159
    \\ \midrule
\multirow{2}{*}{Corrformer \cite{c:2} } & MAE  &1.3157$\pm$0.0007	&0.8055$\pm$0.0014 &0.1559$\pm$0.0001	  &4.6720$\pm$0.0023         &25.4462$\pm$0.1916	&20.2820$\pm$0.1950	         \\
          & RMSE &2.1507$\pm$0.0007		&1.3627$\pm$0.0011		&0.2473$\pm$0.0005		&7.1811$\pm$0.0041
         &37.7988$\pm$0.2390		&30.1798$\pm$0.2621          \\ \midrule
\multirow{2}{*}{MGCRN \cite{c:13} }      & MAE  & 1.3899$\pm$0.0035          & 1.1143$\pm$0.0709          & 1.5930$\pm$0.0035          & 0.4481$\pm$0.0012          & \textbf{18.8580$\pm$0.0413}	&\underline{15.2828$\pm$0.0410}	 \\
          & RMSE & 2.2696$\pm$0.0099          & 1.8806$\pm$0.1187          & 2.5409$\pm$0.0154          & 0.6903$\pm$0.0043        &\underline{30.9136$\pm$0.1235}		&\underline{24.6436$\pm$0.2891}
 \\ \midrule
\multirow{2}{*}{RGSL \cite{yu2022regularized} }       & MAE  & \underline{1.1927$\pm$0.0096}    & \underline{0.7877$\pm$0.0353}    & \underline{1.4911$\pm$0.0015}    & \underline {0.4165$\pm$0.0072}    & 19.5440$\pm$0.2571		&15.7333$\pm$0.2037	     \\
          & RMSE & \underline{1.9342$\pm$0.0161}    & \underline{1.3233$\pm$0.0397}    & \underline{2.4136$\pm$0.0052}    & \underline{0.6385$\pm$0.0097}    &31.0884$\pm$0.1087		&25.0912$\pm$0.2831
 \\ \midrule
\multirow{2}{*}{Ours}       & MAE  &\textbf{0.9014$\pm$0.0091}		&\textbf{0.6329$\pm$0.0131}			&\textbf{1.4086$\pm$0.0037}			&\textbf{0.3761$\pm$0.0018}	&\underline{19.1346$\pm$0.0627}	&\textbf{15.2174$\pm$0.0035}
 \\
           & RMSE	&\textbf{1.4072$\pm$0.0157}		&\textbf{1.0876$\pm$0.0194}		&\textbf{2.3084$\pm$0.0084}			&\textbf{0.5807$\pm$0.0028}		&\textbf{30.6541$\pm$0.0776}		&\textbf{24.4680$\pm$0.0015}
 \\ \hline \hline
\end{tabular}
\begin{tablenotes} 
    \item Results with \textbf{bold} are the overall best performance, and results with \underline{underlines} are the suboptimal performance.
\end{tablenotes}
\label{tab1}
\end{table*}
\subsection{Datasets}
\subsubsection{WeatherBench} WeatherBench \cite{rasp2020weatherbench} is derived from the ERA5 reanalysis dataset, which has been regrided to lower resolutions. We select the hourly dataset from Jan. 1, 2010 to Jan. 1, 2019 with a resolution of 5.625$^{\circ}$ (32$\times$64 grid points), encompassing a total of 2,048 nodes \cite{c:6}. It is worth noting that this dataset is a general dataset in spatio-temporal forecasting tasks, which is rougher in spatial resolution compared with data usually taken for weather forecasting. A more extended application for weather forecasting is under preparation for a forthcoming paper. Our evaluation of the model's performance in weather forecasting involves four datasets aimed at predicting wind speed, global temperature, cloud cover, and humidity. The respective units for these predictions are $\text{ms}^{-1}$, $\rm K$, $\%\times$10$^{-1}$, and $\%\times$10. In the scope of our experiments, we focus on short-term weather forecasting. This entails inputting 12 hours of historical time series data and predicting weather conditions for the subsequent 12 hours. 
\subsubsection{PeMSD4 and PeMSD8} The PeMSD4 and PeMSD8 datasets are derived from PeMS system \cite{chen2001freeway} and sampled at 5-minute intervals by Guo \textit{et al.} \cite{c:15}. PeMSD4 consists of traffic flow records from 3,848 detectors located on 29 roads within San Francisco Bay Area, ranging from Jan. 1, 2018 to Feb. 28, 2018. PeMSD8 comprises traffic records from 1,979 detectors situated across 8 roads in San Bernardino, covering the period from Jul. 1, 2016 to Aug. 31, 2016. Our experiments on these two datasets center on forecasting traffic conditions one hour ahead using the most recent hour of historical data. Both the input and output sequences have a length of 12 data points.
\subsubsection{METR-LA} The dataset comprises traffic records gathered from 207 sensors situated along the Los Angeles County highway \cite{c:13}. The data spans from Mar. 1, 2012 to Jun. 27, 2012, with a sampling interval of 5 minutes. Our prediction task involves forecasting traffic conditions for a one-hour horizon using a historical series spanning one hour. Both input and output sequences have a length of 12 data points.
% The PeMSD4 dataset is derived from PeMS system \cite{chen2001freeway} and processed by \cite{c:15}, which consists of traffic flow records from $3,848$ detectors located on $29$ roads within the San Francisco Bay Area. The data spans from 1 January 2018 to 28 February 2018. Our experiments on this dataset center on forecasting traffic conditions one hour ahead using the most recent hour of historical data. The time series data is sampled at $5$-minute intervals. Therefore, both the input and output sequences have a length of $12$ data points.
% \subsubsection{PeMSD8} The PeMSD8 dataset is another dataset from PeMS system \cite{chen2001freeway} and sampled at $5$-minute intervals by \cite{c:15}, which comprises traffic records in San Bernardino, covering the period from 1 July 2016 to 31 August 2016. This dataset encompasses data from $1,979$ detectors situated across $8$ roads. The length of input and output sequences are the same as the settings of PeMSD4.
% Our forecasting efforts here involve predicting traffic conditions one hour in advance using input and output sequences, both of which have a length of $12$ data points.

\subsection{Implement Details}
The experiments are carried out on a single NVIDIA RTX 8000 48GB GPU. 
The graph network architecture is established using the torch package in Python, along with the torch-geometric package. The training phase of the model encompass 100 epochs, with a batch size set to 32.
For weather forecasting tasks, we employ a data split ratio of 7:1:2 for the training, validation, and test sets, respectively \cite{c:6}. For the traffic flow prediction experiments, the datasets are split in a ratio of 6:2:2 \cite{ji2023signal}. Normalization is performed on the training, testing, and validation sets by using the mean and variance of the training sets before inputting the time series into the model.

The dimension of node features after feature embedding is set to be 64, and the spatial convolution channels are also set to be 64,  consistent with RGSL \cite{yu2022regularized}. There are 4 Graph-FC blocks in the model. Each block consists of a single multi-graph convolution layer, which contains a series aligned graph convolution, a delayed graph convolution and a local graph convolution. In each series-aligned graph convolution, the 1-dimensional convolution layer employs a kernel size of 3. Subsequently, the multi-graph convolution is succeeded by 2 fully-connected layers, each yielding an output channel size of 256 \cite{oreshkin2019n}.
Throughout the experiments, the Mean Absolute Error (MAE) is employed as the loss function, and the model is trained using the Adam optimizer. The learning schedule follows \cite{c:6}, which is a multi-step decay approach. The learning rate is initially set as 0.005, with a learning rate decay ratio of 0.05. The minimum learning rate is constrained to $1e^{-8}$. To evaluate the performance of our proposed SAMSGL, we use the MAE and the Root Mean Square Error (RMSE) metrics.

% \begin{table}[!h]
% \small
% \centering
% \begin{tabular}{@{}c|ccc@{}}
% \toprule
% \diagbox{Methods}{MAE}{Horizon}  & 3    & 6    & 12    \\ \midrule
% STGCN \cite{yu2017spatio}      & 2.88   & 3.47   & 4.59   \\
% DCRNN \cite{li2018diffusion}     & 2.77   & 3.15   & 3.60   \\
% AGCRN \cite{c:9}                  & 2.86   & 3.25   & 3.68   \\
% MGCRN \cite{c:13}                   & \textbf{2.52}   & \textbf{2.93}   & 3.38   \\
% SGP \cite{cini2023scalable}          & 2.69   & 3.05   & 3.45   \\
% Ours                   & 2.97       & 2.99       & \textbf{3.06}       \\ \bottomrule
% \end{tabular}
% \caption{Forecasting performance at different horizons on METR-LA dataset.}
% \label{tab-ml}
% \end{table}
\begin{figure*}[htbp]
\centering
\subfloat[Ground Truth]{\includegraphics[width=0.27\textwidth]{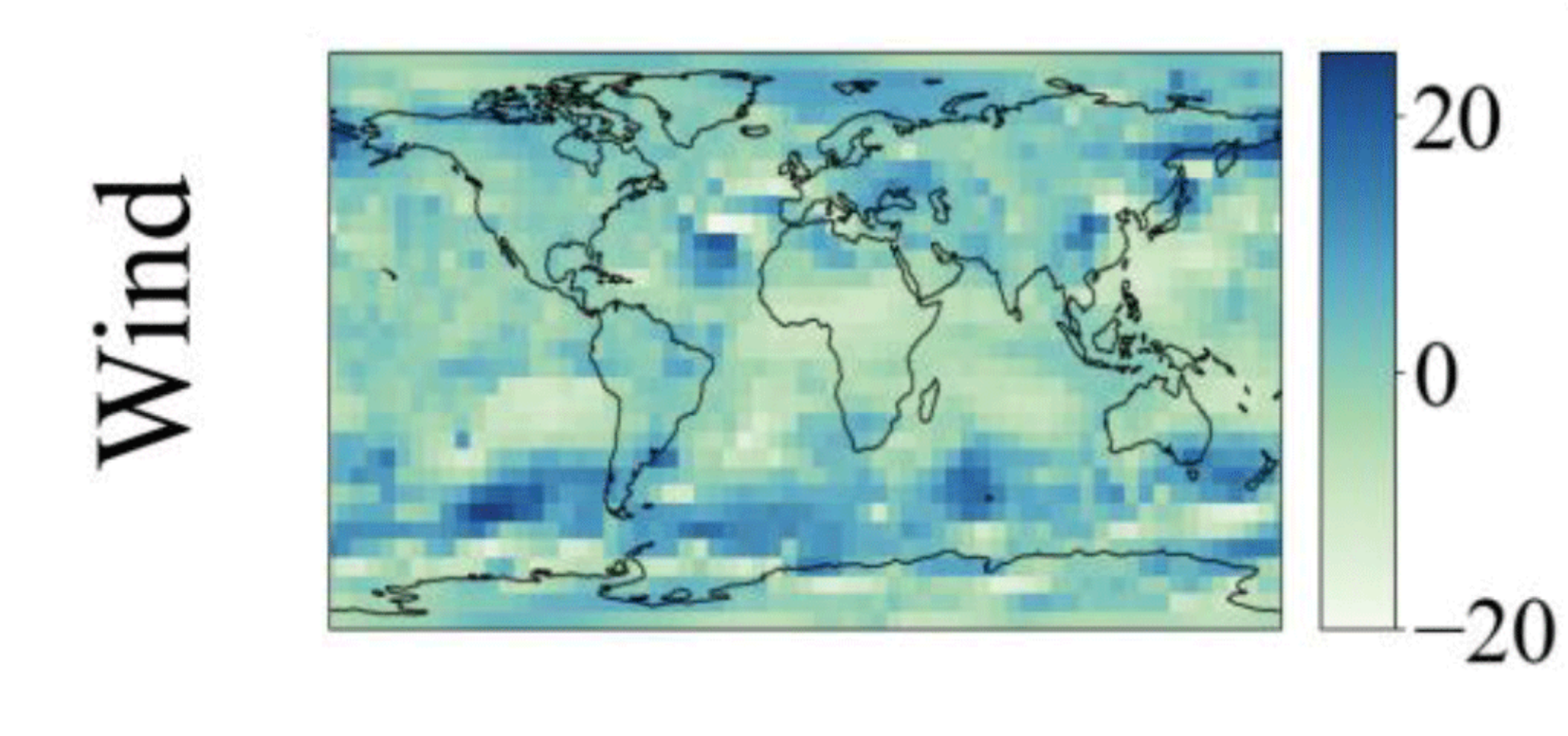}%
\label{e_a}}
\hfil
\subfloat[RGSL]{\includegraphics[width=0.23\textwidth]{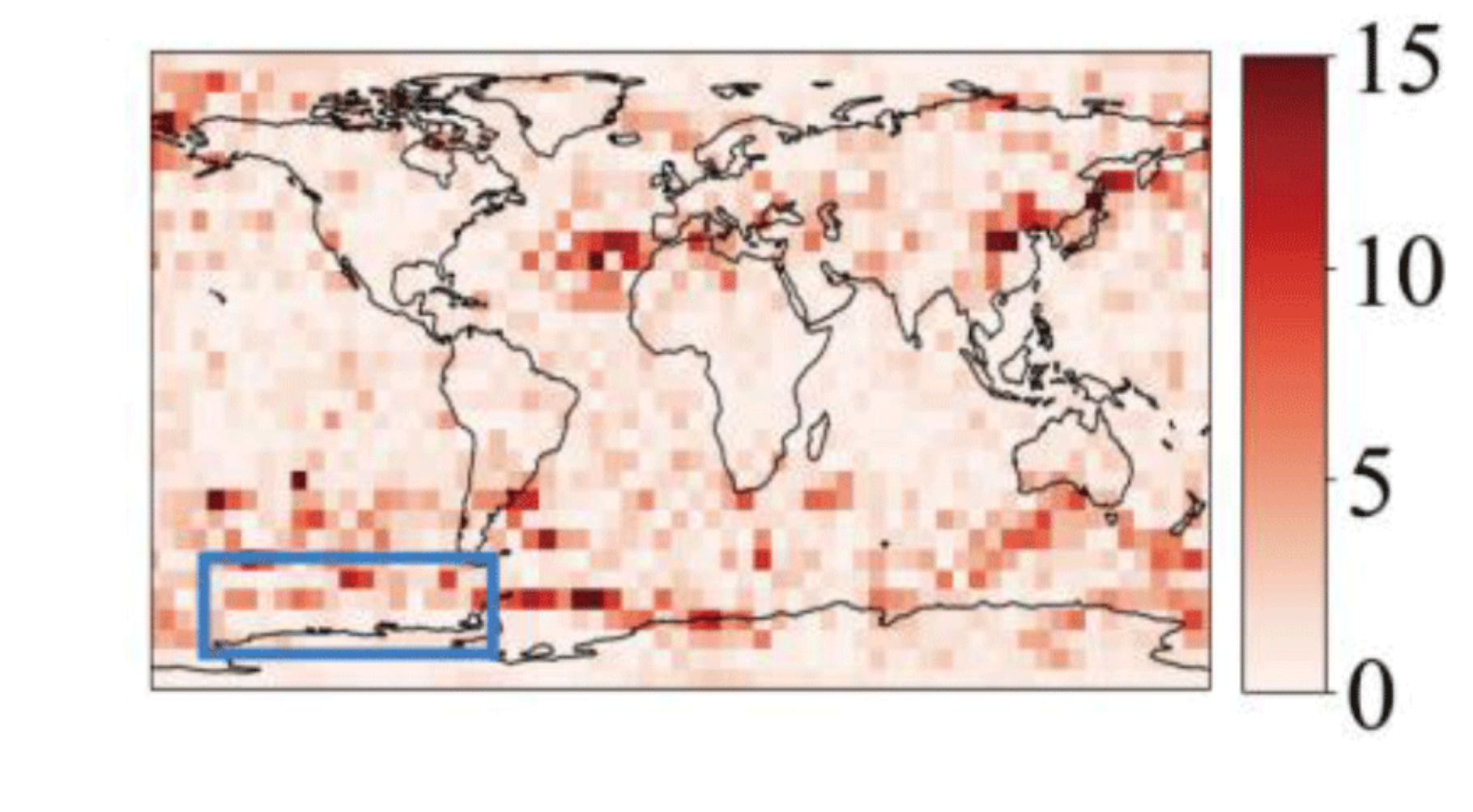}%
\label{e_b}}
\hfil
\subfloat[Corrfomer]{\includegraphics[width=0.23\textwidth]{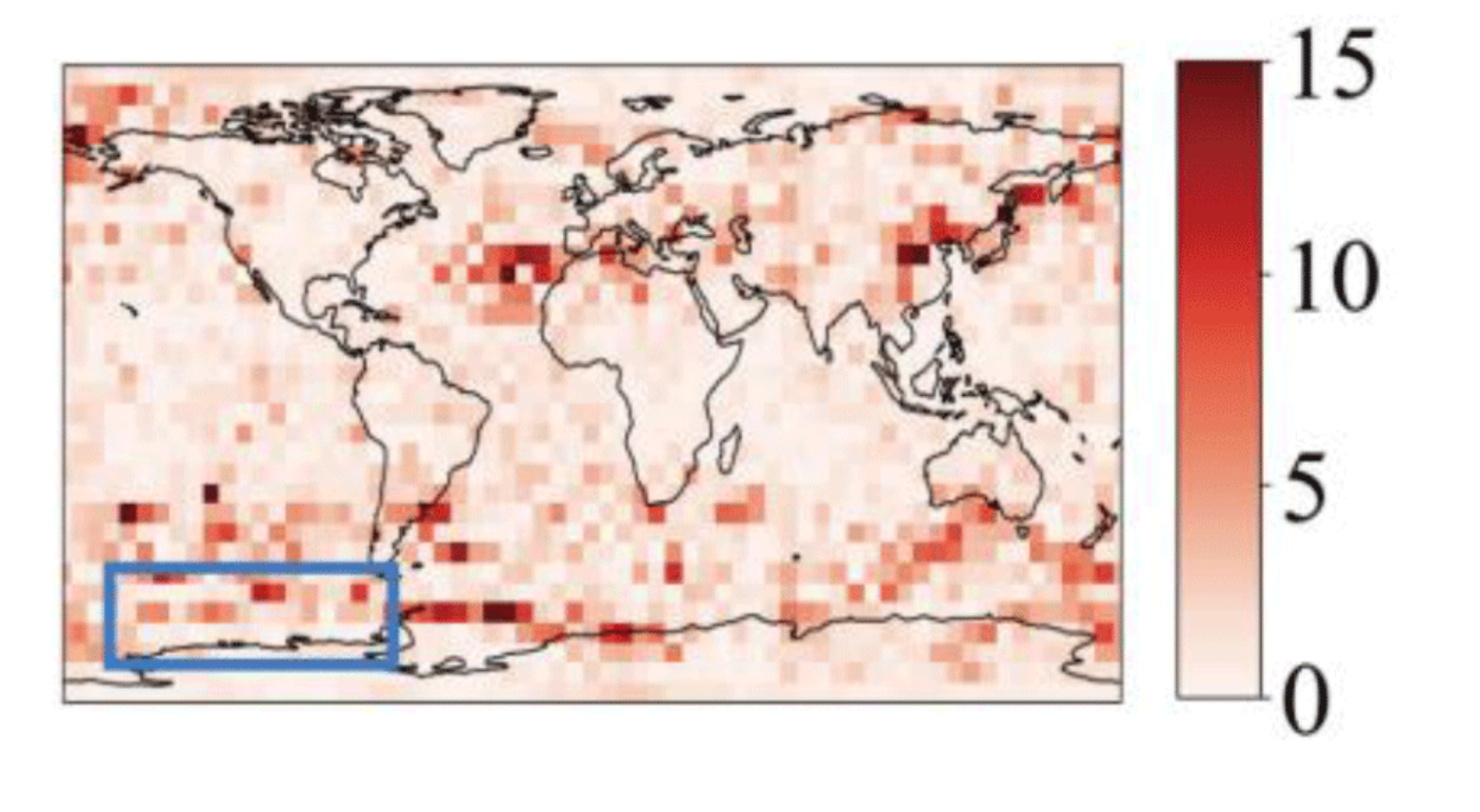}%
\label{e_c}}
\hfil
\subfloat[SAMSGL (Ours)]{\includegraphics[width=0.23\textwidth]{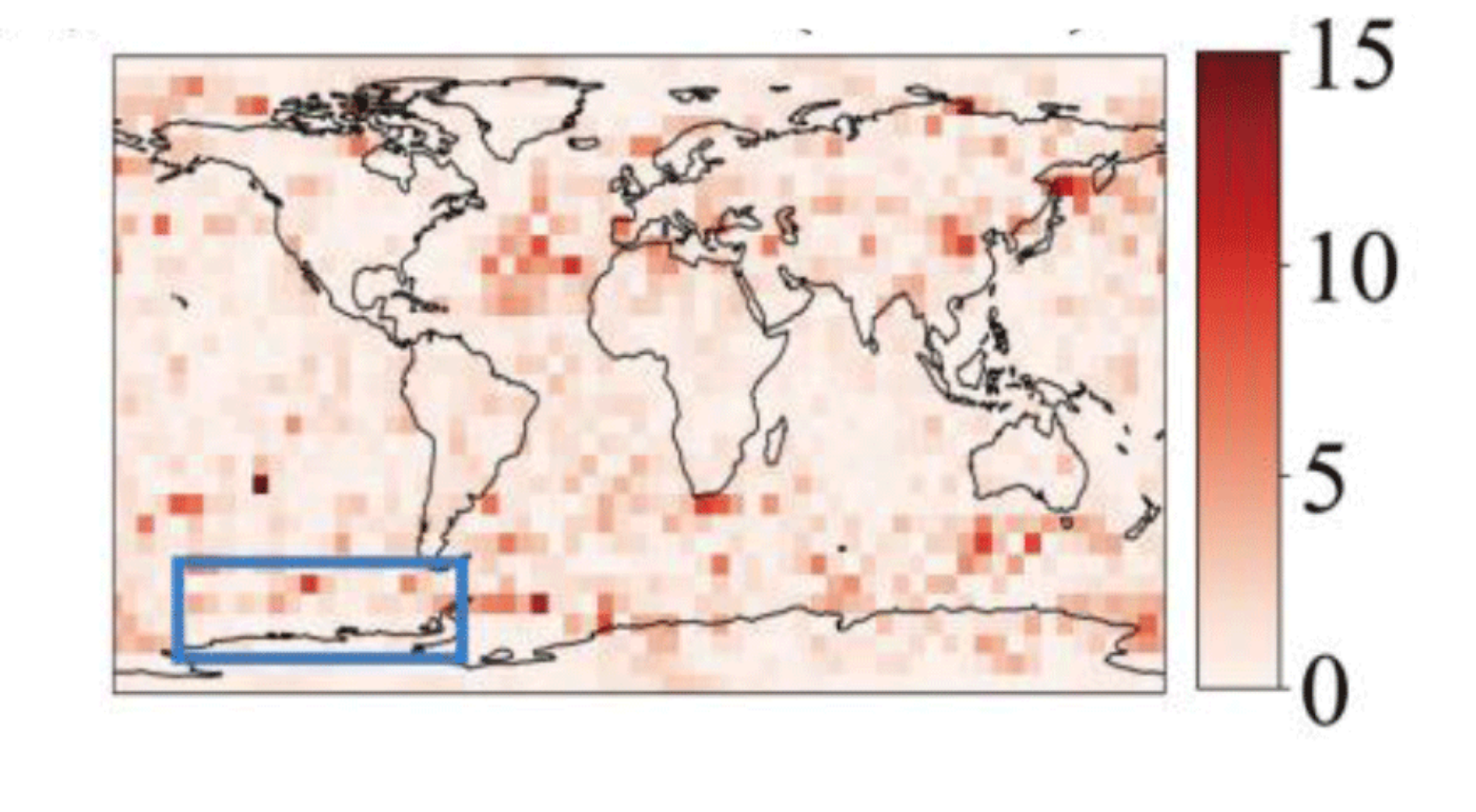}%
\label{e_d}}
\vspace{-1em}
\vfill
\subfloat[Ground Truth]{\includegraphics[width=0.27\textwidth]{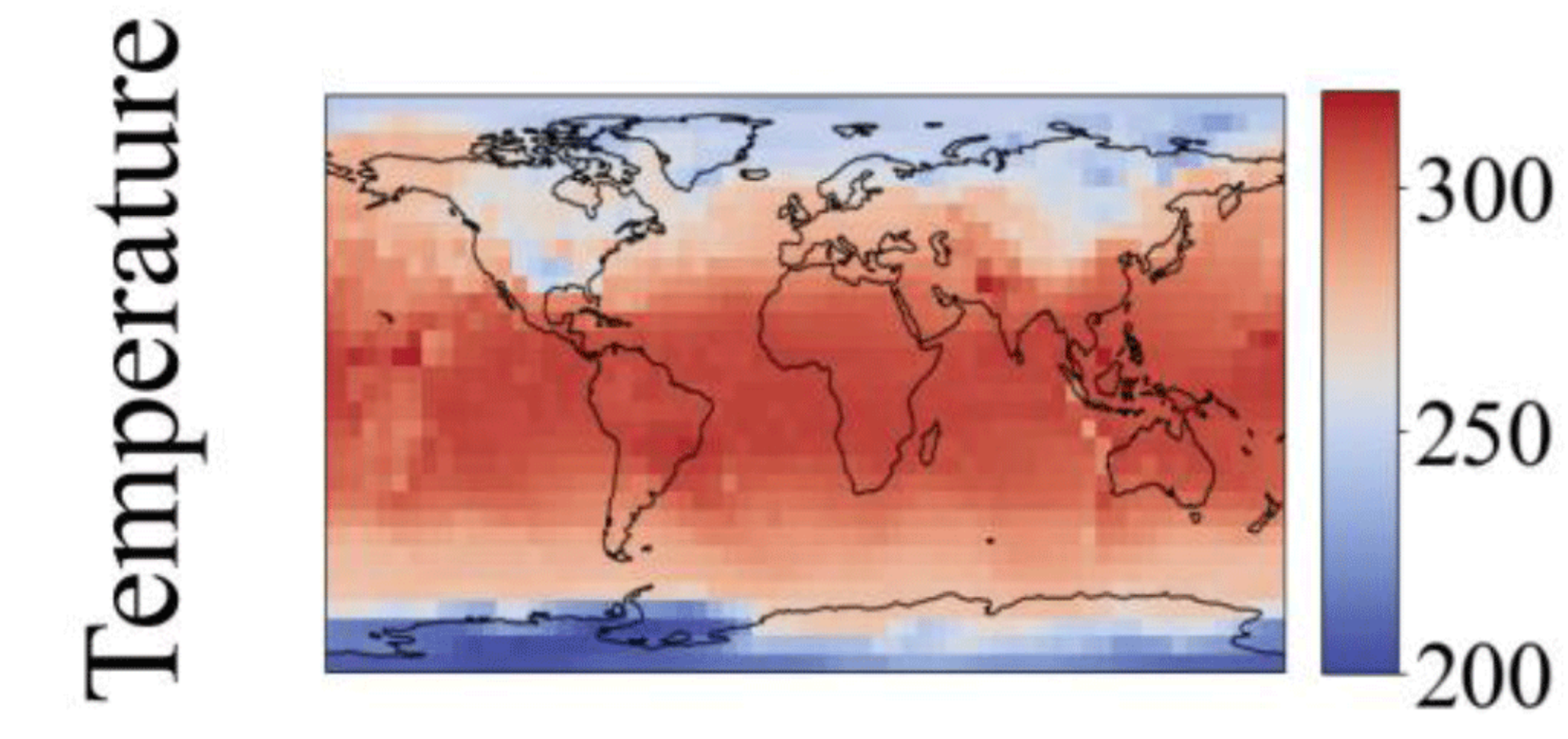}%
\label{e_e}}
\hfil
\subfloat[RGSL]{\includegraphics[width=0.23\textwidth]{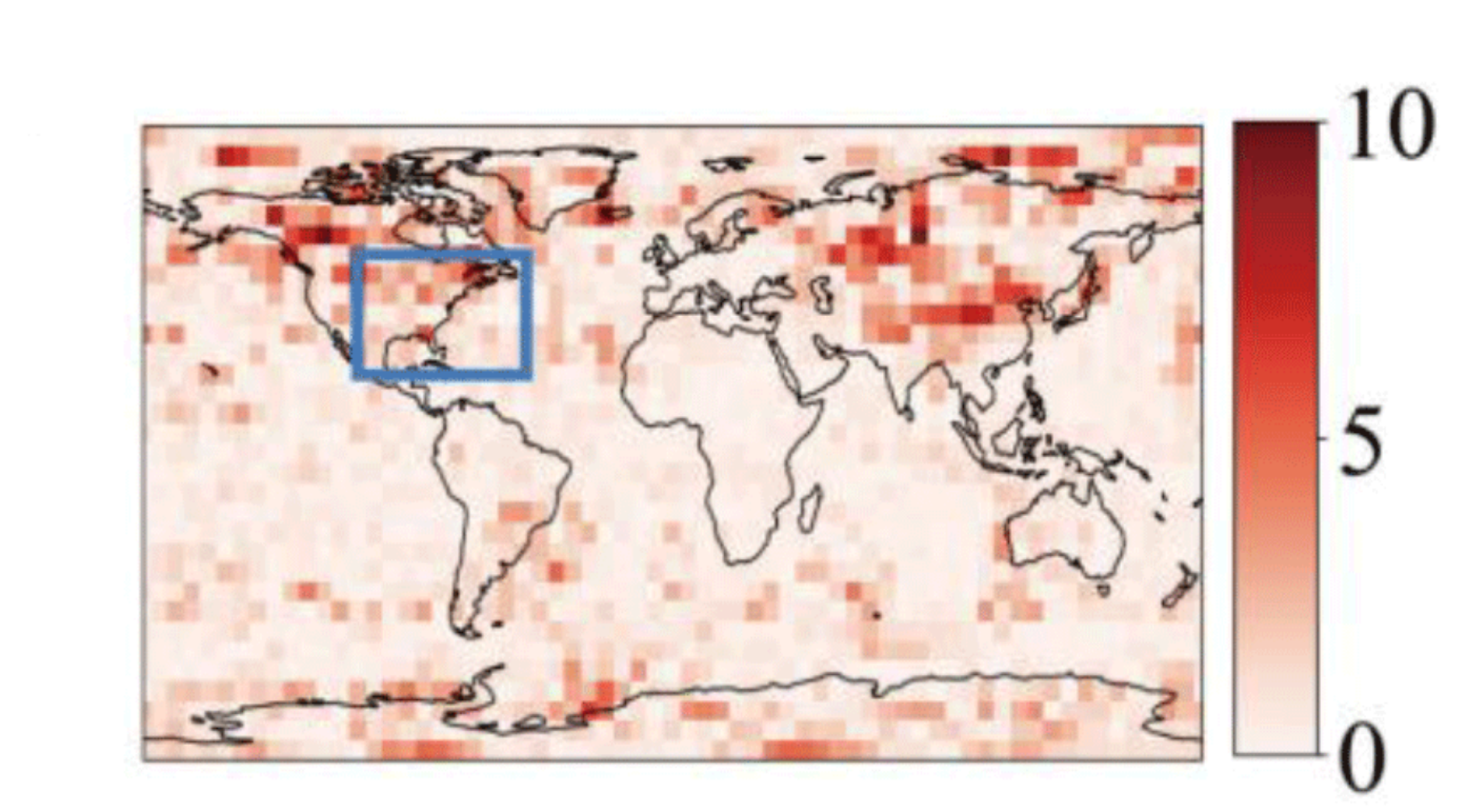}%
\label{e_f}}
\hfil
\subfloat[Corrfomer]{\includegraphics[width=0.23\textwidth]{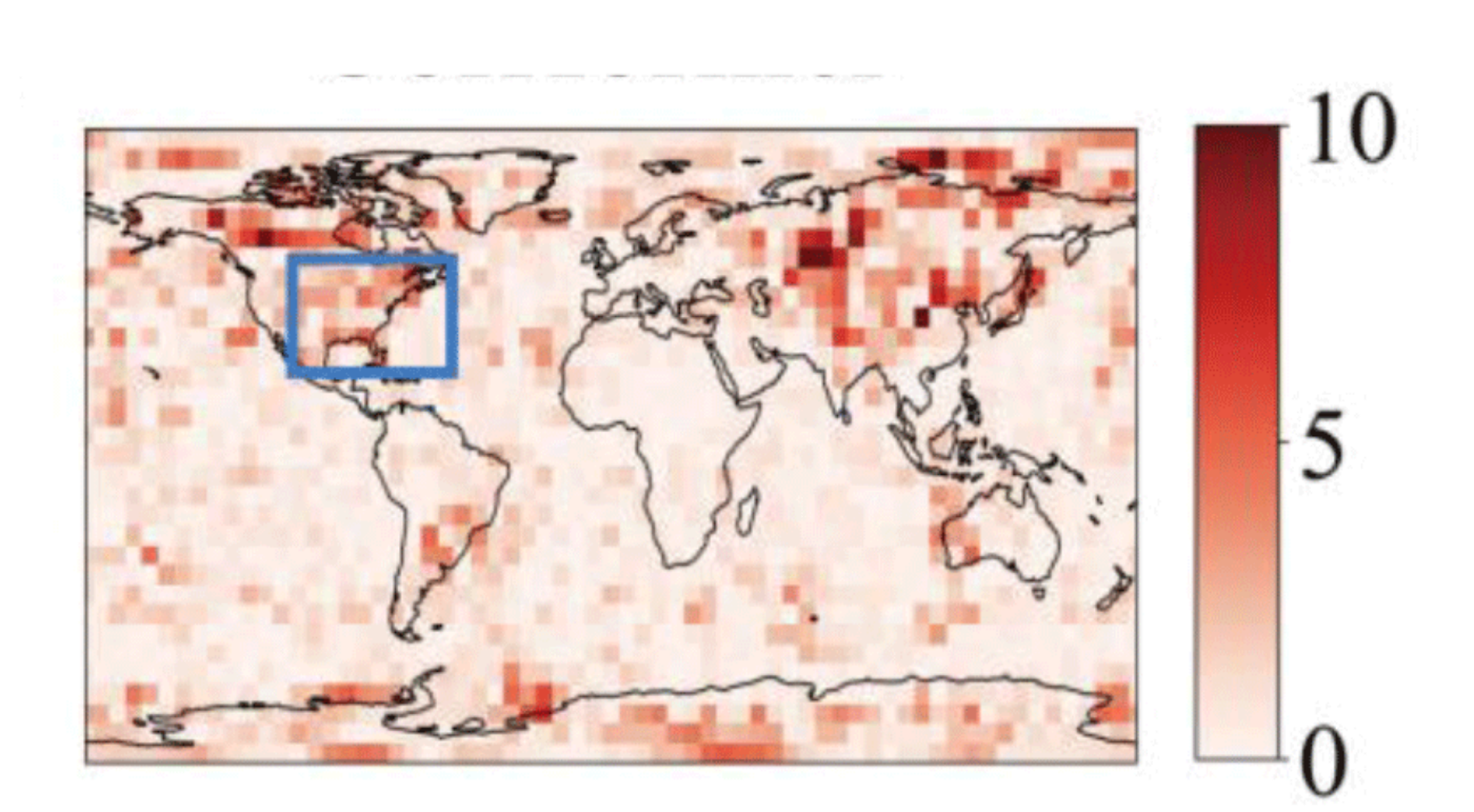}%
\label{e_g}}
\hfil
\subfloat[SAMSGL (Ours)]{\includegraphics[width=0.23\textwidth]{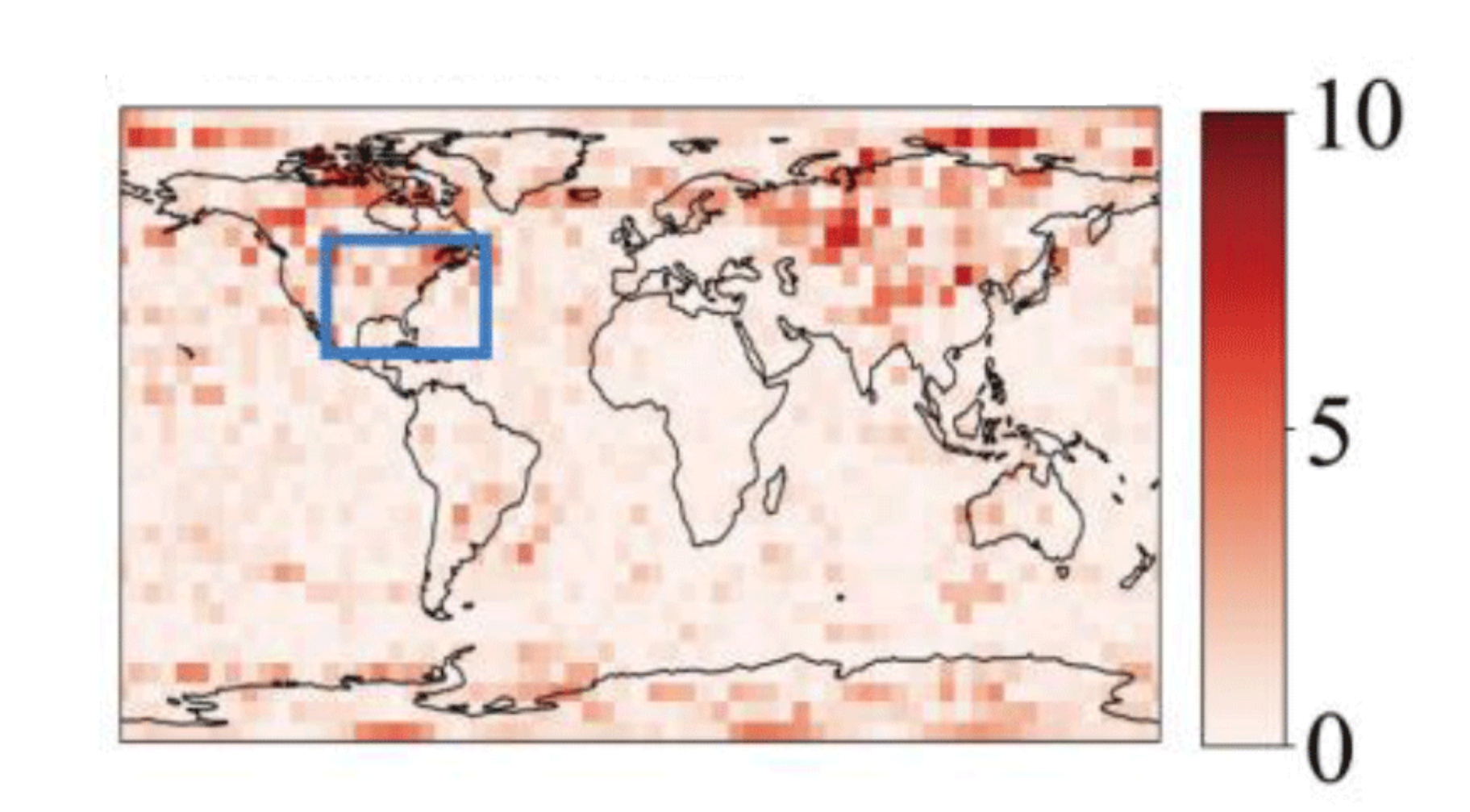}%
\label{e_h}}
\vspace{-1em}
\vfill
\subfloat[Ground Truth]{\includegraphics[width=0.27\textwidth]{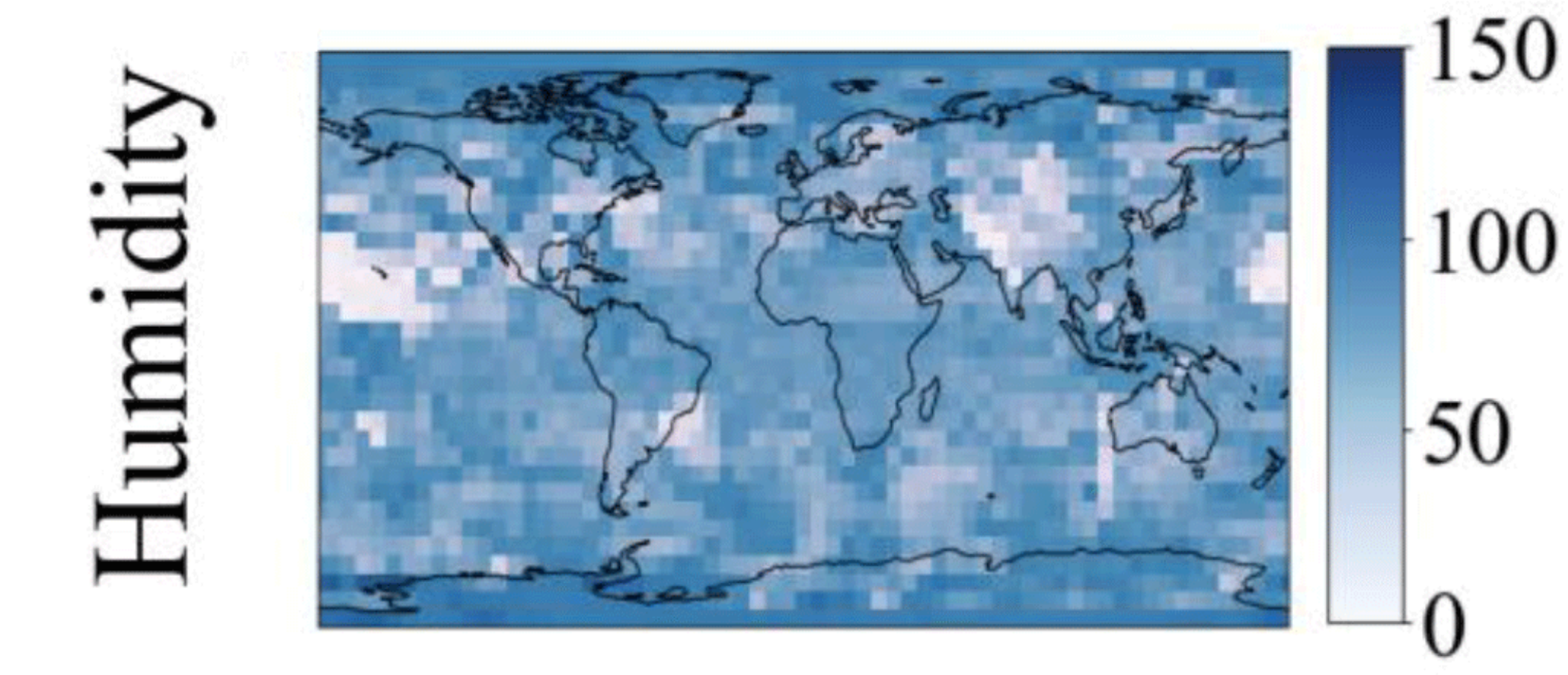}%
\label{e_i}}
\hfil
\subfloat[RGSL]{\includegraphics[width=0.23\textwidth]{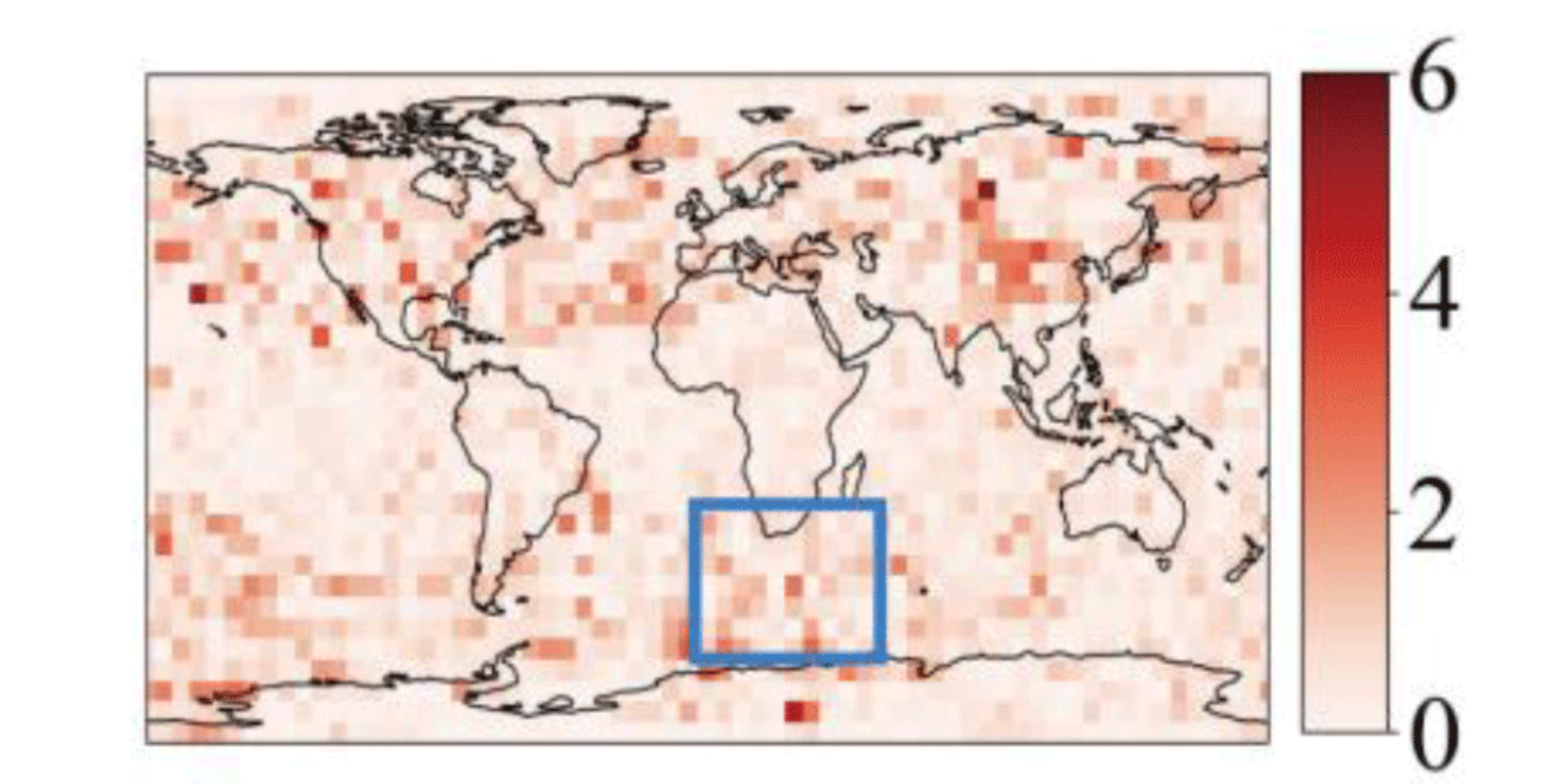}%
\label{e_j}}
\hfil
\subfloat[Corrfomer]{\includegraphics[width=0.23\textwidth]{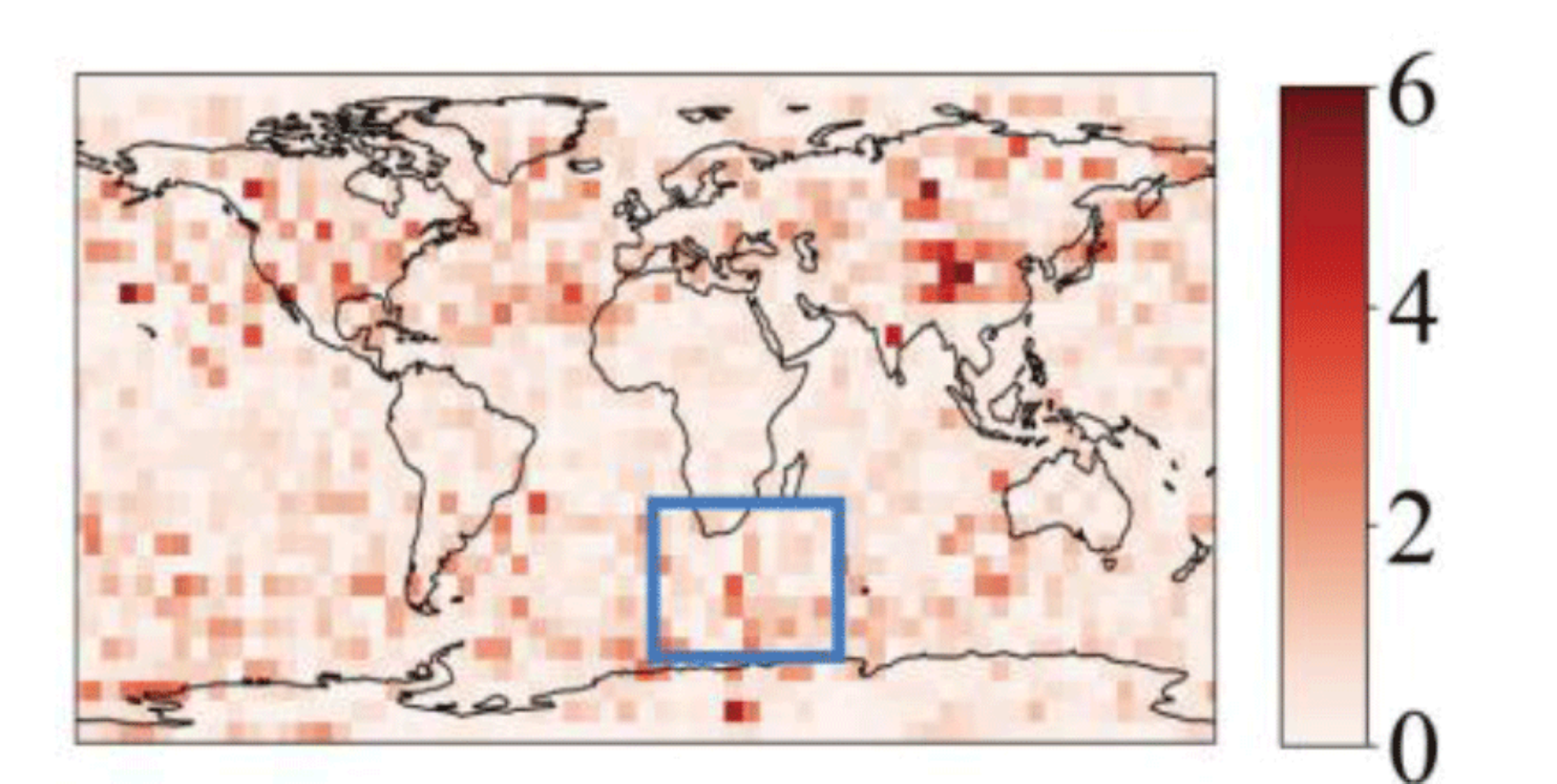}%
\label{e_k}}
\hfil
\subfloat[SAMSGL (Ours)]{\includegraphics[width=0.23\textwidth]{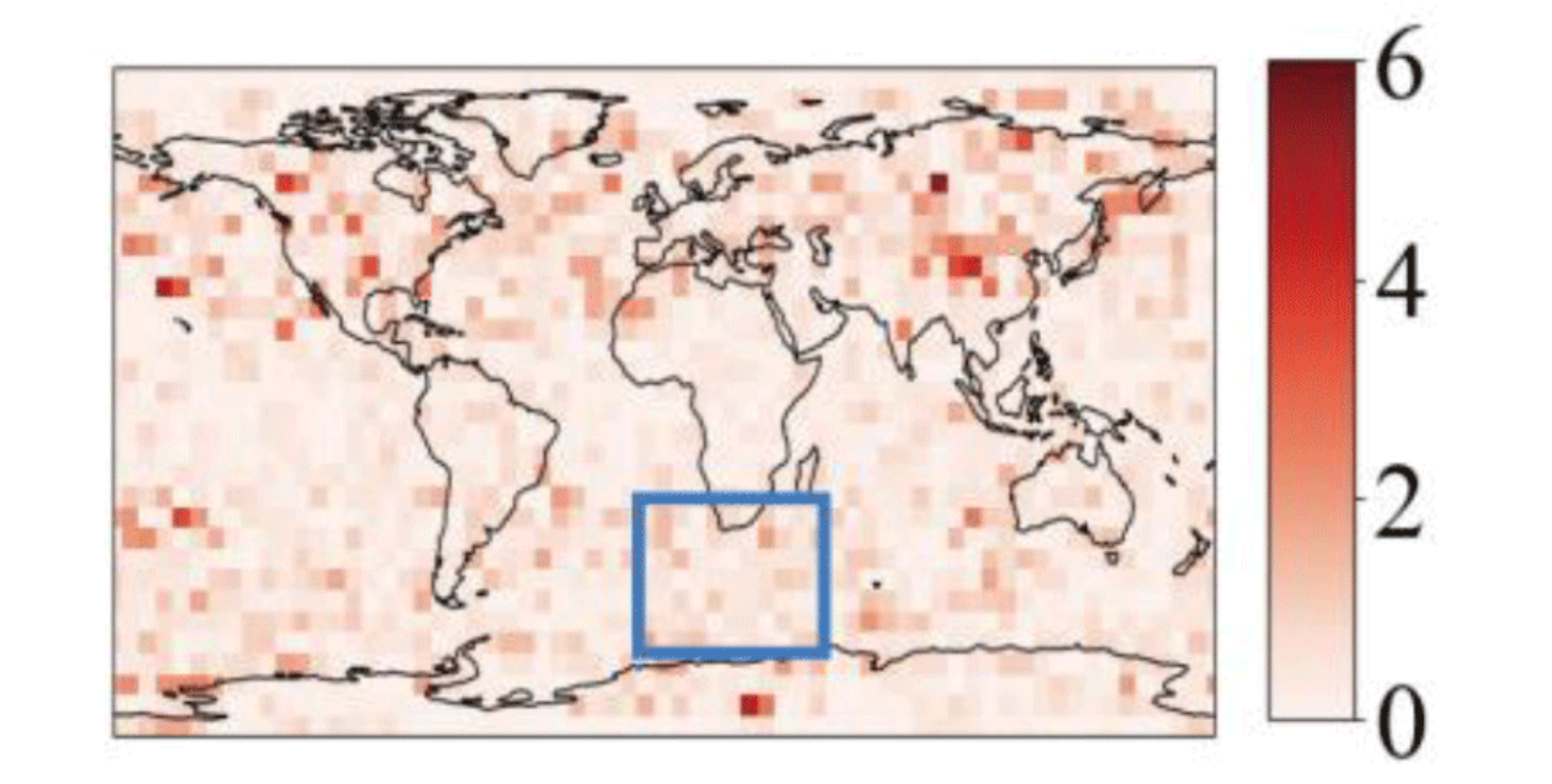}%
\label{e_l}}
\vspace{-1em}
\vfill
\subfloat[Ground Truth]{\includegraphics[width=0.27\textwidth]{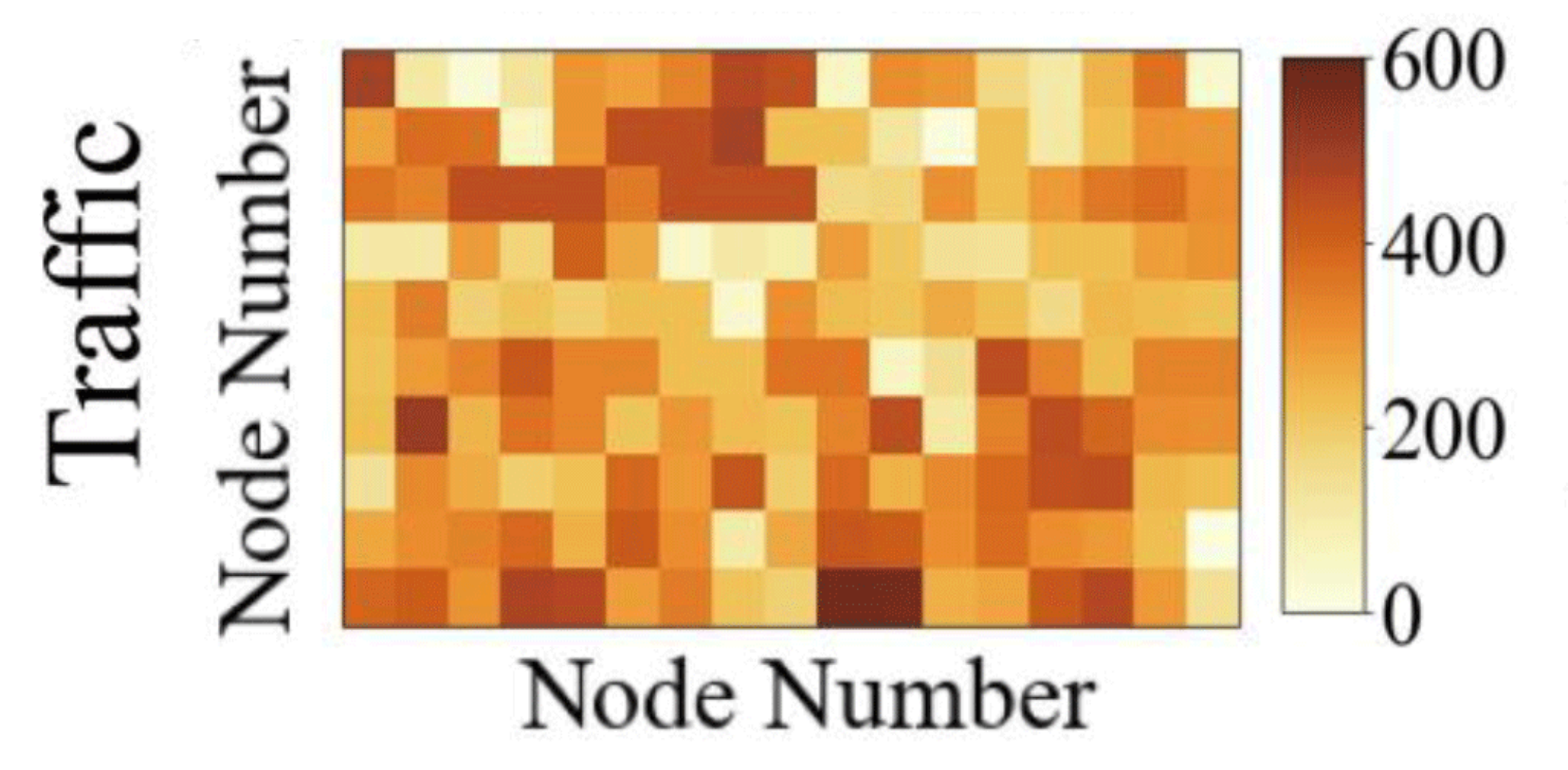}%
\label{e_m}}
\hfil
\subfloat[RGSL]{\includegraphics[width=0.23\textwidth]{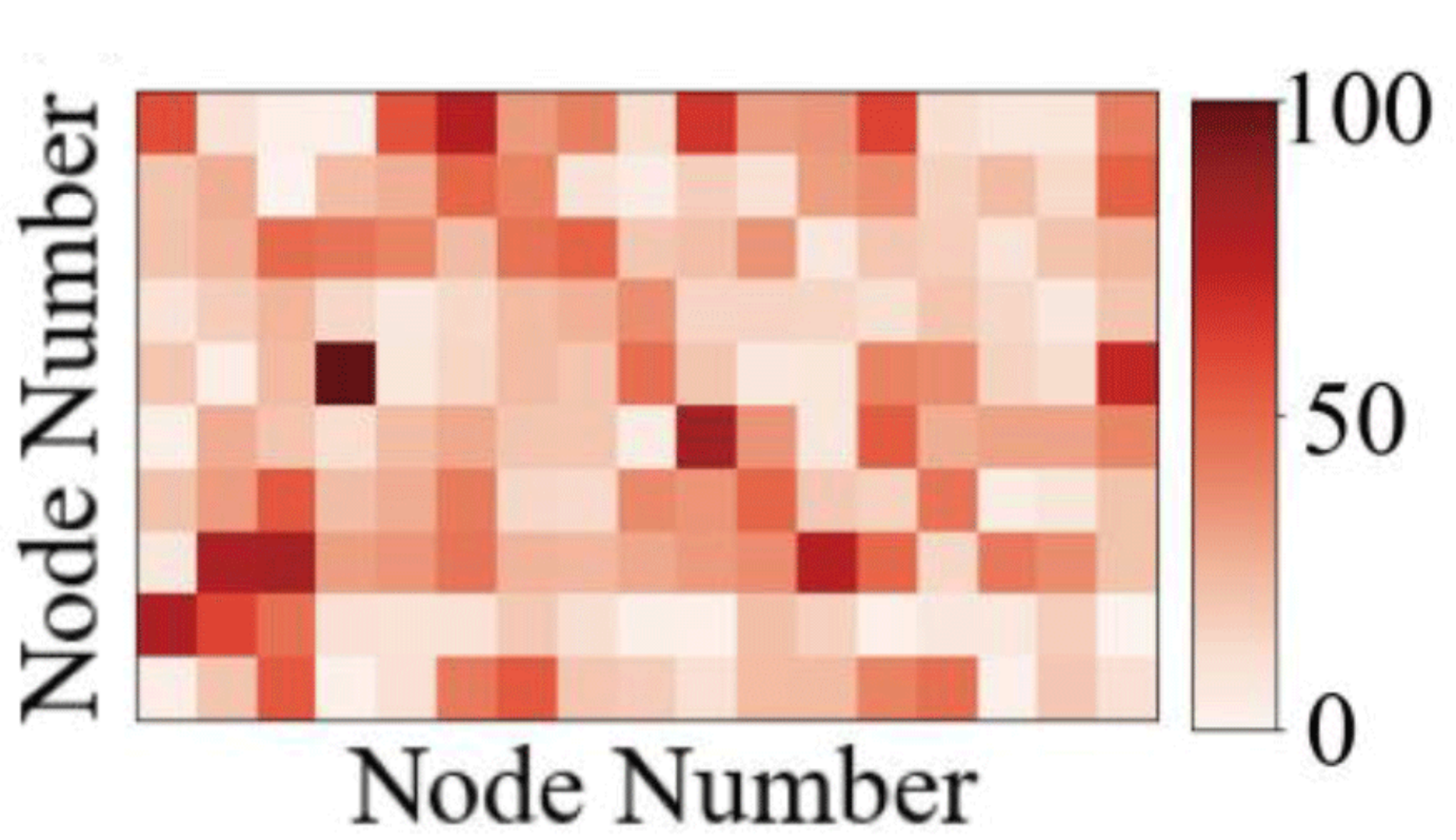}%
\label{e_n}}
\hfil
\subfloat[MGCRN]{\includegraphics[width=0.23\textwidth]{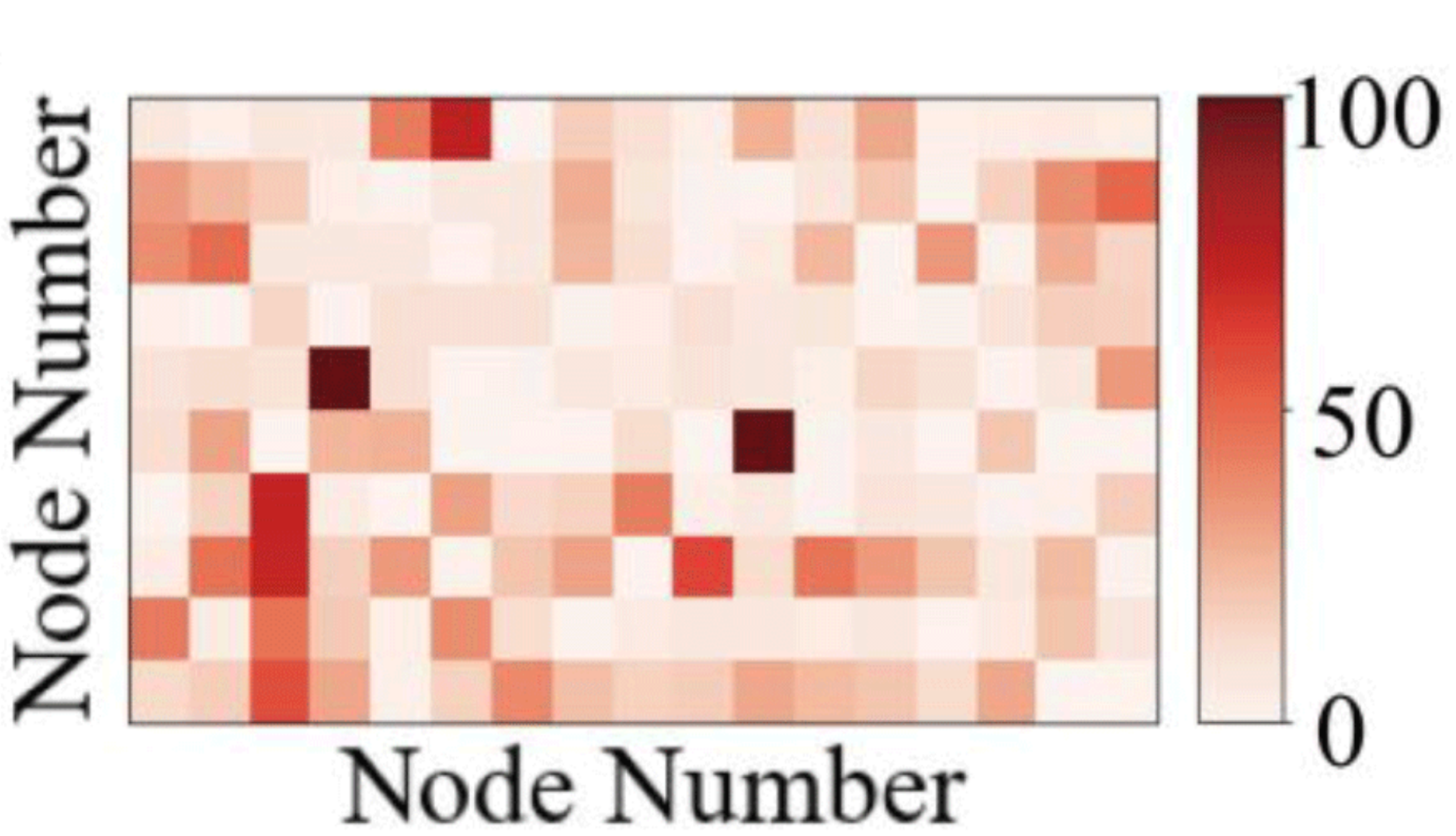}%
\label{e_o}}
\hfil
\subfloat[SAMSGL (Ours)]{\includegraphics[width=0.23\textwidth]{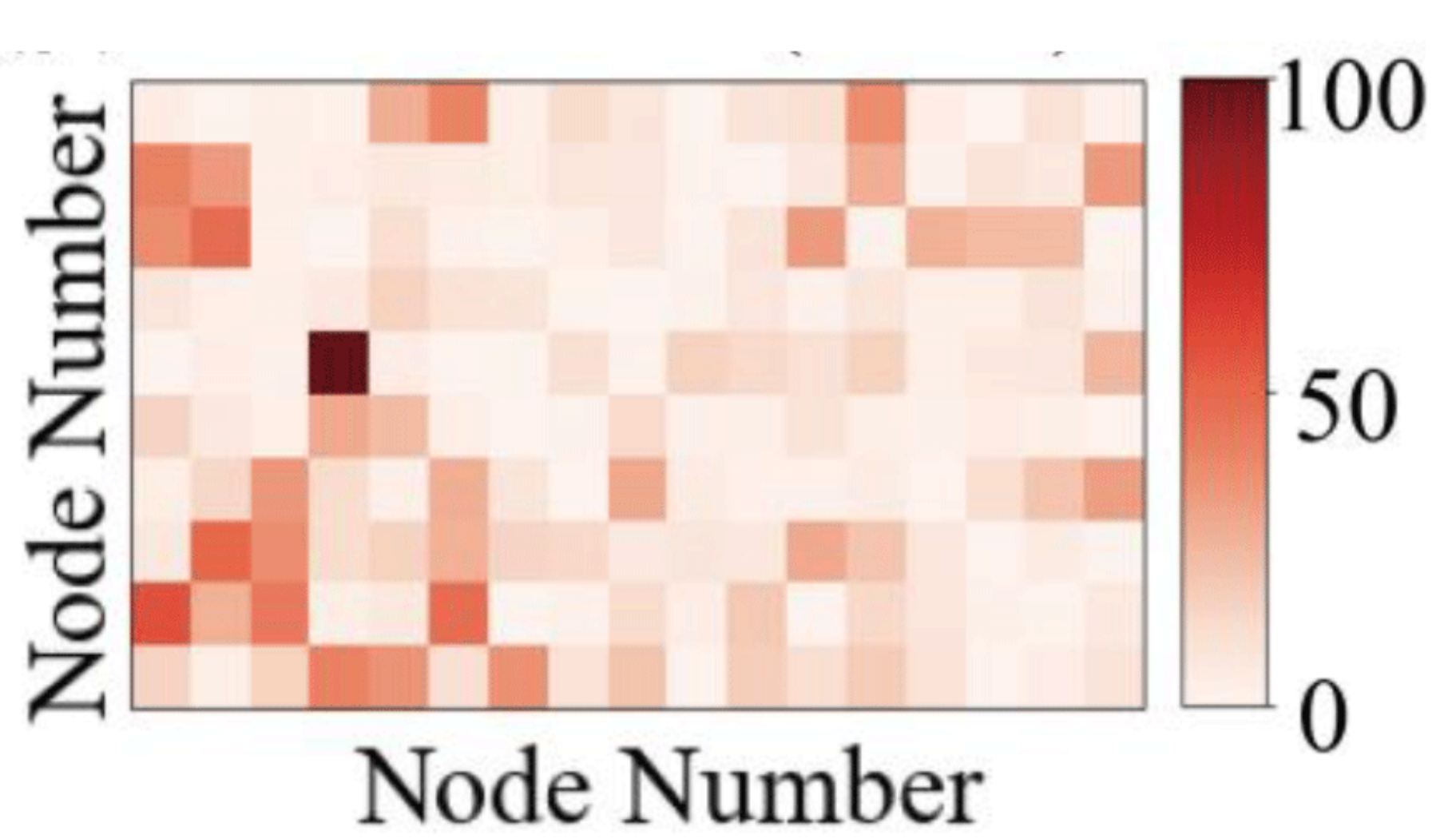}%
\label{e_p}}
\caption{Performance comparison on various spatio-temporal datasets. Panels (a), (e), (i), and (m) are the ground truths. (b-d), (f-h), and (j-l) show, in turn, the errors of RGSL, Corrformer and our SAMSGL model on wind speed, temperature and humidity datasets, respectively. (n-p) display the errors of RGSL, MGCRN and our SAMSGL model on traffic flow dataset, respectively. The blue boxes in panels (b)-(d), (f)-(h), and (j)-(l) highlight regions with significant errors.}
\label{fig_0}
\end{figure*}
\begin{figure*}[!htbp]
\centering
\subfloat[Ground Truth]{\includegraphics[height=1.6in]{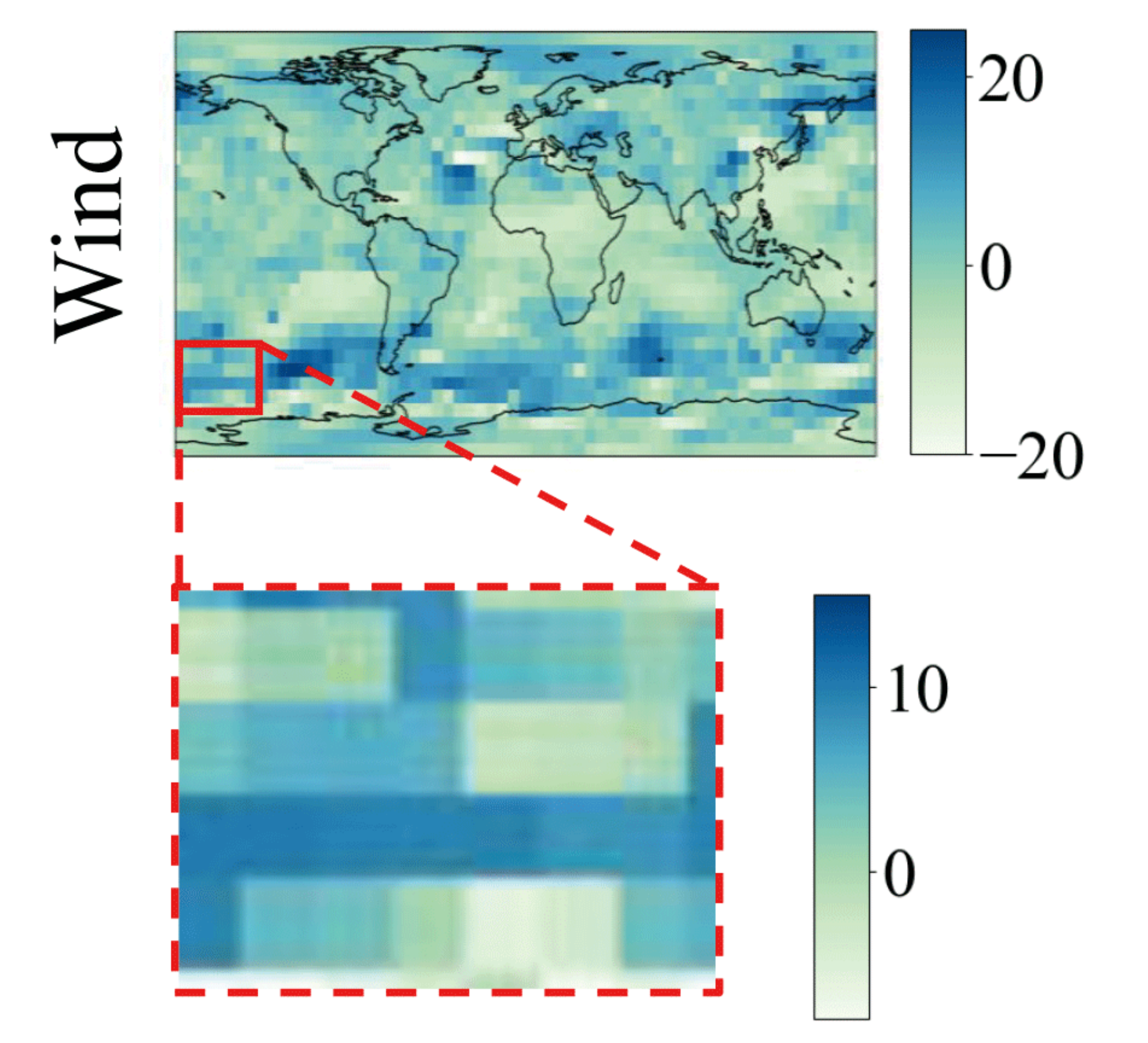}%
\label{g_a}}
\hfil
\subfloat[SAMGSL (Ours)]{\includegraphics[height=1.6in]{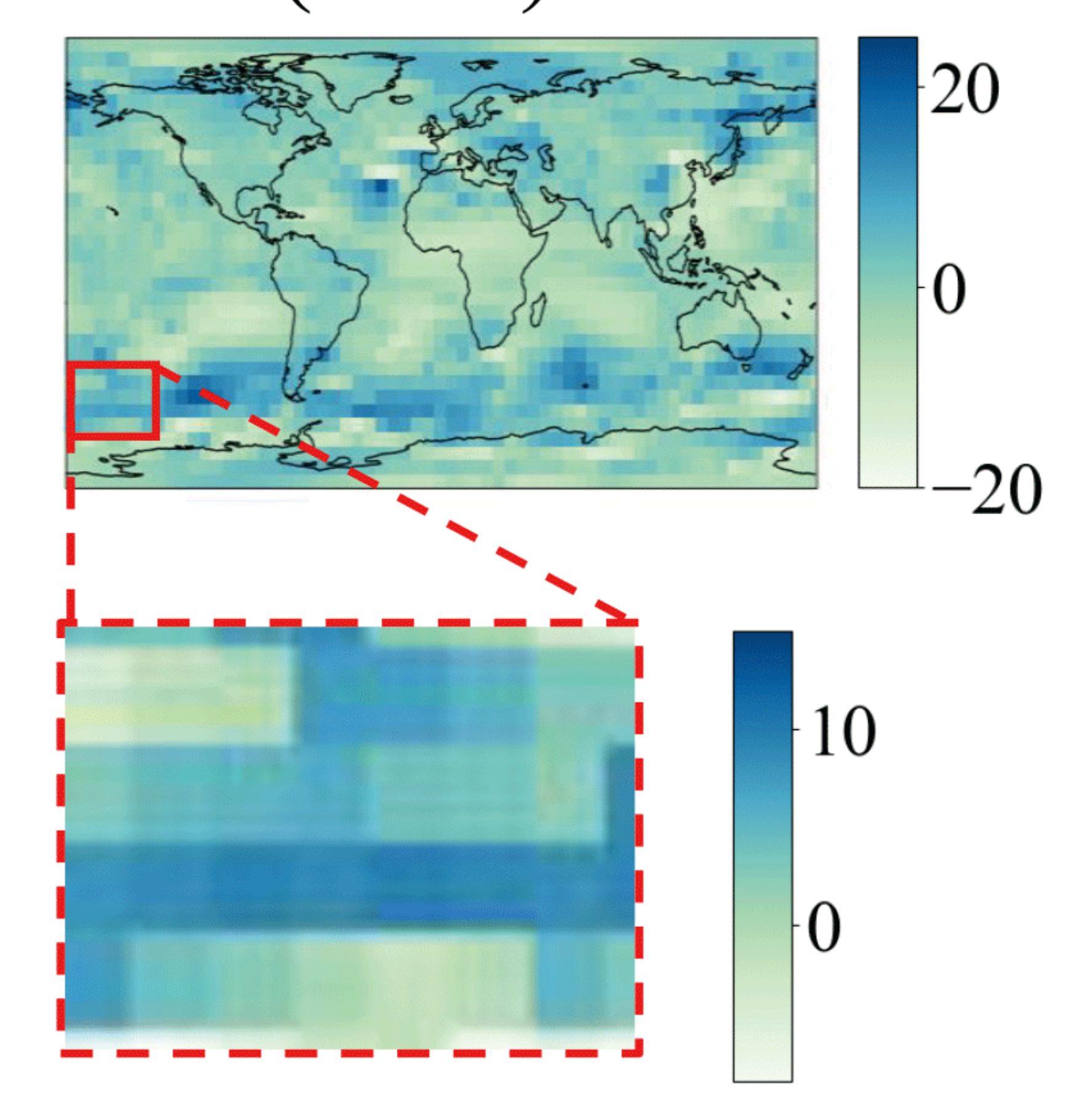}%
\label{g_b}}
\hfil
\subfloat[Corrformer \cite{wu2021autoformer}]{\includegraphics[height=1.6in]{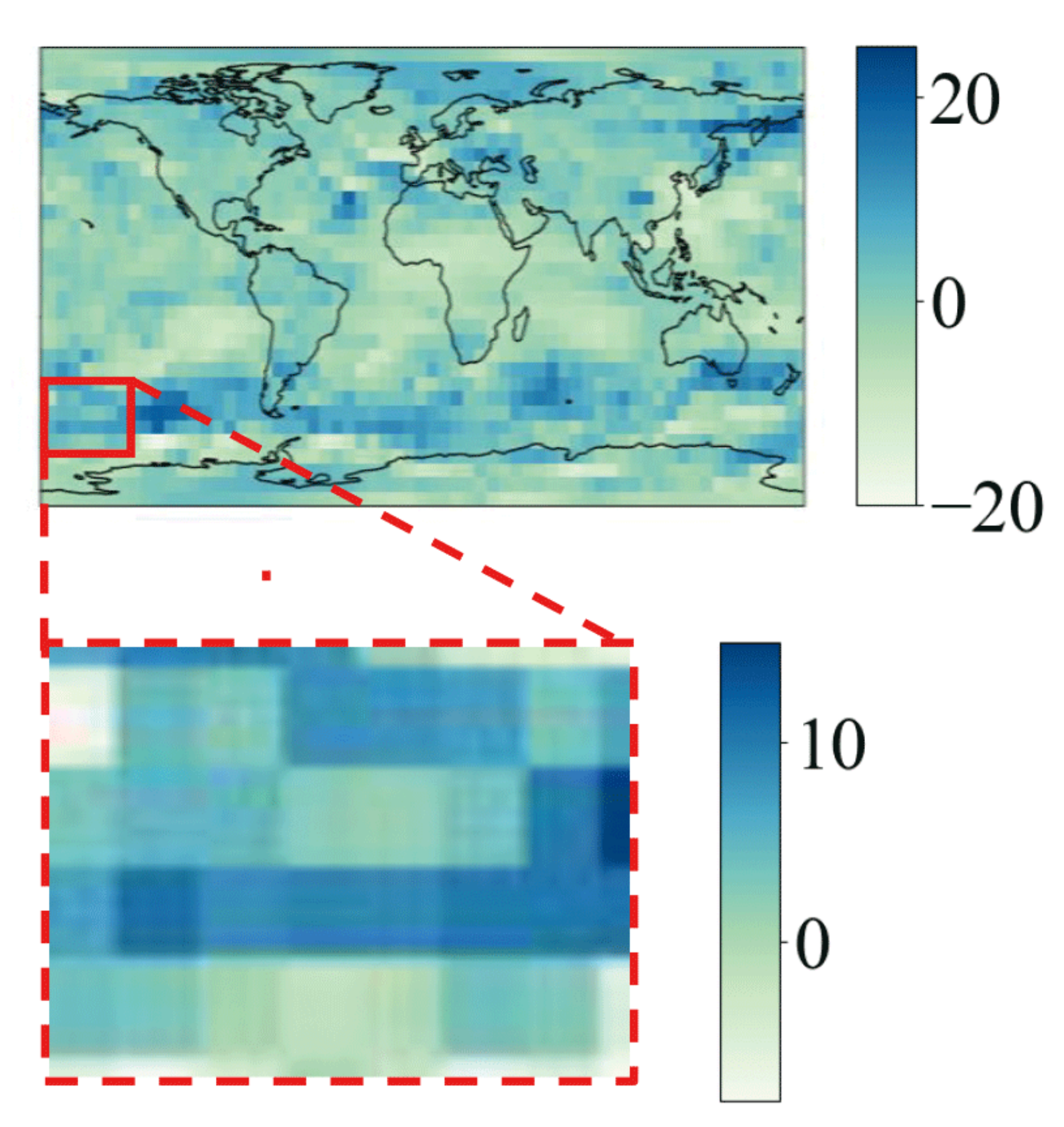}%
\label{g_c}}
\hfil
\subfloat[RGSL \cite{yu2022regularized}]{\includegraphics[height=1.6in]{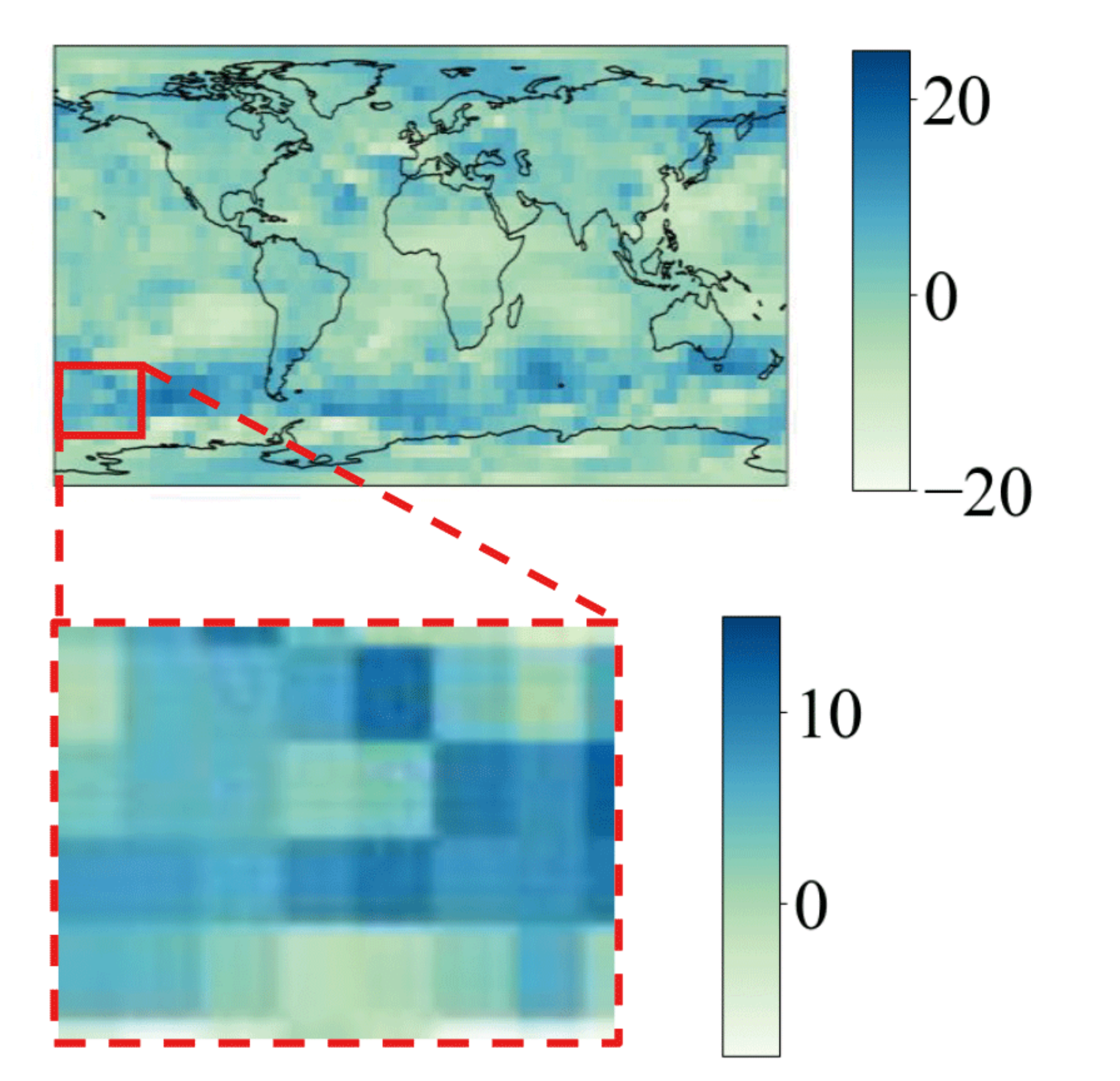}%
\label{g_d}}
\caption{Comparison of forecasting performance at a time step on wind dataset. The red boxes zoom in on typical regions to illustrate the differences in predicted spatial patterns.}
\label{spatial}
\end{figure*}
% \begin{figure*}[!htbp]
% \centering
% \subfloat[]{\includegraphics[width=0.45\textwidth]{ts/ts_add_01}%
% \label{ts-a}}
% \hfil
% \subfloat[]{\includegraphics[width=0.45\textwidth]{ts/ts_add_02}%
% \label{ts-b}}
% \vspace{-1em}
% \vfill
% \subfloat[]{\includegraphics[width=0.45\textwidth]{ts/ts_add_03}%
% \label{ts-c}}
% \hfil
% \subfloat[]{\includegraphics[width=0.45\textwidth]{ts/ts_add_04}%
% \label{ts-d}}
% \vspace{-1em}
% \vfill
% \subfloat[]{\includegraphics[width=0.45\textwidth]{ts/ts_add_05}%
% \label{ts-e}}
% \vspace{-1em}
% \hfil
% \subfloat[]{\includegraphics[width=0.45\textwidth]{ts/ts_add_06}%
% \label{ts-f}}
% \caption{Comparison of forecasting performance in single node's temporal dynamics. (a), (b) are the predicted time series on Temperature dataset. (c), (d) are the predicted time series on Wind  dataset. (e), (f) are the predicted time series on PeMSD8 dataset.}
% \label{ts_6}
% \end{figure*}
\begin{figure*}[!htbp]
\centering
\captionsetup[subfloat]{}%{labelfont={color=blue}}
\subfloat[]{\includegraphics[width=0.33\textwidth]{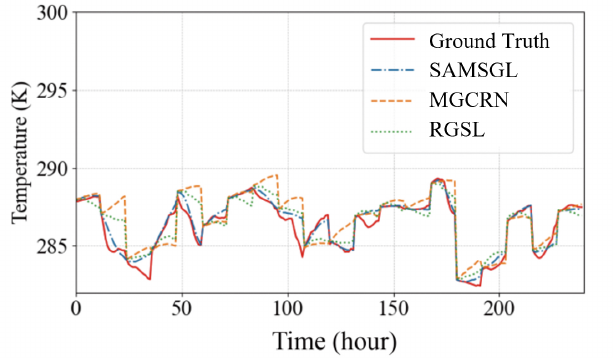}%
\label{ts-a}}
\hfil
\subfloat[]{\includegraphics[width=0.33\textwidth]{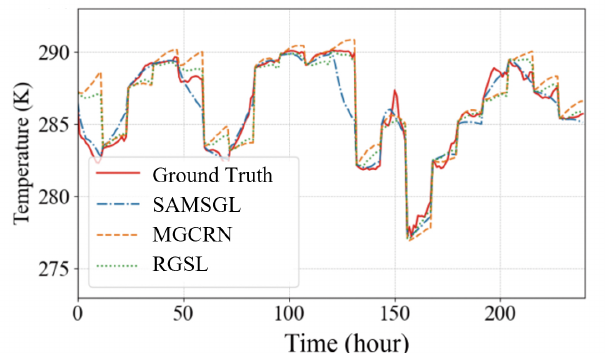}%
\label{ts-b}}
\hfil
\subfloat[]{\includegraphics[width=0.33\textwidth]{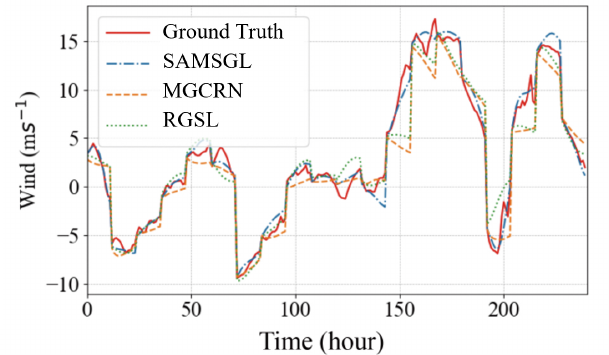}%
\label{ts-c}}
\vspace{-1em}
\vfill
\subfloat[]{\includegraphics[width=0.33\textwidth]{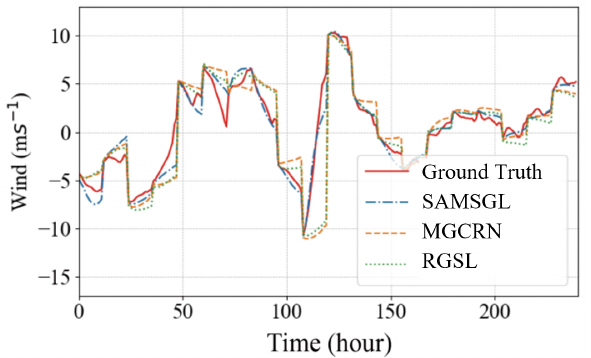}%
\label{ts-d}}
\hfil
\subfloat[]{\includegraphics[width=0.33\textwidth]{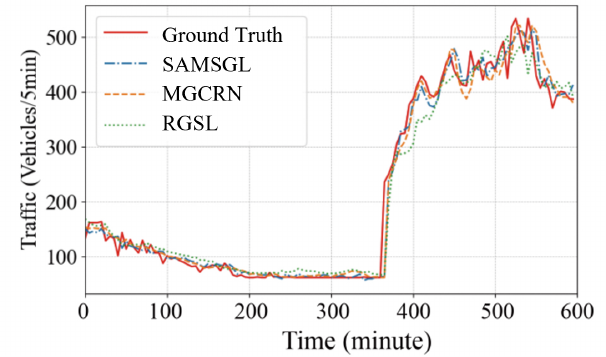}%
\label{ts-e}}
\hfil
\subfloat[]{\includegraphics[width=0.33\textwidth]{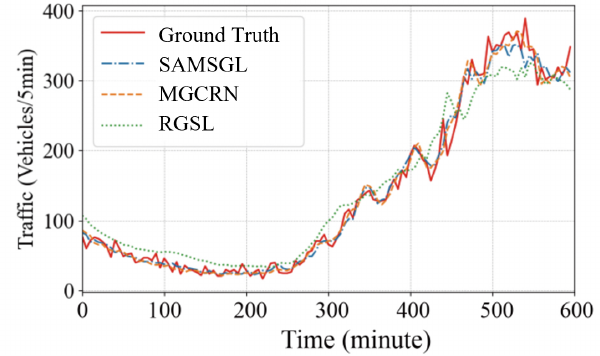}%
\label{ts-f}}
\caption{Comparison of forecasting performance in single node's temporal dynamics. (a), (b) are the predicted time series on Temperature dataset. (c), (d) are the predicted time series on Wind  dataset. (e), (f) are the predicted time series on PeMSD8 dataset.}
\label{ts_6}
\end{figure*}
\begin{figure}[!h]
\centering
\includegraphics[width=\linewidth]{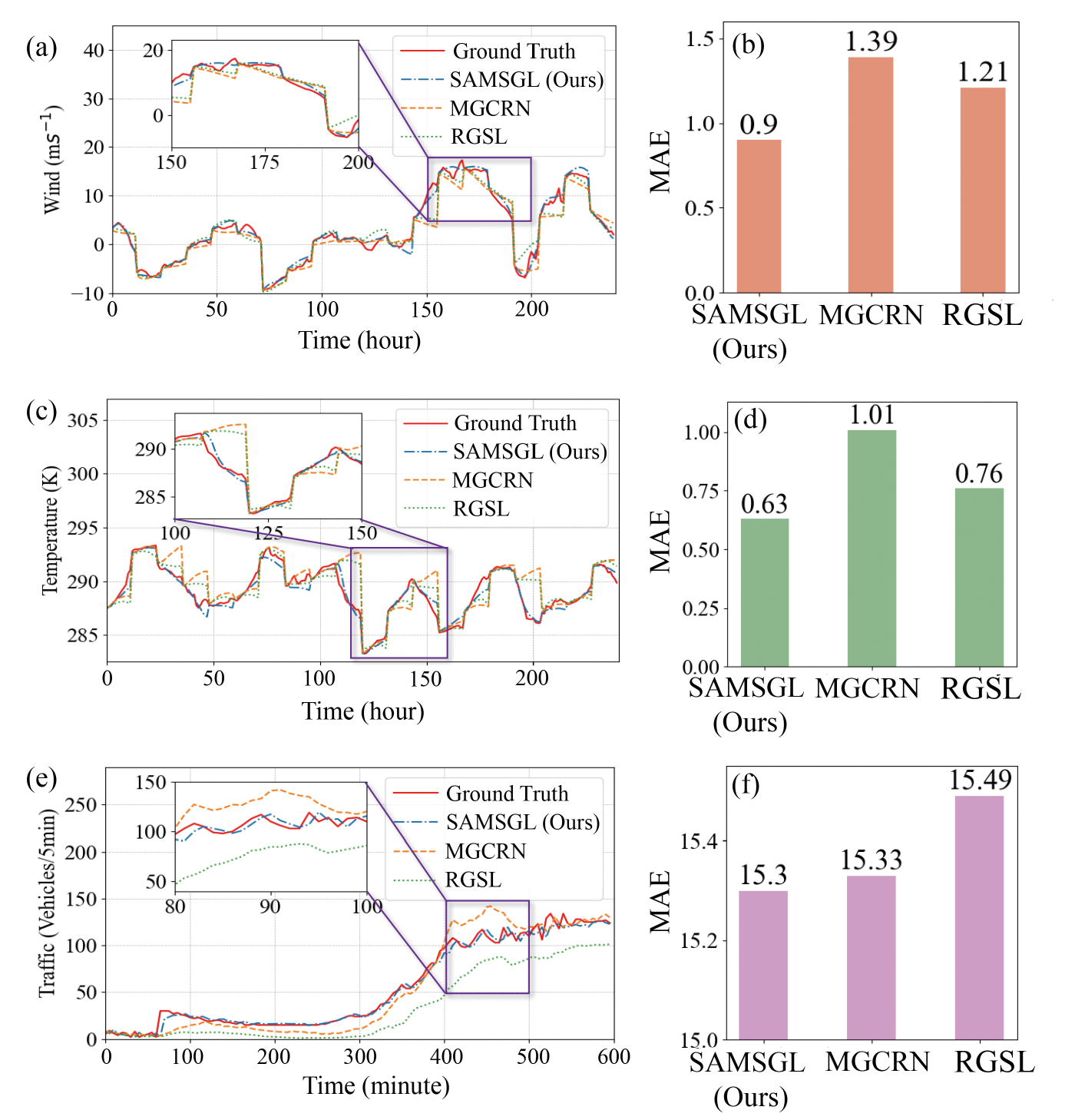} % Reduce the figure size so that it is slightly narrower than the column.
\caption{Comparison of spatio-temporal forecasting performance among SAMSGL (ours), MGCRN and RGSL on various datasets. (a), (c), and (e) present the prediction results on the Wind dataset, the Temperature dataset, and the PeMSD8 dataset, respectively. (b), (d), and (f) display the MAE of the methods on the Wind dataset, the Temperature dataset, and the PeMSD8 dataset, respectively.}
\label{wind_global}
\end{figure}
\begin{figure}[htbp]
\centering
\subfloat[]{\includegraphics[width=0.9\linewidth]{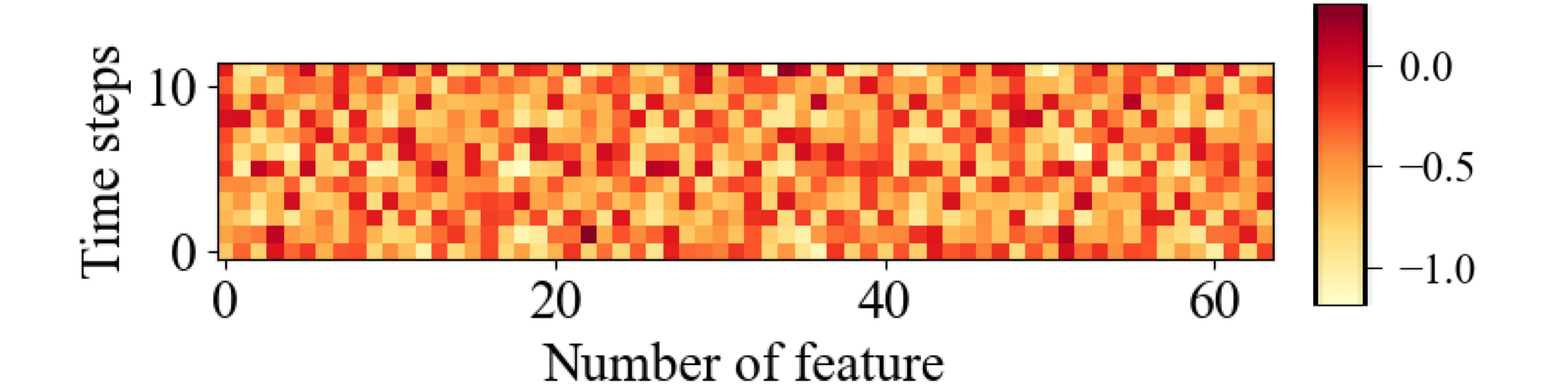}%
\label{ms-a}}
\vspace{-1em}
\vfill
\subfloat[]{\includegraphics[width=0.9\linewidth]{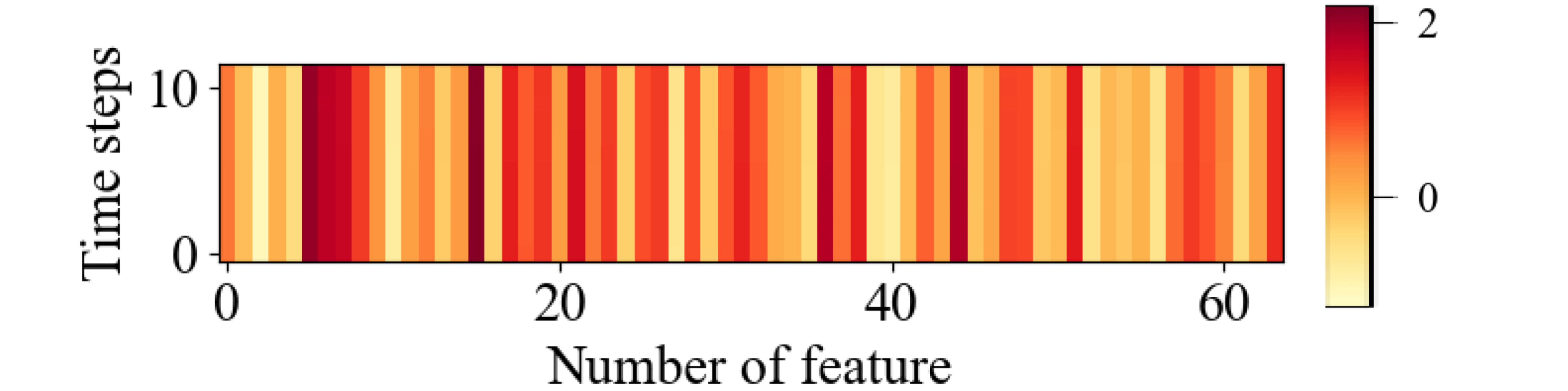}%
\label{ms-b}}
\vspace{-1em}
\vfill
\subfloat[]{\includegraphics[width=0.9\linewidth]{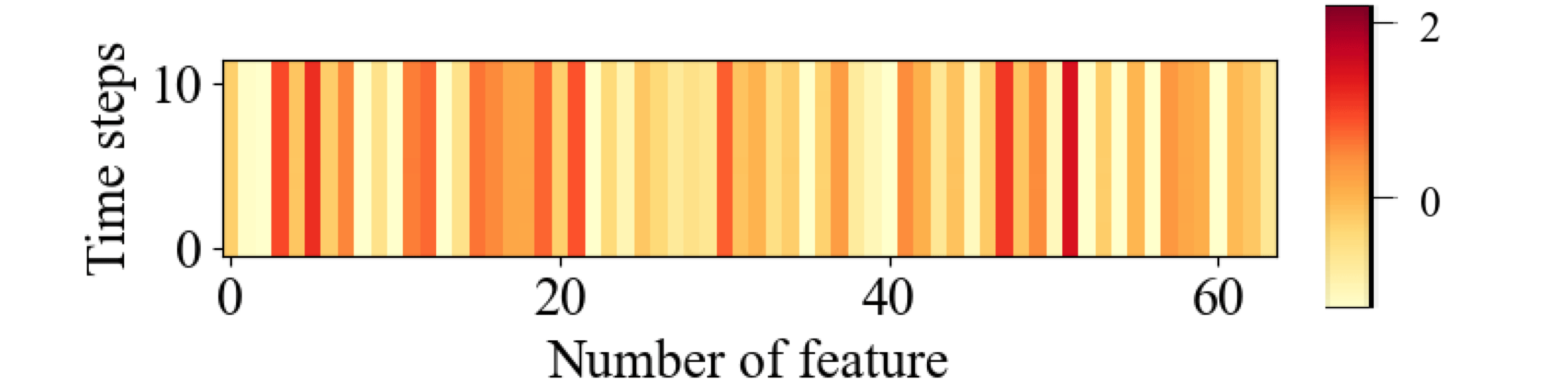}
\label{ms-c}}
\vspace{-1em}
\vfill
\subfloat[]{\includegraphics[width=0.9\linewidth]{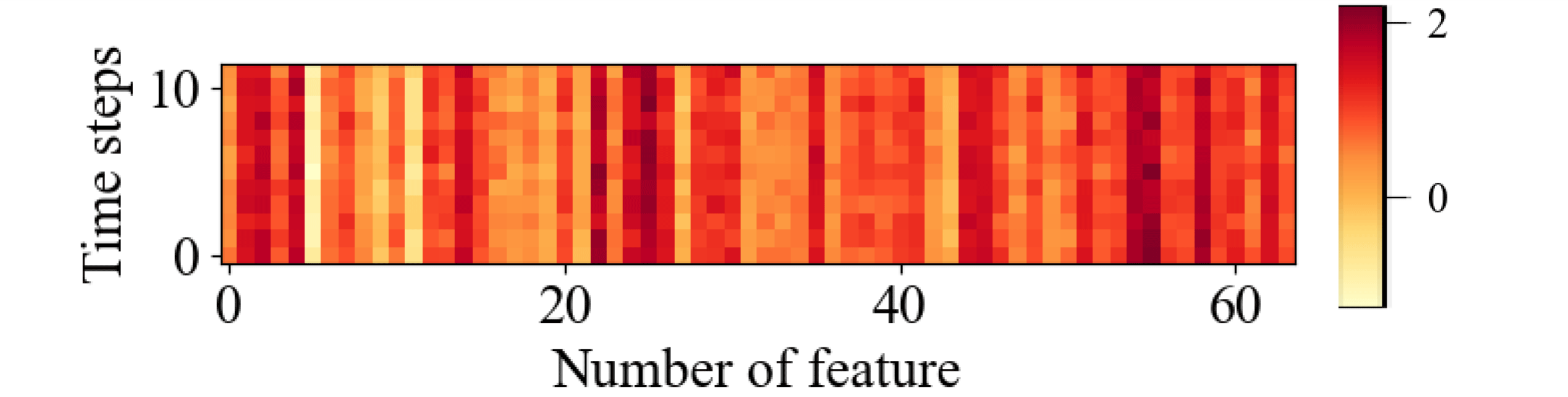}
\label{ms-d}}
\caption{Visualization of multi-graph convolution. Panel (a) illustrates the output of series-aligned graph convolution layer. Panel (b) displays the output of delayed graph convolution. Panel (c) presents the output of local graph convolution. Panel (d) is the output of the whole multi-graph convolution. }
\label{ms}
\end{figure}
\begin{figure*}[htbp]
\centering
\captionsetup[subfloat]{}%{labelfont={color=blue}}
\subfloat[]{\includegraphics[width=0.25\textwidth]{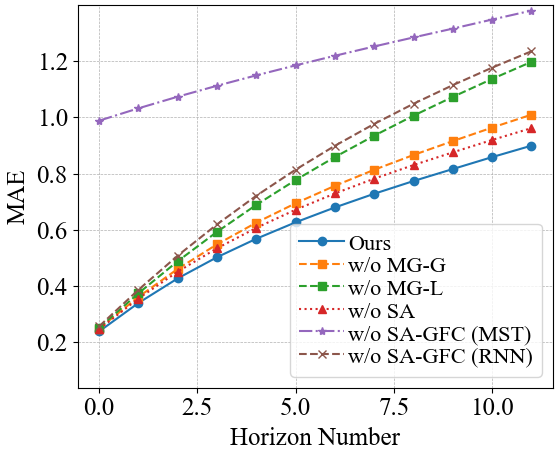}%
\label{abl-a}}
\hfil
\subfloat[]{\includegraphics[width=0.25\textwidth]{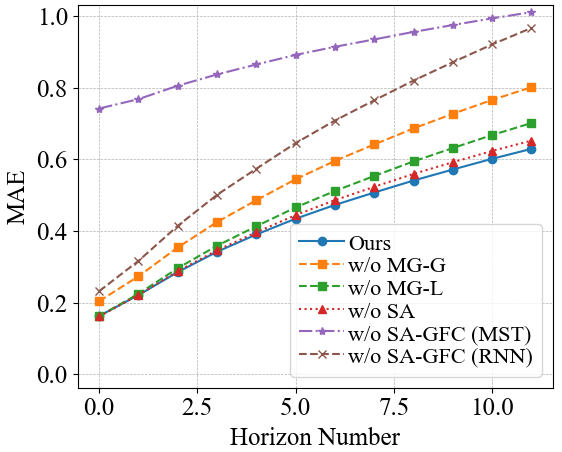}%
\label{abl-c}}
\hfil
\subfloat[]{\includegraphics[width=0.25\textwidth]{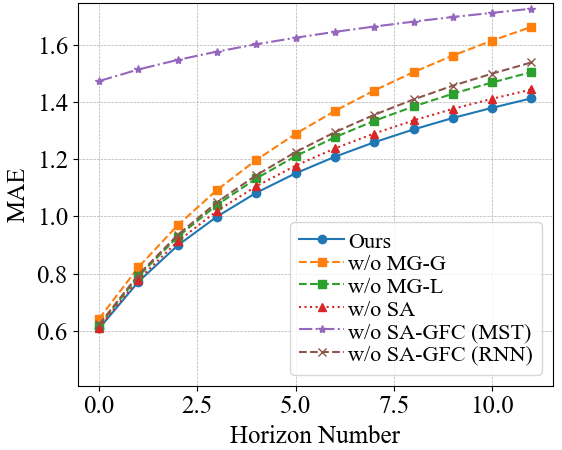}%
\label{abl-e}}
\hfil
\subfloat[]{\includegraphics[width=0.25\textwidth]{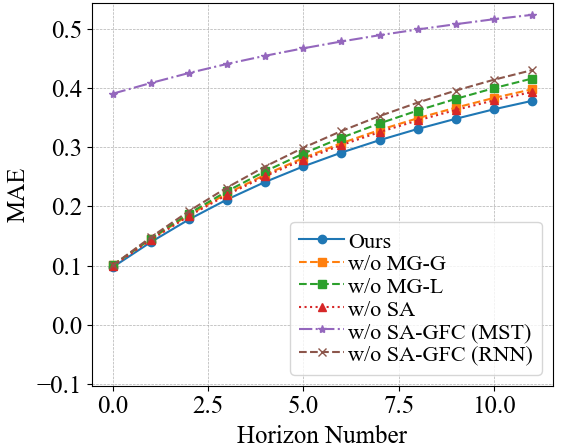}%
\label{abl-g}}
\vspace{-1em}
\vfill
\subfloat[]{\includegraphics[width=0.25\textwidth]{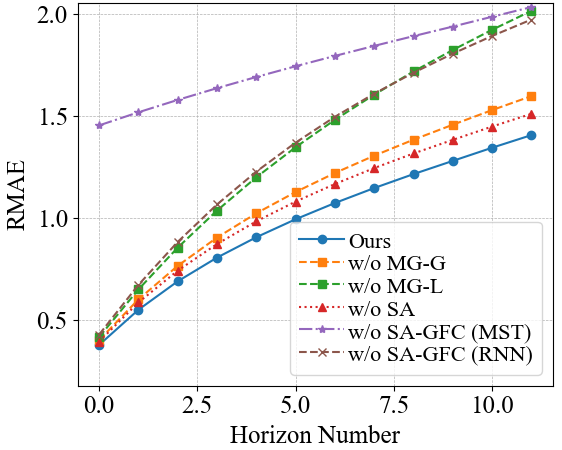}%
\label{abl-b}}
\hfil
\subfloat[]{\includegraphics[width=0.25\textwidth]{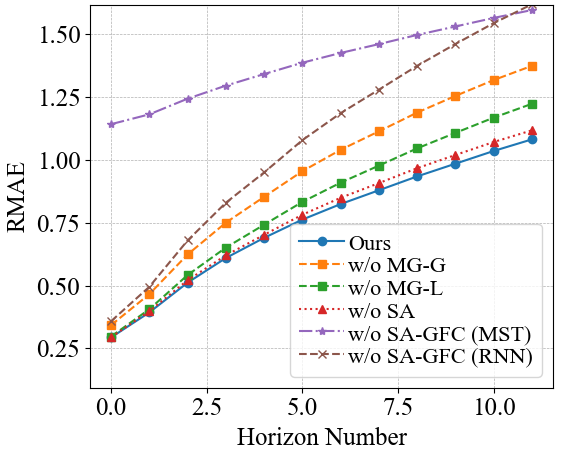}%
\label{abl-d}}
\hfil
\subfloat[]{\includegraphics[width=0.25\textwidth]{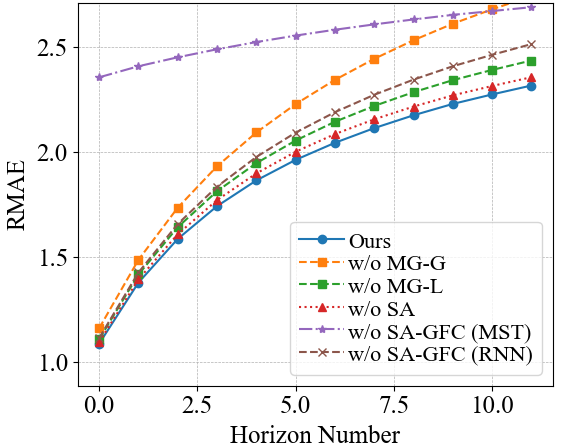}%
\label{abl-f}}
\hfil
\subfloat[]{\includegraphics[width=0.25\textwidth]{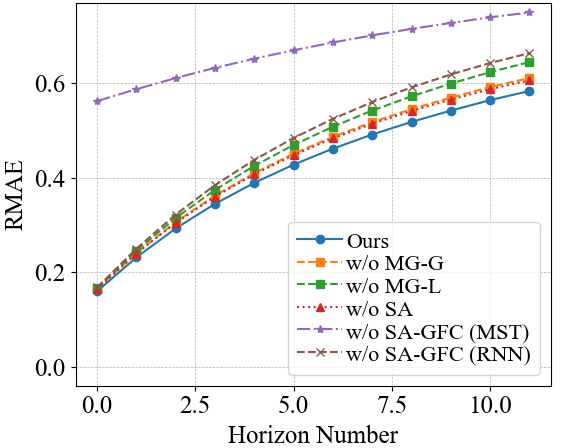}%
\label{abl-h}}
\caption{Ablation studies. The comparison is conducted at each time step in prediction period on (a), (e) the Wind  dataset, (b), (f) the Temperature dataset, (c), (g) the Cloud cover dataset, (d), (h) the humidity dataset.}
\label{abl_all}
\end{figure*}
\subsection{Baseline Methods}
We conduct a comprehensive comparison of our model against two categories models: RNN-based models combining GCN and RNN for capturing temporal and dynamic features,  and time-space alternation models that alternate between temporal and spatial operations. Additionally, we also present the utilization of graph structure learning, as depicted in Table \ref{tab0}. 
We compare our model with these RNN-based model: 1) GCGRU incorporates the GCN and RNN to capture temporal and dynamic features. 2) DCRNN applies a diffusion model to model the traffic diffusion process.
3) TGCN  combines the GCN and the gated recurrent unit.
4) AGCRN proposes a node and data adaptive graph convolution. 5) CLCRN is designed to capture the meteorology dynamics and use a conditional local convolution. 6) SGP uses a randomized recurrent neural network to embed the history of the input time series into high-dimensional state representations. 7) MGCRN uses a Meta Graph Learner to learn the heterogeneity in space and time. 8) RGSL obtains the prior graph information and the implict graph information with Regularized Graph Generation. In these alternating temporal and spatial layers models:
1) STGCN conducts a ST-Conv block to learn spatio-temporal dependencies; 2) MSTGCN captures daily, weekly and recently temporal dependencies with attention module; 3) ASTGCN uses a spatial temporal attention to predict traffic flow.
Moreover, We compare our model with Corrformer, a new global weather forecasting method with spatial multi-scale tree structure.
In models that necessitate the adjacency matrix as input, we construct the adjacency matrix according to distance and road connectivity between nodes. In meteorology datasets, we identify the 25 nodes with the shortest distances as neighboring nodes. In traffic datasets, the adjacency matrix is created based on road connectivity. Corrformer, which incorporates location information, leverages the latitude and longitude of nodes for embedding location features. The RNN-based models are configured with 32 hidden units and 2 graph recurrent layers. For the time-space alternation methods, the spatial and convolutional channels are set to be 64, with 4 temporal and spatial layers, respectively. The Corrformer model maintains the same hyper-parameters as its original vision. 

\begin{table}[htpb]
\caption{Forecasting performance across various horizons on the METR-LA dataset.}
% \small
\centering
\begin{tabular}{@{\hskip0.2cm}c|ccc@{}}
\hline \hline
\multirow{2}{4cm}{\diagbox[width=4.0cm, height=0.67cm]{Methods}{Horizons \& Metrics}}  & 3    & 6    & 12 \\
                                &MAE    &MAE    &MAE                      \\ \midrule
STGCN \cite{yu2017spatio}      & 2.88   & 3.47   & 4.59   \\
DCRNN \cite{li2018diffusion}     & 2.77   & 3.15   & 3.60   \\
AGCRN \cite{c:9}                  & 2.86   & 3.25   & 3.68   \\
MGCRN \cite{c:13}                   & \textbf{2.52}   & \textbf{2.93}   & 3.38   \\
SGP \cite{cini2023scalable}          & 2.69   & 3.05   & 3.45   \\
Ours                   & 2.97       & 2.99       & \textbf{3.06} \\      \hline \hline
\end{tabular}
\label{tab-ml}
\end{table}
\begin{table*}[htpb]
\caption{Forecasting performance across various horizons on the meteorological datasets.}
\centering
% \small
\begin{tabular}{@{\hskip0.2cm}c|ccc|ccc|ccc|ccc@{}}%\hskip0.2cm
\hline \hline
Datasets             & \multicolumn{3}{c|}{Wind} &\multicolumn{3}{c|}{Temperature} & \multicolumn{3}{c|}{Humidity} & \multicolumn{3}{c}{Cloud cover} \\ \midrule
\multirow{2}{*}{\diagbox[width=4.1cm, height=0.67cm]{Methods}{{Horizons \& Metrics}}} & 3      & 6     & 12   & 3      & 6     & 12    & 3    & 6    & 12    & 3     & 6     & 12     \\

                     & MAE       & MAE      & MAE    & MAE       & MAE      & MAE       & MAE      & MAE     & MAE      & MAE       & MAE      & MAE       \\ \midrule
CLCRN \cite{c:6}       & 0.73         & 1.23         & 1.81        & 0.78      & 1.45     & 2.29      & 0.29     & 0.49    & 0.65     & 1.21      & 1.60     & 1.89      \\
MGCRN \cite{c:13}      & 0.81     & 1.43    & 2.18         & 0.61      & 1.11     & 1.48      & 0.29     & 0.47    & 0.65     & 1.25      & 1.69     & 2.07      \\
RGSL \cite{yu2022regularized}   & 0.75 & 1.24  & 1.83             & 0.50      & 0.78     & 1.12      & 0.27     & 0.44    & 0.59     & 1.19      & 1.52     & 1.83      \\
Corrformer \cite{c:2}       & 0.81 & 1.33 & 2.02     & 0.52      & 0.85     & 1.16      & 0.31     & 0.49    & 0.65     & 1.26      & 1.67     & 2.02      \\
Ours               & \textbf{0.60} & \textbf{0.92} & \textbf{1.35}   & \textbf{0.41}      & \textbf{0.65}     & \textbf{0.93}      & \textbf{0.25}     & \textbf{0.40}    & \textbf{0.53}     & \textbf{1.15}      & \textbf{1.49}     & \textbf{1.78}      \\ \hline \hline
\end{tabular}
\label{tab-wind}
\end{table*}

\begin{table*}[htbp]
\caption{Comonent-wise analysis and ablation study on meteorology and traffic datasets.}
\centering
% \small
\begin{tabular}{@{\hskip0.2cm}c|cccccc@{}}
\hline \hline
\multirow{2}{4cm}{\diagbox[width=4.0cm, height=0.67cm]{\hskip0.3cm{Ablation}}{\hskip0.3cm{Datasets \& Metrics}\hskip0.1cm}} & Wind               & Temperature        & Cloud cover        & Humidity           & PeMSD4               & PeMSD8               \\
                          & MAE/RMSE           & MAE/RMSE           & MAE/RMSE           & MAE/RMSE           & MAE/RMSE             & MAE/RMSE             \\ \midrule
w/o MG-G                   & 1.01/1.60          & 0.80/1.37          & 1.66/2.74          & 0.40/0.61          & 31.60/48.51          & 25.20/38.68                \\
w/o MG-L & 1.19/2.01         & 0.70/1.22          &  1.50/2.44            & 0.42/0.65          &19.70/31.60                & 15.31/24.66              \\
w/o SA                    & 0.96/1.51          & 0.65/1.12          & 1.44/2.36          & 0.39/0.61          & 19.22/30.77          & 15.43/25.09          \\
w/o SA-GFC (MST)           & 1.38/2.03         & 1.01/1.60          & 1.72/2.69          & 0.52/0.75          &  21.58/33.75         &  17.60/27.89              \\
w/o SA-GFC (RNN)           &1.23/1.97          &0.96/1.61        & 1.54/2.51         & 0.63/0.64          &  20.58/32.14        & 16.12/25.33                     \\
Ours                      & \textbf{0.90/1.41} & \textbf{0.63/1.08} & \textbf{1.41/2.32} & \textbf{0.38/0.58} & \textbf{19.21/30.76} & \textbf{15.29/24.57} \\ \hline \hline
\end{tabular}
\label{tab2}
\end{table*}

\subsection{Experimental Results}
Table \ref{tab1} displays the evaluation results of our SAMSGL method, highlighting its superiority over all existing state-of-the-art AI-based approaches on meteorology datasets and comparable performance on traffic datasets. 
To ensure the reliability of our experiments, we carried out five repetitions with distinct random seeds (2021, 2022, 2023, 2024, 2025), following the procedures outlined in CLCRN \cite{c:6}, and present the mean and standard deviation of the experimental results. 
Notably, our SAMSGL achieves a remarkable improvement of up to 27.25\% in the RMSE metric for the wind dataset. Moreover, our SAMSGL outperforms other methods in all metrics in meteorology datasets. On the traffic datasets, our method show comparable performance.
Among the baseline models, RGSL and Corrformer exhibit robust performance in meteorology datasets, while MGCRN and RGSL surpass other baseline models in traffic forecasting tasks. RGSL and MGCRN derive benefits from learning the latent graph structure, whereas Corrformer demonstrates the efficiency of series alignment for considering the time delay of nodes. Our findings from Table \ref{tab1} indicate that although CLCRN performs admirably in meteorology datasets, it faces challenges in predicting future dynamics in traffic datasets. This discrepancy can be attributed to the conditional graph convolution within CLCRN, which aims to ensure node smoothness and computes attention scores based on location similarity. These characteristics closely mimic meteorological propagation but differ significantly from the road connectivity inherent in traffic forecasting. 
% In comparison of the performance of spatio-temporal architectures, RNN-based methods perform better than spatio-temporal attention blocks in general. 

% This also showcases the efficiency of graph structure learning and series alignment for considering the time delay of nodes.

To showcase the forecasting performance concerning spatial patterns, we compare the MAE and the predicted spatial patterns of various methods in Fig. \ref{fig_0} and Fig. \ref{spatial}, respectively. As is shown in Fig. \ref{fig_0}, the prediction errors are mainly concentrated in specific areas, where our model exhibits less deviation from the ground truth. The zoomed-out view in Fig. \ref{spatial}\subref{g_b} closely resembles the ground truth in Fig. \ref{spatial}\subref{g_a}, whereas other methods, despite having low forecasting errors according to the metrics, struggle to replicate the spatial patterns.

In Fig. \ref{ts_6}, we present the predicted temporal dynamics by MGCRN, RGSL, and our proposed SAMSGL. These methods can capture temporal evolution trends effectively but tend to perform less effectively in dealing with sudden, extreme changes over time. To further assess the methods' ability to predict temporal changes, we highlight specific examples in Fig. \ref{wind_global} to present a zoomed-out view of the predicted series. Fig. \ref{wind_global} provides a detailed visualization of the time delays between the ground truth and predicted results. A comparative analysis against other methods underscores our model's superior performance in forecasting both the overall trend and the fluctuations within the time series. Our predictions exhibit a higher level of accuracy compared to alternative methods. Remarkably, specific attention should be drawn to the segments highlighted by purple boxes. Here, our method excels by promptly adjusting to changes in trend within the time series, displaying minimal delays. In contrast, other methods might demonstrate discrepancies in values or encounter delays in forecasting the evolving trend. This visual comparison further solidifies the efficacy of our approach in capturing intricate temporal dynamics and making more accurate predictions.
%width=0.25\textwidth
% \begin{table}[!h]
% \caption{Forecasting performance across various horizons on the Wind dataset.}
% \small
% \centering
% \begin{tabular}{@{\hskip0.2cm}c|ccc@{}}
% \toprule
% \multirow{2}{*}{\diagbox[width=3.9cm, height=0.75cm]{Methods}{\hskip0.3cm{Horizons \& Metrics}}} & 3    & 6    & 12   \\
%                       & MAE      & MAE      & MAE      \\ \midrule
% CLCRN \cite{c:6}      & 0.73         & 1.23         & 1.81         \\
% MGCRN  \cite{c:13}               & 0.81  & 1.43 & 2.18 \\
% RGSL  \cite{yu2022regularized}                & 0.75 & 1.24  & 1.83 \\
% Corrformer \cite{c:2}           & 0.81 & 1.33 & 2.02 \\
% Ours                & \textbf{0.60} & \textbf{0.92} & \textbf{1.35} \\ \bottomrule
% \end{tabular}
% \label{tab-wind}
% \end{table}
To evaluate the forecasting capabilities of different methods across different horizons, we analyze the MAE of these methods with 3-rd, 6-th, and 12-th horizons. The forecasting error increase with the increase of time steps.
As shown in Table \ref{tab-ml}, our method exhibits stable forecasting performance in the Metr-LA dataset, with minimal variations in prediction errors across different time steps. Notably, MGCRN and SGP outperform in the 3rd and 6th predictions, which can be attributed to their ability to capture local short-term dynamics through their recurrent structures. In contrast, our model simultaneously predicts results for all time steps.
% As shown in Table \ref{tab-ml}, our method exhibits stable forecasting performance in the Metr-LA dataset, with minimal variations in prediction errors across different time steps. 
Additionally, in Table \ref{tab-wind}, our method consistently outperforms the competition at various time steps on meteorological datasets.  

To illustrate the function of multi-scale graph structure learning and distinguish between our series-aligned graph convolution and the conventional spatial graph convolution, we provide visualizations of the outputs from the series-aligned graph convolution, delayed graph convolution, local graph convolution, and the fusion of these outputs within the multi-graph convolution block. We focus on the outputs from the first Graph-FC block (where $m=1$). We display the feature patterns of a single node in $H_{\rm nd}$, $H_{\rm d}$, $H_{\rm l}$, and $H_0^{1}$, all of which have dimensions of $\mathbb{R}^{{L}\times{D}}$. As depicted in Fig. \ref{ms}, we observe distinctions in the outputs of series-aligned graph convolution (Fig. \ref{ms}\subref{ms-a}), delayed graph convolution (Fig. \ref{ms}\subref{ms-b}), and local graph convolution (Fig. \ref{ms}\subref{ms-c}). Notably, the output of series-aligned graph convolution exhibits variations over time steps, while the outputs of delayed graph convolution and local graph convolution remain consistent at each time step and only vary with changes in the number of features. This indicates that the series alignment process aggregates messages from neighboring nodes at different time steps based on the time delays between them, capturing spatial-temporal dependencies. Additionally, the outputs of delayed graph convolution and local graph convolution differ, suggesting variations in feature extraction between global and local graphs. This phenomenon highlights that these graph convolution methods learn spatial representations at multiple scales, underscoring the utility of our multi-scale graph learning approach.\par
Additionally, we compare the performance of selecting different nodes as the reference node for alignment. As shown in Table \ref{ref_nod}, utilizing the average series of all nodes as the reference node yields better performance, while randomly selecting a node lacks robustness.
\begin{table}[h]
\caption{Forecasting performance of various reference node selection strategies.}
\begin{tabular}{c|cc}
\hline \hline
\multirow{2}{*}{Reference node} & \multicolumn{2}{c}{Wind} \\
  & MAE        & RMSE        \\ \hline
node 1         & 0.92       &  1.43        \\
node 100       & 0.91        & 1.42         \\
node 1000      & 0.91        & 1.42       \\ 
mean node      & 0.90       & 1.40 \\ \hline \hline
\end{tabular}
\label{ref_nod}

\end{table}
\subsection{Ablation Studies}
To comprehensively assess the efficiency of each component, we conduct ablation studies on both meteorology and traffic datasets. Our evaluation focuses on isolating and analyzing the contributions of specific modules within our SAMSGL model, including the Series-Alignment Graph Convolution (SA), the Multi-Scale Graph Structure Learning module (MG), and the Graph-FC blocks (GFC). The following strategies are employed for comparison:
\begin{itemize}
\item \textbf{w/o MG-G}: The SAMSGL model is examined without the global graphs. Spatial interactions are established solely based on node locations and the distances between them.
\item \textbf{w/o MG-L}: The SAMSGL model is examined without local graph. Only global graphs learned from node embedding are utilized to represent the spatial relations. 
\item \textbf{w/o SA}: The SAMSGL model is examined without the SA. Spatial information is aggregated using traditional spatial GCN, while the model still retains the MG and the GFC.
\item \textbf{w/o SA-GFC (MST)}: The SA and the GFC are both omitted and substituted with the Multi-Component Spatial-Temporal Graph Convolution block (MST) \cite{c:15}.
\item \textbf{w/o SA-GFC (RNN)}: The SA and the GFC are both omitted and substituted with a graph RNN.
\end{itemize}

% It is worth noting that spatio-temporal architectures like time convolution layers hinder the process of finding the initial points of series, while RNN-based methods serve the node feature at each time step as hidden states, rendering alignment unnecessary. Based on the above analysis and unique characteristics of SA, we choose to exclude both the SA and the GFC, instead of solely replacing GFC with other spatio-temporal modules. 
We don't solely choose to replace GFC with other spatio-temporal modules due to the unique characteristics of SA. Spatio-temporal architectures like time convolution layers fusing temporal information, which obstacles the process of finding the initial points of series, while RNN-based methods serve the node feature at each time step as hidden states, rendering alignment unnecessary.
% SA focuses on aligning time series, while the spatio-temporal architectures with time convolution layers aggregates comprehensive temporal information, capturing features over a specific time span. Combining these temporal fusion capabilities with series alignment would introduce conflicting objectives. Additionally, the hidden states within the RNN already serve as node features at each time step, rendering alignment unnecessary. 
Based on the above analysis and unique characteristics of SA, we choose to exclude both the SA and the GFC, instead of solely replacing GFC with other spatio-temporal modules.

As depicted in Fig. \ref{abl_all}, the model with all components consistently delivers a strong performance at every forecasting time step onvarous datasets. Table \ref{tab2} clearly illustrates a significant degradation in prediction accuracy when substituting the GFC with other spatio-temporal structures. Both local and global graphs play a crucial role in influencing prediction accuracy, underscoring the effectiveness of learning node relations across multiple scales. Additionally, the SA module contributes to maintaining high accuracy. All components demonstrate efficiency across diverse forecasting tasks, indicating the generalizability of these modules.
Overall, this analysis reaffirms the collective importance of these components in enabling our model to consistently deliver reliable forecasts.

\section{Conclusion}\label{sec:section5}
We propose a spatio-temporal forecasting model, called SAMSGL, which consists of three key components: the Series-Aligned Graph Convolution, the Multi-Scale Graph Structure Learning, and the Graph-FC Blocks. 
The Series-Aligned Graph Convolution effectively reduces the influence of time delays, leading to improved spatial aggregation of node features. The multi-scale graphs learned node interactions from scales of the local graph, the global delayed graph, and the global non-delayed graph, in combination with Graph-FC blocks, aid high-dimensional interaction comprehension and spatio-temporal dynamics extraction. Our experimental evaluations on meteorology and traffic flow datasets produce compelling quantitative and visual results, consistently demonstrating the superiority of our proposed method over existing state-of-the-art approaches. Considering the computational cost of conducting graph convolution at muiltiple graphs, we will further improve the the structure of our network for the reduction of computation. Furthermore, within our SAMSGL model, we currently utilize a linear layer to combine features from multi-scale graphs. Future efforts will focus on refining graph fusion methods for improved performance and reduction of computational cost.

\begin{acknowledgments}
This work was supported by the National Natural Science Foundation of China (62293502, 62293504, 62173147).

\end{acknowledgments}
\section*{Data Availability Statement}
The WeatherBench data employed in this paper are available from the website \url{https://github.com/pangeo-data/WeatherBench}. The PeMSD4 and PeMSD8 datasets are available from the website \url{https://github.com/ant-research/RGSL/tree/main/data}. The METR-LA dataset is available from the website \url{https://github.com/deepkashiwa20/MegaCRN/tree/main/METRLA}
% \nocite{*}
\bibliography{main}% Produces the bibliography via BibTeX.

\begin{thebibliography}{54}%
\makeatletter
\providecommand \@ifxundefined [1]{%
 \@ifx{#1\undefined}
}%
\providecommand \@ifnum [1]{%
 \ifnum #1\expandafter \@firstoftwo
 \else \expandafter \@secondoftwo
 \fi
}%
\providecommand \@ifx [1]{%
 \ifx #1\expandafter \@firstoftwo
 \else \expandafter \@secondoftwo
 \fi
}%
\providecommand \natexlab [1]{#1}%
\providecommand \enquote  [1]{``#1''}%
\providecommand \bibnamefont  [1]{#1}%
\providecommand \bibfnamefont [1]{#1}%
\providecommand \citenamefont [1]{#1}%
\providecommand \href@noop [0]{\@secondoftwo}%
\providecommand \href [0]{\begingroup \@sanitize@url \@href}%
\providecommand \@href[1]{\@@startlink{#1}\@@href}%
\providecommand \@@href[1]{\endgroup#1\@@endlink}%
\providecommand \@sanitize@url [0]{\catcode `\\12\catcode `\$12\catcode `\&12\catcode `\#12\catcode `\^12\catcode `\_12\catcode `\%12\relax}%
\providecommand \@@startlink[1]{}%
\providecommand \@@endlink[0]{}%
\providecommand \url  [0]{\begingroup\@sanitize@url \@url }%
\providecommand \@url [1]{\endgroup\@href {#1}{\urlprefix }}%
\providecommand \urlprefix  [0]{URL }%
\providecommand \Eprint [0]{\href }%
\providecommand \doibase [0]{http://dx.doi.org/}%
\providecommand \selectlanguage [0]{\@gobble}%
\providecommand \bibinfo  [0]{\@secondoftwo}%
\providecommand \bibfield  [0]{\@secondoftwo}%
\providecommand \translation [1]{[#1]}%
\providecommand \BibitemOpen [0]{}%
\providecommand \bibitemStop [0]{}%
\providecommand \bibitemNoStop [0]{.\EOS\space}%
\providecommand \EOS [0]{\spacefactor3000\relax}%
\providecommand \BibitemShut  [1]{\csname bibitem#1\endcsname}%
\let\auto@bib@innerbib\@empty
%</preamble>
\bibitem [{\citenamefont {Liu}\ \emph {et~al.}(2024)\citenamefont {Liu}, \citenamefont {Zhao}, \citenamefont {Sun},\ and\ \citenamefont {Zhou}}]{chaotic1}%
  \BibitemOpen
  \bibfield  {author} {\bibinfo {author} {\bibfnamefont {T.}~\bibnamefont {Liu}}, \bibinfo {author} {\bibfnamefont {X.}~\bibnamefont {Zhao}}, \bibinfo {author} {\bibfnamefont {P.}~\bibnamefont {Sun}}, \ and\ \bibinfo {author} {\bibfnamefont {J.}~\bibnamefont {Zhou}},\ }\bibfield  {title} {\enquote {\bibinfo {title} {{A hybrid proper orthogonal decomposition and next generation reservoir computing approach for high-dimensional chaotic prediction: Application to flow-induced vibration of tube bundles}},}\ }\href@noop {} {\bibfield  {journal} {\bibinfo  {journal} {Chaos: An Interdisciplinary Journal of Nonlinear Science}\ }\textbf {\bibinfo {volume} {34}},\ \bibinfo {pages} {033125} (\bibinfo {year} {2024})}\BibitemShut {NoStop}%
\bibitem [{\citenamefont {Wang}\ \emph {et~al.}(2024)\citenamefont {Wang}, \citenamefont {Feng}, \citenamefont {Xu},\ and\ \citenamefont {Kurths}}]{chaotic2}%
  \BibitemOpen
  \bibfield  {author} {\bibinfo {author} {\bibfnamefont {X.}~\bibnamefont {Wang}}, \bibinfo {author} {\bibfnamefont {J.}~\bibnamefont {Feng}}, \bibinfo {author} {\bibfnamefont {Y.}~\bibnamefont {Xu}}, \ and\ \bibinfo {author} {\bibfnamefont {J.}~\bibnamefont {Kurths}},\ }\bibfield  {title} {\enquote {\bibinfo {title} {{Deep learning-based state prediction of the Lorenz system with control parameters}},}\ }\href@noop {} {\bibfield  {journal} {\bibinfo  {journal} {Chaos: An Interdisciplinary Journal of Nonlinear Science}\ }\textbf {\bibinfo {volume} {34}},\ \bibinfo {pages} {033108} (\bibinfo {year} {2024})}\BibitemShut {NoStop}%
\bibitem [{\citenamefont {Aksoy}(2024)}]{chaotic3}%
  \BibitemOpen
  \bibfield  {author} {\bibinfo {author} {\bibfnamefont {A.}~\bibnamefont {Aksoy}},\ }\bibfield  {title} {\enquote {\bibinfo {title} {{A Monte Carlo approach to understanding the impacts of initial-condition uncertainty, model uncertainty, and simulation variability on the predictability of chaotic systems: Perspectives from the one-dimensional logistic map}},}\ }\href@noop {} {\bibfield  {journal} {\bibinfo  {journal} {Chaos: An Interdisciplinary Journal of Nonlinear Science}\ }\textbf {\bibinfo {volume} {34}},\ \bibinfo {pages} {011102} (\bibinfo {year} {2024})}\BibitemShut {NoStop}%
\bibitem [{\citenamefont {Kwon}, \citenamefont {Jung},\ and\ \citenamefont {Eom}(2023)}]{traffic1}%
  \BibitemOpen
  \bibfield  {author} {\bibinfo {author} {\bibfnamefont {Y.}~\bibnamefont {Kwon}}, \bibinfo {author} {\bibfnamefont {J.-H.}\ \bibnamefont {Jung}}, \ and\ \bibinfo {author} {\bibfnamefont {Y.-H.}\ \bibnamefont {Eom}},\ }\bibfield  {title} {\enquote {\bibinfo {title} {{Global efficiency and network structure of urban traffic flows: A percolation-based empirical analysis}},}\ }\href@noop {} {\bibfield  {journal} {\bibinfo  {journal} {Chaos: An Interdisciplinary Journal of Nonlinear Science}\ }\textbf {\bibinfo {volume} {33}},\ \bibinfo {pages} {113104} (\bibinfo {year} {2023})}\BibitemShut {NoStop}%
\bibitem [{\citenamefont {Wu}\ \emph {et~al.}(2023)\citenamefont {Wu}, \citenamefont {Zhou}, \citenamefont {Long},\ and\ \citenamefont {Wang}}]{c:2}%
  \BibitemOpen
  \bibfield  {author} {\bibinfo {author} {\bibfnamefont {H.}~\bibnamefont {Wu}}, \bibinfo {author} {\bibfnamefont {H.}~\bibnamefont {Zhou}}, \bibinfo {author} {\bibfnamefont {M.}~\bibnamefont {Long}}, \ and\ \bibinfo {author} {\bibfnamefont {J.}~\bibnamefont {Wang}},\ }\bibfield  {title} {\enquote {\bibinfo {title} {Interpretable weather forecasting for worldwide stations with a unified deep model},}\ }\href@noop {} {\bibfield  {journal} {\bibinfo  {journal} {Nature Machine Intelligence}\ }\textbf {\bibinfo {volume} {5}},\ \bibinfo {pages} {602--611} (\bibinfo {year} {2023})}\BibitemShut {NoStop}%
\bibitem [{\citenamefont {Cheng}, \citenamefont {Guo},\ and\ \citenamefont {Arcucci}(2023)}]{cheng2023generative}%
  \BibitemOpen
  \bibfield  {author} {\bibinfo {author} {\bibfnamefont {S.}~\bibnamefont {Cheng}}, \bibinfo {author} {\bibfnamefont {Y.}~\bibnamefont {Guo}}, \ and\ \bibinfo {author} {\bibfnamefont {R.}~\bibnamefont {Arcucci}},\ }\bibfield  {title} {\enquote {\bibinfo {title} {A generative model for surrogates of spatial-temporal wildfire nowcasting},}\ }\href@noop {} {\bibfield  {journal} {\bibinfo  {journal} {IEEE Transactions on Emerging Topics in Computational Intelligence}\ }\textbf {\bibinfo {volume} {7}},\ \bibinfo {pages} {1420--1430} (\bibinfo {year} {2023})}\BibitemShut {NoStop}%
\bibitem [{\citenamefont {Titz}\ \emph {et~al.}(2024)\citenamefont {Titz}, \citenamefont {Kaiser}, \citenamefont {Kruse},\ and\ \citenamefont {Witthaut}}]{power1}%
  \BibitemOpen
  \bibfield  {author} {\bibinfo {author} {\bibfnamefont {M.}~\bibnamefont {Titz}}, \bibinfo {author} {\bibfnamefont {F.}~\bibnamefont {Kaiser}}, \bibinfo {author} {\bibfnamefont {J.}~\bibnamefont {Kruse}}, \ and\ \bibinfo {author} {\bibfnamefont {D.}~\bibnamefont {Witthaut}},\ }\bibfield  {title} {\enquote {\bibinfo {title} {{Predicting dynamic stability from static features in power grid models using machine learning}},}\ }\href@noop {} {\bibfield  {journal} {\bibinfo  {journal} {Chaos: An Interdisciplinary Journal of Nonlinear Science}\ }\textbf {\bibinfo {volume} {34}},\ \bibinfo {pages} {013139} (\bibinfo {year} {2024})}\BibitemShut {NoStop}%
\bibitem [{\citenamefont {Xiong}\ \emph {et~al.}(2022)\citenamefont {Xiong}, \citenamefont {Tang}, \citenamefont {Mao}, \citenamefont {Liu}, \citenamefont {Meng}, \citenamefont {Dong},\ and\ \citenamefont {Qian}}]{xiong2022two}%
  \BibitemOpen
  \bibfield  {author} {\bibinfo {author} {\bibfnamefont {L.}~\bibnamefont {Xiong}}, \bibinfo {author} {\bibfnamefont {Y.}~\bibnamefont {Tang}}, \bibinfo {author} {\bibfnamefont {S.}~\bibnamefont {Mao}}, \bibinfo {author} {\bibfnamefont {H.}~\bibnamefont {Liu}}, \bibinfo {author} {\bibfnamefont {K.}~\bibnamefont {Meng}}, \bibinfo {author} {\bibfnamefont {Z.}~\bibnamefont {Dong}}, \ and\ \bibinfo {author} {\bibfnamefont {F.}~\bibnamefont {Qian}},\ }\bibfield  {title} {\enquote {\bibinfo {title} {A two-level energy management strategy for multi-microgrid systems with interval prediction and reinforcement learning},}\ }\href@noop {} {\bibfield  {journal} {\bibinfo  {journal} {IEEE Transactions on Circuits and Systems I: Regular Papers}\ }\textbf {\bibinfo {volume} {69}},\ \bibinfo {pages} {1788--1799} (\bibinfo {year} {2022})}\BibitemShut {NoStop}%
\bibitem [{\citenamefont {Macioszek}\ and\ \citenamefont {Kurek}(2021)}]{macioszek2021road}%
  \BibitemOpen
  \bibfield  {author} {\bibinfo {author} {\bibfnamefont {E.}~\bibnamefont {Macioszek}}\ and\ \bibinfo {author} {\bibfnamefont {A.}~\bibnamefont {Kurek}},\ }\bibfield  {title} {\enquote {\bibinfo {title} {Road traffic distribution on public holidays and workdays on selected road transport network elements},}\ }\href@noop {} {\bibfield  {journal} {\bibinfo  {journal} {Transport Problems}\ }\textbf {\bibinfo {volume} {16}} (\bibinfo {year} {2021})}\BibitemShut {NoStop}%
\bibitem [{\citenamefont {Boers}\ \emph {et~al.}(2019)\citenamefont {Boers}, \citenamefont {Goswami}, \citenamefont {Rheinwalt}, \citenamefont {Bookhagen}, \citenamefont {Hoskins},\ and\ \citenamefont {Kurths}}]{boers2019complex}%
  \BibitemOpen
  \bibfield  {author} {\bibinfo {author} {\bibfnamefont {N.}~\bibnamefont {Boers}}, \bibinfo {author} {\bibfnamefont {B.}~\bibnamefont {Goswami}}, \bibinfo {author} {\bibfnamefont {A.}~\bibnamefont {Rheinwalt}}, \bibinfo {author} {\bibfnamefont {B.}~\bibnamefont {Bookhagen}}, \bibinfo {author} {\bibfnamefont {B.}~\bibnamefont {Hoskins}}, \ and\ \bibinfo {author} {\bibfnamefont {J.}~\bibnamefont {Kurths}},\ }\bibfield  {title} {\enquote {\bibinfo {title} {Complex networks reveal global pattern of extreme-rainfall teleconnections},}\ }\href@noop {} {\bibfield  {journal} {\bibinfo  {journal} {Nature}\ }\textbf {\bibinfo {volume} {566}},\ \bibinfo {pages} {373--377} (\bibinfo {year} {2019})}\BibitemShut {NoStop}%
\bibitem [{\citenamefont {Tang}\ \emph {et~al.}(2020)\citenamefont {Tang}, \citenamefont {Kurths}, \citenamefont {Lin}, \citenamefont {Ott},\ and\ \citenamefont {Kocarev}}]{tang2020introduction}%
  \BibitemOpen
  \bibfield  {author} {\bibinfo {author} {\bibfnamefont {Y.}~\bibnamefont {Tang}}, \bibinfo {author} {\bibfnamefont {J.}~\bibnamefont {Kurths}}, \bibinfo {author} {\bibfnamefont {W.}~\bibnamefont {Lin}}, \bibinfo {author} {\bibfnamefont {E.}~\bibnamefont {Ott}}, \ and\ \bibinfo {author} {\bibfnamefont {L.}~\bibnamefont {Kocarev}},\ }\bibfield  {title} {\enquote {\bibinfo {title} {{Introduction to Focus Issue: When machine learning meets complex systems: Networks, chaos, and nonlinear dynamics}},}\ }\href@noop {} {\bibfield  {journal} {\bibinfo  {journal} {Chaos: An Interdisciplinary Journal of Nonlinear Science}\ }\textbf {\bibinfo {volume} {30}},\ \bibinfo {pages} {063151} (\bibinfo {year} {2020})}\BibitemShut {NoStop}%
\bibitem [{\citenamefont {Ji}\ \emph {et~al.}(2023)\citenamefont {Ji}, \citenamefont {Ye}, \citenamefont {Mu}, \citenamefont {Lin}, \citenamefont {Tian}, \citenamefont {Hens}, \citenamefont {Perc}, \citenamefont {Tang}, \citenamefont {Sun},\ and\ \citenamefont {Kurths}}]{ji2023signal}%
  \BibitemOpen
  \bibfield  {author} {\bibinfo {author} {\bibfnamefont {P.}~\bibnamefont {Ji}}, \bibinfo {author} {\bibfnamefont {J.}~\bibnamefont {Ye}}, \bibinfo {author} {\bibfnamefont {Y.}~\bibnamefont {Mu}}, \bibinfo {author} {\bibfnamefont {W.}~\bibnamefont {Lin}}, \bibinfo {author} {\bibfnamefont {Y.}~\bibnamefont {Tian}}, \bibinfo {author} {\bibfnamefont {C.}~\bibnamefont {Hens}}, \bibinfo {author} {\bibfnamefont {M.}~\bibnamefont {Perc}}, \bibinfo {author} {\bibfnamefont {Y.}~\bibnamefont {Tang}}, \bibinfo {author} {\bibfnamefont {J.}~\bibnamefont {Sun}}, \ and\ \bibinfo {author} {\bibfnamefont {J.}~\bibnamefont {Kurths}},\ }\bibfield  {title} {\enquote {\bibinfo {title} {Signal propagation in complex networks},}\ }\href@noop {} {\bibfield  {journal} {\bibinfo  {journal} {Physics Reports}\ }\textbf {\bibinfo {volume} {1017}},\ \bibinfo {pages} {1--96} (\bibinfo {year} {2023})}\BibitemShut {NoStop}%
\bibitem [{\citenamefont {Chen}\ \emph {et~al.}(2001)\citenamefont {Chen}, \citenamefont {Petty}, \citenamefont {Skabardonis}, \citenamefont {Varaiya},\ and\ \citenamefont {Jia}}]{chen2001freeway}%
  \BibitemOpen
  \bibfield  {author} {\bibinfo {author} {\bibfnamefont {C.}~\bibnamefont {Chen}}, \bibinfo {author} {\bibfnamefont {K.}~\bibnamefont {Petty}}, \bibinfo {author} {\bibfnamefont {A.}~\bibnamefont {Skabardonis}}, \bibinfo {author} {\bibfnamefont {P.}~\bibnamefont {Varaiya}}, \ and\ \bibinfo {author} {\bibfnamefont {Z.}~\bibnamefont {Jia}},\ }\bibfield  {title} {\enquote {\bibinfo {title} {Freeway performance measurement system: mining loop detector data},}\ }\href@noop {} {\bibfield  {journal} {\bibinfo  {journal} {Transportation Research Record}\ }\textbf {\bibinfo {volume} {1748}},\ \bibinfo {pages} {96--102} (\bibinfo {year} {2001})}\BibitemShut {NoStop}%
\bibitem [{\citenamefont {Lam}\ \emph {et~al.}(2023)\citenamefont {Lam}, \citenamefont {Sanchez-Gonzalez}, \citenamefont {Willson}, \citenamefont {Wirnsberger}, \citenamefont {Fortunato}, \citenamefont {Alet}, \citenamefont {Ravuri}, \citenamefont {Ewalds}, \citenamefont {Eaton-Rosen}, \citenamefont {Hu}, \citenamefont {Merose}, \citenamefont {Hoyer}, \citenamefont {Holland}, \citenamefont {Vinyals}, \citenamefont {Stott}, \citenamefont {Pritzel}, \citenamefont {Mohamed},\ and\ \citenamefont {Battaglia}}]{c:5}%
  \BibitemOpen
  \bibfield  {author} {\bibinfo {author} {\bibfnamefont {R.}~\bibnamefont {Lam}}, \bibinfo {author} {\bibfnamefont {A.}~\bibnamefont {Sanchez-Gonzalez}}, \bibinfo {author} {\bibfnamefont {M.}~\bibnamefont {Willson}}, \bibinfo {author} {\bibfnamefont {P.}~\bibnamefont {Wirnsberger}}, \bibinfo {author} {\bibfnamefont {M.}~\bibnamefont {Fortunato}}, \bibinfo {author} {\bibfnamefont {F.}~\bibnamefont {Alet}}, \bibinfo {author} {\bibfnamefont {S.}~\bibnamefont {Ravuri}}, \bibinfo {author} {\bibfnamefont {T.}~\bibnamefont {Ewalds}}, \bibinfo {author} {\bibfnamefont {Z.}~\bibnamefont {Eaton-Rosen}}, \bibinfo {author} {\bibfnamefont {W.}~\bibnamefont {Hu}}, \bibinfo {author} {\bibfnamefont {A.}~\bibnamefont {Merose}}, \bibinfo {author} {\bibfnamefont {S.}~\bibnamefont {Hoyer}}, \bibinfo {author} {\bibfnamefont {G.}~\bibnamefont {Holland}}, \bibinfo {author} {\bibfnamefont {O.}~\bibnamefont {Vinyals}}, \bibinfo {author} {\bibfnamefont {J.}~\bibnamefont {Stott}}, \bibinfo {author} {\bibfnamefont {A.}~\bibnamefont
  {Pritzel}}, \bibinfo {author} {\bibfnamefont {S.}~\bibnamefont {Mohamed}}, \ and\ \bibinfo {author} {\bibfnamefont {P.}~\bibnamefont {Battaglia}},\ }\bibfield  {title} {\enquote {\bibinfo {title} {Learning skillful medium-range global weather forecasting},}\ }\href@noop {} {\bibfield  {journal} {\bibinfo  {journal} {Science}\ }\textbf {\bibinfo {volume} {382}},\ \bibinfo {pages} {1416--1421} (\bibinfo {year} {2023})}\BibitemShut {NoStop}%
\bibitem [{\citenamefont {Kipf}\ and\ \citenamefont {Welling}(2017)}]{kipf2016semi}%
  \BibitemOpen
  \bibfield  {author} {\bibinfo {author} {\bibfnamefont {T.~N.}\ \bibnamefont {Kipf}}\ and\ \bibinfo {author} {\bibfnamefont {M.}~\bibnamefont {Welling}},\ }\bibfield  {title} {\enquote {\bibinfo {title} {Semi-supervised classification with graph convolutional networks},}\ }in\ \href@noop {} {\emph {\bibinfo {booktitle} {Proceedings of the 5th International Conference on Learning Representations}}}\ (\bibinfo {year} {2017})\BibitemShut {NoStop}%
\bibitem [{\citenamefont {Li}\ and\ \citenamefont {Zhu}(2021)}]{li2021spatial}%
  \BibitemOpen
  \bibfield  {author} {\bibinfo {author} {\bibfnamefont {M.}~\bibnamefont {Li}}\ and\ \bibinfo {author} {\bibfnamefont {Z.}~\bibnamefont {Zhu}},\ }\bibfield  {title} {\enquote {\bibinfo {title} {Spatial-temporal fusion graph neural networks for traffic flow forecasting},}\ }in\ \href@noop {} {\emph {\bibinfo {booktitle} {Proceedings of the AAAI Conference on Artificial Intelligence}}},\ Vol.~\bibinfo {volume} {35}\ (\bibinfo {year} {2021})\ pp.\ \bibinfo {pages} {4189--4196}\BibitemShut {NoStop}%
\bibitem [{\citenamefont {Yu}\ \emph {et~al.}(2022)\citenamefont {Yu}, \citenamefont {Li}, \citenamefont {Yu}, \citenamefont {Li}, \citenamefont {Huang}, \citenamefont {Wang},\ and\ \citenamefont {Liu}}]{yu2022regularized}%
  \BibitemOpen
  \bibfield  {author} {\bibinfo {author} {\bibfnamefont {H.}~\bibnamefont {Yu}}, \bibinfo {author} {\bibfnamefont {T.}~\bibnamefont {Li}}, \bibinfo {author} {\bibfnamefont {W.}~\bibnamefont {Yu}}, \bibinfo {author} {\bibfnamefont {J.}~\bibnamefont {Li}}, \bibinfo {author} {\bibfnamefont {Y.}~\bibnamefont {Huang}}, \bibinfo {author} {\bibfnamefont {L.}~\bibnamefont {Wang}}, \ and\ \bibinfo {author} {\bibfnamefont {A.}~\bibnamefont {Liu}},\ }\bibfield  {title} {\enquote {\bibinfo {title} {Regularized graph structure learning with semantic knowledge for multi-variates time-series forecasting},}\ }in\ \href@noop {} {\emph {\bibinfo {booktitle} {Proceedings of the Thirty-First International Joint Conference on Artificial Intelligence, {IJCAI-22}}}}\ (\bibinfo {year} {2022})\ pp.\ \bibinfo {pages} {2362--2368}\BibitemShut {NoStop}%
\bibitem [{\citenamefont {Shao}\ \emph {et~al.}(2022)\citenamefont {Shao}, \citenamefont {Jin}, \citenamefont {Wang}, \citenamefont {Kang}, \citenamefont {Xiao}, \citenamefont {Menouar}, \citenamefont {Zhang}, \citenamefont {Zhang},\ and\ \citenamefont {Salim}}]{c:14}%
  \BibitemOpen
  \bibfield  {author} {\bibinfo {author} {\bibfnamefont {W.}~\bibnamefont {Shao}}, \bibinfo {author} {\bibfnamefont {Z.}~\bibnamefont {Jin}}, \bibinfo {author} {\bibfnamefont {S.}~\bibnamefont {Wang}}, \bibinfo {author} {\bibfnamefont {Y.}~\bibnamefont {Kang}}, \bibinfo {author} {\bibfnamefont {X.}~\bibnamefont {Xiao}}, \bibinfo {author} {\bibfnamefont {H.}~\bibnamefont {Menouar}}, \bibinfo {author} {\bibfnamefont {Z.}~\bibnamefont {Zhang}}, \bibinfo {author} {\bibfnamefont {J.}~\bibnamefont {Zhang}}, \ and\ \bibinfo {author} {\bibfnamefont {F.}~\bibnamefont {Salim}},\ }\bibfield  {title} {\enquote {\bibinfo {title} {Long-term spatio-temporal forecasting via dynamic multiple-graph attention},}\ }in\ \href@noop {} {\emph {\bibinfo {booktitle} {Proceedings of the Thirty-First International Joint Conference on Artificial Intelligence, {IJCAI-22}}}}\ (\bibinfo {year} {2022})\ pp.\ \bibinfo {pages} {2225--2232}\BibitemShut {NoStop}%
\bibitem [{\citenamefont {Bai}\ \emph {et~al.}(2020)\citenamefont {Bai}, \citenamefont {Yao}, \citenamefont {Li}, \citenamefont {Wang},\ and\ \citenamefont {Wang}}]{c:9}%
  \BibitemOpen
  \bibfield  {author} {\bibinfo {author} {\bibfnamefont {L.}~\bibnamefont {Bai}}, \bibinfo {author} {\bibfnamefont {L.}~\bibnamefont {Yao}}, \bibinfo {author} {\bibfnamefont {C.}~\bibnamefont {Li}}, \bibinfo {author} {\bibfnamefont {X.}~\bibnamefont {Wang}}, \ and\ \bibinfo {author} {\bibfnamefont {C.}~\bibnamefont {Wang}},\ }\bibfield  {title} {\enquote {\bibinfo {title} {Adaptive graph convolutional recurrent network for traffic forecasting},}\ }in\ \href@noop {} {\emph {\bibinfo {booktitle} {Advances in Neural Information Processing Systems}}},\ Vol.~\bibinfo {volume} {33}\ (\bibinfo {year} {2020})\ pp.\ \bibinfo {pages} {17804--17815}\BibitemShut {NoStop}%
\bibitem [{\citenamefont {Jiang}\ \emph {et~al.}(2023{\natexlab{a}})\citenamefont {Jiang}, \citenamefont {Wang}, \citenamefont {Yong}, \citenamefont {Jeph}, \citenamefont {Chen}, \citenamefont {Kobayashi}, \citenamefont {Song}, \citenamefont {Fukushima},\ and\ \citenamefont {Suzumura}}]{c:13}%
  \BibitemOpen
  \bibfield  {author} {\bibinfo {author} {\bibfnamefont {R.}~\bibnamefont {Jiang}}, \bibinfo {author} {\bibfnamefont {Z.}~\bibnamefont {Wang}}, \bibinfo {author} {\bibfnamefont {J.}~\bibnamefont {Yong}}, \bibinfo {author} {\bibfnamefont {P.}~\bibnamefont {Jeph}}, \bibinfo {author} {\bibfnamefont {Q.}~\bibnamefont {Chen}}, \bibinfo {author} {\bibfnamefont {Y.}~\bibnamefont {Kobayashi}}, \bibinfo {author} {\bibfnamefont {X.}~\bibnamefont {Song}}, \bibinfo {author} {\bibfnamefont {S.}~\bibnamefont {Fukushima}}, \ and\ \bibinfo {author} {\bibfnamefont {T.}~\bibnamefont {Suzumura}},\ }\bibfield  {title} {\enquote {\bibinfo {title} {Spatio-temporal meta-graph learning for traffic forecasting},}\ }in\ \href@noop {} {\emph {\bibinfo {booktitle} {Proceedings of the AAAI Conference on Artificial Intelligence}}},\ Vol.~\bibinfo {volume} {37}\ (\bibinfo {year} {2023})\ pp.\ \bibinfo {pages} {8078--8086}\BibitemShut {NoStop}%
\bibitem [{\citenamefont {Guo}\ \emph {et~al.}(2019)\citenamefont {Guo}, \citenamefont {Lin}, \citenamefont {Feng}, \citenamefont {Song},\ and\ \citenamefont {Wan}}]{c:15}%
  \BibitemOpen
  \bibfield  {author} {\bibinfo {author} {\bibfnamefont {S.}~\bibnamefont {Guo}}, \bibinfo {author} {\bibfnamefont {Y.}~\bibnamefont {Lin}}, \bibinfo {author} {\bibfnamefont {N.}~\bibnamefont {Feng}}, \bibinfo {author} {\bibfnamefont {C.}~\bibnamefont {Song}}, \ and\ \bibinfo {author} {\bibfnamefont {H.}~\bibnamefont {Wan}},\ }\bibfield  {title} {\enquote {\bibinfo {title} {Attention based spatial-temporal graph convolutional networks for traffic flow forecasting},}\ }in\ \href@noop {} {\emph {\bibinfo {booktitle} {Proceedings of the AAAI Conference on Artificial Intelligence}}},\ Vol.~\bibinfo {volume} {33}\ (\bibinfo {year} {2019})\ pp.\ \bibinfo {pages} {922--929}\BibitemShut {NoStop}%
\bibitem [{\citenamefont {Li}\ \emph {et~al.}(2022)\citenamefont {Li}, \citenamefont {Zhang}, \citenamefont {Li}, \citenamefont {Zhang},\ and\ \citenamefont {Zhang}}]{li2022dmgan}%
  \BibitemOpen
  \bibfield  {author} {\bibinfo {author} {\bibfnamefont {R.}~\bibnamefont {Li}}, \bibinfo {author} {\bibfnamefont {F.}~\bibnamefont {Zhang}}, \bibinfo {author} {\bibfnamefont {T.}~\bibnamefont {Li}}, \bibinfo {author} {\bibfnamefont {N.}~\bibnamefont {Zhang}}, \ and\ \bibinfo {author} {\bibfnamefont {T.}~\bibnamefont {Zhang}},\ }\bibfield  {title} {\enquote {\bibinfo {title} {Dmgan: Dynamic multi-hop graph attention network for traffic forecasting},}\ }\href@noop {} {\bibfield  {journal} {\bibinfo  {journal} {IEEE Transactions on Knowledge and Data Engineering}\ } (\bibinfo {year} {2022})}\BibitemShut {NoStop}%
\bibitem [{\citenamefont {Bruna}\ \emph {et~al.}(2014)\citenamefont {Bruna}, \citenamefont {Zaremba}, \citenamefont {Szlam},\ and\ \citenamefont {LeCun}}]{bruna2013spectral}%
  \BibitemOpen
  \bibfield  {author} {\bibinfo {author} {\bibfnamefont {J.}~\bibnamefont {Bruna}}, \bibinfo {author} {\bibfnamefont {W.}~\bibnamefont {Zaremba}}, \bibinfo {author} {\bibfnamefont {A.}~\bibnamefont {Szlam}}, \ and\ \bibinfo {author} {\bibfnamefont {Y.}~\bibnamefont {LeCun}},\ }\bibfield  {title} {\enquote {\bibinfo {title} {Spectral networks and deep locally connected networks on graphs},}\ }in\ \href@noop {} {\emph {\bibinfo {booktitle} {Proceedings of the 2nd International Conference on Learning Representations}}}\ (\bibinfo {year} {2014})\BibitemShut {NoStop}%
\bibitem [{\citenamefont {Defferrard}, \citenamefont {Bresson},\ and\ \citenamefont {Vandergheynst}(2016)}]{defferrard2016convolutional}%
  \BibitemOpen
  \bibfield  {author} {\bibinfo {author} {\bibfnamefont {M.}~\bibnamefont {Defferrard}}, \bibinfo {author} {\bibfnamefont {X.}~\bibnamefont {Bresson}}, \ and\ \bibinfo {author} {\bibfnamefont {P.}~\bibnamefont {Vandergheynst}},\ }\bibfield  {title} {\enquote {\bibinfo {title} {Convolutional neural networks on graphs with fast localized spectral filtering},}\ }in\ \href@noop {} {\emph {\bibinfo {booktitle} {Proceedings of the 30th International Conference on Neural Information Processing Systems}}},\ Vol.~\bibinfo {volume} {29}\ (\bibinfo {year} {2016})\ pp.\ \bibinfo {pages} {3844--3852}\BibitemShut {NoStop}%
\bibitem [{\citenamefont {Ju}\ \emph {et~al.}(2022)\citenamefont {Ju}, \citenamefont {Hou}, \citenamefont {Fan}, \citenamefont {Zhao}, \citenamefont {Ye},\ and\ \citenamefont {Zhao}}]{c:7}%
  \BibitemOpen
  \bibfield  {author} {\bibinfo {author} {\bibfnamefont {M.}~\bibnamefont {Ju}}, \bibinfo {author} {\bibfnamefont {S.}~\bibnamefont {Hou}}, \bibinfo {author} {\bibfnamefont {Y.}~\bibnamefont {Fan}}, \bibinfo {author} {\bibfnamefont {J.}~\bibnamefont {Zhao}}, \bibinfo {author} {\bibfnamefont {Y.}~\bibnamefont {Ye}}, \ and\ \bibinfo {author} {\bibfnamefont {L.}~\bibnamefont {Zhao}},\ }\bibfield  {title} {\enquote {\bibinfo {title} {Adaptive kernel graph neural network},}\ }in\ \href@noop {} {\emph {\bibinfo {booktitle} {Proceedings of the AAAI Conference on Artificial Intelligence}}},\ Vol.~\bibinfo {volume} {36}\ (\bibinfo {year} {2022})\ pp.\ \bibinfo {pages} {7051--7058}\BibitemShut {NoStop}%
\bibitem [{\citenamefont {Guo}\ \emph {et~al.}(2022)\citenamefont {Guo}, \citenamefont {Zhou}, \citenamefont {Hu}, \citenamefont {Li}, \citenamefont {Chang},\ and\ \citenamefont {Wang}}]{guo2022orthogonal}%
  \BibitemOpen
  \bibfield  {author} {\bibinfo {author} {\bibfnamefont {K.}~\bibnamefont {Guo}}, \bibinfo {author} {\bibfnamefont {K.}~\bibnamefont {Zhou}}, \bibinfo {author} {\bibfnamefont {X.}~\bibnamefont {Hu}}, \bibinfo {author} {\bibfnamefont {Y.}~\bibnamefont {Li}}, \bibinfo {author} {\bibfnamefont {Y.}~\bibnamefont {Chang}}, \ and\ \bibinfo {author} {\bibfnamefont {X.}~\bibnamefont {Wang}},\ }\bibfield  {title} {\enquote {\bibinfo {title} {Orthogonal graph neural networks},}\ }in\ \href@noop {} {\emph {\bibinfo {booktitle} {Proceedings of the AAAI Conference on Artificial Intelligence}}},\ Vol.~\bibinfo {volume} {36}\ (\bibinfo {year} {2022})\ pp.\ \bibinfo {pages} {3996--4004}\BibitemShut {NoStop}%
\bibitem [{\citenamefont {Jiang}\ \emph {et~al.}(2023{\natexlab{b}})\citenamefont {Jiang}, \citenamefont {Wang}, \citenamefont {Cheng}, \citenamefont {Tang},\ and\ \citenamefont {Luo}}]{jiang2021gpens}%
  \BibitemOpen
  \bibfield  {author} {\bibinfo {author} {\bibfnamefont {B.}~\bibnamefont {Jiang}}, \bibinfo {author} {\bibfnamefont {L.}~\bibnamefont {Wang}}, \bibinfo {author} {\bibfnamefont {J.}~\bibnamefont {Cheng}}, \bibinfo {author} {\bibfnamefont {J.}~\bibnamefont {Tang}}, \ and\ \bibinfo {author} {\bibfnamefont {B.}~\bibnamefont {Luo}},\ }\bibfield  {title} {\enquote {\bibinfo {title} {Gpens: Graph data learning with graph propagation-embedding networks},}\ }\href@noop {} {\bibfield  {journal} {\bibinfo  {journal} {IEEE Transactions on Neural Networks and Learning Systems}\ }\textbf {\bibinfo {volume} {34}},\ \bibinfo {pages} {3925--3938} (\bibinfo {year} {2023}{\natexlab{b}})}\BibitemShut {NoStop}%
\bibitem [{\citenamefont {Zhang}, \citenamefont {Wu},\ and\ \citenamefont {Yan}(2023)}]{zhang2021learning}%
  \BibitemOpen
  \bibfield  {author} {\bibinfo {author} {\bibfnamefont {T.}~\bibnamefont {Zhang}}, \bibinfo {author} {\bibfnamefont {Q.}~\bibnamefont {Wu}}, \ and\ \bibinfo {author} {\bibfnamefont {J.}~\bibnamefont {Yan}},\ }\bibfield  {title} {\enquote {\bibinfo {title} {Learning high-order graph convolutional networks via adaptive layerwise aggregation combination},}\ }\href@noop {} {\bibfield  {journal} {\bibinfo  {journal} {IEEE Transactions on Neural Networks and Learning Systems}\ }\textbf {\bibinfo {volume} {34}},\ \bibinfo {pages} {5144--5155} (\bibinfo {year} {2023})}\BibitemShut {NoStop}%
\bibitem [{\citenamefont {Bose}\ and\ \citenamefont {Das}(2023)}]{bose2023can}%
  \BibitemOpen
  \bibfield  {author} {\bibinfo {author} {\bibfnamefont {K.}~\bibnamefont {Bose}}\ and\ \bibinfo {author} {\bibfnamefont {S.}~\bibnamefont {Das}},\ }\bibfield  {title} {\enquote {\bibinfo {title} {Can graph neural networks go deeper without over-smoothing? yes, with a randomized path exploration!}}\ }\href@noop {} {\bibfield  {journal} {\bibinfo  {journal} {IEEE Transactions on Emerging Topics in Computational Intelligence}\ } (\bibinfo {year} {2023})}\BibitemShut {NoStop}%
\bibitem [{\citenamefont {Sawhney}\ \emph {et~al.}(2021)\citenamefont {Sawhney}, \citenamefont {Agarwal}, \citenamefont {Wadhwa},\ and\ \citenamefont {Shah}}]{sawhney2021exploring}%
  \BibitemOpen
  \bibfield  {author} {\bibinfo {author} {\bibfnamefont {R.}~\bibnamefont {Sawhney}}, \bibinfo {author} {\bibfnamefont {S.}~\bibnamefont {Agarwal}}, \bibinfo {author} {\bibfnamefont {A.}~\bibnamefont {Wadhwa}}, \ and\ \bibinfo {author} {\bibfnamefont {R.}~\bibnamefont {Shah}},\ }\bibfield  {title} {\enquote {\bibinfo {title} {Exploring the scale-free nature of stock markets: Hyperbolic graph learning for algorithmic trading},}\ }in\ \href@noop {} {\emph {\bibinfo {booktitle} {Proceedings of the Web Conference}}}\ (\bibinfo {year} {2021})\ pp.\ \bibinfo {pages} {11--22}\BibitemShut {NoStop}%
\bibitem [{\citenamefont {Guo}\ \emph {et~al.}(2024)\citenamefont {Guo}, \citenamefont {Sun}, \citenamefont {Lv}, \citenamefont {Ma}, \citenamefont {Niu}, \citenamefont {Gao},\ and\ \citenamefont {Wang}}]{graph}%
  \BibitemOpen
  \bibfield  {author} {\bibinfo {author} {\bibfnamefont {W.}~\bibnamefont {Guo}}, \bibinfo {author} {\bibfnamefont {X.}~\bibnamefont {Sun}}, \bibinfo {author} {\bibfnamefont {D.}~\bibnamefont {Lv}}, \bibinfo {author} {\bibfnamefont {W.}~\bibnamefont {Ma}}, \bibinfo {author} {\bibfnamefont {W.}~\bibnamefont {Niu}}, \bibinfo {author} {\bibfnamefont {Z.}~\bibnamefont {Gao}}, \ and\ \bibinfo {author} {\bibfnamefont {Y.}~\bibnamefont {Wang}},\ }\bibfield  {title} {\enquote {\bibinfo {title} {{Motion states identification of underwater glider based on complex networks and graph convolutional networks}},}\ }\href@noop {} {\bibfield  {journal} {\bibinfo  {journal} {Chaos: An Interdisciplinary Journal of Nonlinear Science}\ }\textbf {\bibinfo {volume} {34}},\ \bibinfo {pages} {023108} (\bibinfo {year} {2024})}\BibitemShut {NoStop}%
\bibitem [{\citenamefont {Ta}\ \emph {et~al.}(2022)\citenamefont {Ta}, \citenamefont {Liu}, \citenamefont {Hu}, \citenamefont {Yu}, \citenamefont {Sun},\ and\ \citenamefont {Du}}]{ta2022adaptive}%
  \BibitemOpen
  \bibfield  {author} {\bibinfo {author} {\bibfnamefont {X.}~\bibnamefont {Ta}}, \bibinfo {author} {\bibfnamefont {Z.}~\bibnamefont {Liu}}, \bibinfo {author} {\bibfnamefont {X.}~\bibnamefont {Hu}}, \bibinfo {author} {\bibfnamefont {L.}~\bibnamefont {Yu}}, \bibinfo {author} {\bibfnamefont {L.}~\bibnamefont {Sun}}, \ and\ \bibinfo {author} {\bibfnamefont {B.}~\bibnamefont {Du}},\ }\bibfield  {title} {\enquote {\bibinfo {title} {Adaptive spatio-temporal graph neural network for traffic forecasting},}\ }\href@noop {} {\bibfield  {journal} {\bibinfo  {journal} {Knowledge-Based Systems}\ }\textbf {\bibinfo {volume} {242}},\ \bibinfo {pages} {108199} (\bibinfo {year} {2022})}\BibitemShut {NoStop}%
\bibitem [{\citenamefont {Chen}, \citenamefont {Wu},\ and\ \citenamefont {Zaki}(2020)}]{chen2020iterative}%
  \BibitemOpen
  \bibfield  {author} {\bibinfo {author} {\bibfnamefont {Y.}~\bibnamefont {Chen}}, \bibinfo {author} {\bibfnamefont {L.}~\bibnamefont {Wu}}, \ and\ \bibinfo {author} {\bibfnamefont {M.}~\bibnamefont {Zaki}},\ }\bibfield  {title} {\enquote {\bibinfo {title} {Iterative deep graph learning for graph neural networks: Better and robust node embeddings},}\ }in\ \href@noop {} {\emph {\bibinfo {booktitle} {Advances in Neural Information Processing Systems}}},\ Vol.~\bibinfo {volume} {33}\ (\bibinfo {year} {2020})\ pp.\ \bibinfo {pages} {19314--19326}\BibitemShut {NoStop}%
\bibitem [{\citenamefont {Zhang}\ \emph {et~al.}(2023)\citenamefont {Zhang}, \citenamefont {Chen}, \citenamefont {Zhang}, \citenamefont {Qian},\ and\ \citenamefont {Wang}}]{zhang2023ctfnet}%
  \BibitemOpen
  \bibfield  {author} {\bibinfo {author} {\bibfnamefont {Z.}~\bibnamefont {Zhang}}, \bibinfo {author} {\bibfnamefont {Y.}~\bibnamefont {Chen}}, \bibinfo {author} {\bibfnamefont {D.}~\bibnamefont {Zhang}}, \bibinfo {author} {\bibfnamefont {Y.}~\bibnamefont {Qian}}, \ and\ \bibinfo {author} {\bibfnamefont {H.}~\bibnamefont {Wang}},\ }\bibfield  {title} {\enquote {\bibinfo {title} {Ctfnet: Long-sequence time-series forecasting based on convolution and time–frequency analysis},}\ }\href@noop {} {\bibfield  {journal} {\bibinfo  {journal} {IEEE Transactions on Neural Networks and Learning Systems}\ ,\ \bibinfo {pages} {1--15}} (\bibinfo {year} {2023})}\BibitemShut {NoStop}%
\bibitem [{\citenamefont {Feng}\ \emph {et~al.}(2024)\citenamefont {Feng}, \citenamefont {Gao}, \citenamefont {Xiao},\ and\ \citenamefont {Duan}}]{earlywarning}%
  \BibitemOpen
  \bibfield  {author} {\bibinfo {author} {\bibfnamefont {L.}~\bibnamefont {Feng}}, \bibinfo {author} {\bibfnamefont {T.}~\bibnamefont {Gao}}, \bibinfo {author} {\bibfnamefont {W.}~\bibnamefont {Xiao}}, \ and\ \bibinfo {author} {\bibfnamefont {J.}~\bibnamefont {Duan}},\ }\bibfield  {title} {\enquote {\bibinfo {title} {{Early warning indicators via latent stochastic dynamical systems}},}\ }\href@noop {} {\bibfield  {journal} {\bibinfo  {journal} {Chaos: An Interdisciplinary Journal of Nonlinear Science}\ }\textbf {\bibinfo {volume} {34}},\ \bibinfo {pages} {031101} (\bibinfo {year} {2024})}\BibitemShut {NoStop}%
\bibitem [{\citenamefont {Kothapalli}\ and\ \citenamefont {Totad}(2017)}]{kothapalli2017real}%
  \BibitemOpen
  \bibfield  {author} {\bibinfo {author} {\bibfnamefont {S.}~\bibnamefont {Kothapalli}}\ and\ \bibinfo {author} {\bibfnamefont {S.}~\bibnamefont {Totad}},\ }\bibfield  {title} {\enquote {\bibinfo {title} {A real-time weather forecasting and analysis},}\ }in\ \href@noop {} {\emph {\bibinfo {booktitle} {2017 IEEE International Conference on Power, Control, Signals and Instrumentation Engineering}}}\ (\bibinfo {year} {2017})\ pp.\ \bibinfo {pages} {1567--1570}\BibitemShut {NoStop}%
\bibitem [{\citenamefont {Shao}\ and\ \citenamefont {Soong}(2016)}]{shao2016traffic}%
  \BibitemOpen
  \bibfield  {author} {\bibinfo {author} {\bibfnamefont {H.}~\bibnamefont {Shao}}\ and\ \bibinfo {author} {\bibfnamefont {B.-H.}\ \bibnamefont {Soong}},\ }\bibfield  {title} {\enquote {\bibinfo {title} {Traffic flow prediction with long short-term memory networks (lstms)},}\ }in\ \href@noop {} {\emph {\bibinfo {booktitle} {IEEE Region 10 Conference}}}\ (\bibinfo {year} {2016})\ pp.\ \bibinfo {pages} {2986--2989}\BibitemShut {NoStop}%
\bibitem [{\citenamefont {Rebei}, \citenamefont {Amayri},\ and\ \citenamefont {Bouguila}(2023)}]{rebei2023fsnet}%
  \BibitemOpen
  \bibfield  {author} {\bibinfo {author} {\bibfnamefont {A.}~\bibnamefont {Rebei}}, \bibinfo {author} {\bibfnamefont {M.}~\bibnamefont {Amayri}}, \ and\ \bibinfo {author} {\bibfnamefont {N.}~\bibnamefont {Bouguila}},\ }\bibfield  {title} {\enquote {\bibinfo {title} {Fsnet: A hybrid model for seasonal forecasting},}\ }\href@noop {} {\bibfield  {journal} {\bibinfo  {journal} {IEEE Transactions on Emerging Topics in Computational Intelligence}\ } (\bibinfo {year} {2023})}\BibitemShut {NoStop}%
\bibitem [{\citenamefont {Jiang}\ \emph {et~al.}(2024)\citenamefont {Jiang}, \citenamefont {Yu}, \citenamefont {Anh}, \citenamefont {Lee},\ and\ \citenamefont {Zhou}}]{ensemble_air}%
  \BibitemOpen
  \bibfield  {author} {\bibinfo {author} {\bibfnamefont {S.}~\bibnamefont {Jiang}}, \bibinfo {author} {\bibfnamefont {Z.-G.}\ \bibnamefont {Yu}}, \bibinfo {author} {\bibfnamefont {V.~V.}\ \bibnamefont {Anh}}, \bibinfo {author} {\bibfnamefont {T.}~\bibnamefont {Lee}}, \ and\ \bibinfo {author} {\bibfnamefont {Y.}~\bibnamefont {Zhou}},\ }\bibfield  {title} {\enquote {\bibinfo {title} {{An ensemble multi-scale framework for long-term forecasting of air quality}},}\ }\href@noop {} {\bibfield  {journal} {\bibinfo  {journal} {Chaos: An Interdisciplinary Journal of Nonlinear Science}\ }\textbf {\bibinfo {volume} {34}},\ \bibinfo {pages} {013110} (\bibinfo {year} {2024})}\BibitemShut {NoStop}%
\bibitem [{\citenamefont {Wu}\ \emph {et~al.}(2022)\citenamefont {Wu}, \citenamefont {Hu}, \citenamefont {Liu}, \citenamefont {Zhou}, \citenamefont {Wang},\ and\ \citenamefont {Long}}]{wu2022timesnet}%
  \BibitemOpen
  \bibfield  {author} {\bibinfo {author} {\bibfnamefont {H.}~\bibnamefont {Wu}}, \bibinfo {author} {\bibfnamefont {T.}~\bibnamefont {Hu}}, \bibinfo {author} {\bibfnamefont {Y.}~\bibnamefont {Liu}}, \bibinfo {author} {\bibfnamefont {H.}~\bibnamefont {Zhou}}, \bibinfo {author} {\bibfnamefont {J.}~\bibnamefont {Wang}}, \ and\ \bibinfo {author} {\bibfnamefont {M.}~\bibnamefont {Long}},\ }\bibfield  {title} {\enquote {\bibinfo {title} {Timesnet: Temporal 2d-variation modeling for general time series analysis},}\ }in\ \href@noop {} {\emph {\bibinfo {booktitle} {Proceedings of the Eleventh International Conference on Learning Representations}}}\ (\bibinfo {year} {2022})\BibitemShut {NoStop}%
\bibitem [{\citenamefont {Zhang}\ and\ \citenamefont {Yan}(2022)}]{zhang2022crossformer}%
  \BibitemOpen
  \bibfield  {author} {\bibinfo {author} {\bibfnamefont {Y.}~\bibnamefont {Zhang}}\ and\ \bibinfo {author} {\bibfnamefont {J.}~\bibnamefont {Yan}},\ }\bibfield  {title} {\enquote {\bibinfo {title} {Crossformer: Transformer utilizing cross-dimension dependency for multivariate time series forecasting},}\ }in\ \href@noop {} {\emph {\bibinfo {booktitle} {Proceedings of the Eleventh International Conference on Learning Representations}}}\ (\bibinfo {year} {2022})\BibitemShut {NoStop}%
\bibitem [{\citenamefont {Cirstea}\ \emph {et~al.}(2022)\citenamefont {Cirstea}, \citenamefont {Guo}, \citenamefont {Yang}, \citenamefont {Kieu}, \citenamefont {Dong},\ and\ \citenamefont {Pan}}]{cirstea2022triformer}%
  \BibitemOpen
  \bibfield  {author} {\bibinfo {author} {\bibfnamefont {R.-G.}\ \bibnamefont {Cirstea}}, \bibinfo {author} {\bibfnamefont {C.}~\bibnamefont {Guo}}, \bibinfo {author} {\bibfnamefont {B.}~\bibnamefont {Yang}}, \bibinfo {author} {\bibfnamefont {T.}~\bibnamefont {Kieu}}, \bibinfo {author} {\bibfnamefont {X.}~\bibnamefont {Dong}}, \ and\ \bibinfo {author} {\bibfnamefont {S.}~\bibnamefont {Pan}},\ }\bibfield  {title} {\enquote {\bibinfo {title} {Triformer: Triangular, variable-specific attentions for long sequence multivariate time series forecasting--full version},}\ }\href@noop {} {\bibfield  {journal} {\bibinfo  {journal} {arXiv preprint arXiv:2204.13767}\ } (\bibinfo {year} {2022})}\BibitemShut {NoStop}%
\bibitem [{\citenamefont {Jiang}\ \emph {et~al.}(2023{\natexlab{c}})\citenamefont {Jiang}, \citenamefont {Han}, \citenamefont {Zhao},\ and\ \citenamefont {Wang}}]{jiang2023pdformer}%
  \BibitemOpen
  \bibfield  {author} {\bibinfo {author} {\bibfnamefont {J.}~\bibnamefont {Jiang}}, \bibinfo {author} {\bibfnamefont {C.}~\bibnamefont {Han}}, \bibinfo {author} {\bibfnamefont {W.~X.}\ \bibnamefont {Zhao}}, \ and\ \bibinfo {author} {\bibfnamefont {J.}~\bibnamefont {Wang}},\ }\bibfield  {title} {\enquote {\bibinfo {title} {Pdformer: Propagation delay-aware dynamic long-range transformer for traffic flow prediction},}\ }in\ \href@noop {} {\emph {\bibinfo {booktitle} {Proceedings of the AAAI Conference on Artificial Intelligence}}},\ Vol.~\bibinfo {volume} {37}\ (\bibinfo {year} {2023})\ pp.\ \bibinfo {pages} {4365--4373}\BibitemShut {NoStop}%
\bibitem [{\citenamefont {Zou}\ \emph {et~al.}(2024)\citenamefont {Zou}, \citenamefont {Zhang}, \citenamefont {Wang},\ and\ \citenamefont {Hu}}]{power2}%
  \BibitemOpen
  \bibfield  {author} {\bibinfo {author} {\bibfnamefont {Y.}~\bibnamefont {Zou}}, \bibinfo {author} {\bibfnamefont {H.}~\bibnamefont {Zhang}}, \bibinfo {author} {\bibfnamefont {H.}~\bibnamefont {Wang}}, \ and\ \bibinfo {author} {\bibfnamefont {J.}~\bibnamefont {Hu}},\ }\bibfield  {title} {\enquote {\bibinfo {title} {{Predicting Braess's paradox of power grids using graph neural networks}},}\ }\href@noop {} {\bibfield  {journal} {\bibinfo  {journal} {Chaos: An Interdisciplinary Journal of Nonlinear Science}\ }\textbf {\bibinfo {volume} {34}},\ \bibinfo {pages} {013127} (\bibinfo {year} {2024})}\BibitemShut {NoStop}%
\bibitem [{\citenamefont {Nauck}\ \emph {et~al.}(2023)\citenamefont {Nauck}, \citenamefont {Lindner}, \citenamefont {Schürholt},\ and\ \citenamefont {Hellmann}}]{power_graph}%
  \BibitemOpen
  \bibfield  {author} {\bibinfo {author} {\bibfnamefont {C.}~\bibnamefont {Nauck}}, \bibinfo {author} {\bibfnamefont {M.}~\bibnamefont {Lindner}}, \bibinfo {author} {\bibfnamefont {K.}~\bibnamefont {Schürholt}}, \ and\ \bibinfo {author} {\bibfnamefont {F.}~\bibnamefont {Hellmann}},\ }\bibfield  {title} {\enquote {\bibinfo {title} {{Toward dynamic stability assessment of power grid topologies using graph neural networks}},}\ }\href@noop {} {\bibfield  {journal} {\bibinfo  {journal} {Chaos: An Interdisciplinary Journal of Nonlinear Science}\ }\textbf {\bibinfo {volume} {33}},\ \bibinfo {pages} {103103} (\bibinfo {year} {2023})}\BibitemShut {NoStop}%
\bibitem [{\citenamefont {Zhao}\ \emph {et~al.}(2020)\citenamefont {Zhao}, \citenamefont {Song}, \citenamefont {Zhang}, \citenamefont {Liu}, \citenamefont {Wang}, \citenamefont {Lin}, \citenamefont {Deng},\ and\ \citenamefont {Li}}]{zhao2019t}%
  \BibitemOpen
  \bibfield  {author} {\bibinfo {author} {\bibfnamefont {L.}~\bibnamefont {Zhao}}, \bibinfo {author} {\bibfnamefont {Y.}~\bibnamefont {Song}}, \bibinfo {author} {\bibfnamefont {C.}~\bibnamefont {Zhang}}, \bibinfo {author} {\bibfnamefont {Y.}~\bibnamefont {Liu}}, \bibinfo {author} {\bibfnamefont {P.}~\bibnamefont {Wang}}, \bibinfo {author} {\bibfnamefont {T.}~\bibnamefont {Lin}}, \bibinfo {author} {\bibfnamefont {M.}~\bibnamefont {Deng}}, \ and\ \bibinfo {author} {\bibfnamefont {H.}~\bibnamefont {Li}},\ }\bibfield  {title} {\enquote {\bibinfo {title} {T-gcn: A temporal graph convolutional network for traffic prediction},}\ }\href@noop {} {\bibfield  {journal} {\bibinfo  {journal} {IEEE Transactions on Intelligent Transportation Systems}\ }\textbf {\bibinfo {volume} {21}},\ \bibinfo {pages} {3848--3858} (\bibinfo {year} {2020})}\BibitemShut {NoStop}%
\bibitem [{\citenamefont {Li}\ \emph {et~al.}(2018)\citenamefont {Li}, \citenamefont {Yu}, \citenamefont {Shahabi},\ and\ \citenamefont {Liu}}]{li2018diffusion}%
  \BibitemOpen
  \bibfield  {author} {\bibinfo {author} {\bibfnamefont {Y.}~\bibnamefont {Li}}, \bibinfo {author} {\bibfnamefont {R.}~\bibnamefont {Yu}}, \bibinfo {author} {\bibfnamefont {C.}~\bibnamefont {Shahabi}}, \ and\ \bibinfo {author} {\bibfnamefont {Y.}~\bibnamefont {Liu}},\ }\bibfield  {title} {\enquote {\bibinfo {title} {Diffusion convolutional recurrent neural network: Data-driven traffic forecasting},}\ }in\ \href@noop {} {\emph {\bibinfo {booktitle} {Proceedings of the 6th International Conference on Learning Representations}}}\ (\bibinfo {year} {2018})\BibitemShut {NoStop}%
\bibitem [{\citenamefont {Lin}\ \emph {et~al.}(2022)\citenamefont {Lin}, \citenamefont {Gao}, \citenamefont {Xu}, \citenamefont {Wu}, \citenamefont {Li},\ and\ \citenamefont {Li}}]{c:6}%
  \BibitemOpen
  \bibfield  {author} {\bibinfo {author} {\bibfnamefont {H.}~\bibnamefont {Lin}}, \bibinfo {author} {\bibfnamefont {Z.}~\bibnamefont {Gao}}, \bibinfo {author} {\bibfnamefont {Y.}~\bibnamefont {Xu}}, \bibinfo {author} {\bibfnamefont {L.}~\bibnamefont {Wu}}, \bibinfo {author} {\bibfnamefont {L.}~\bibnamefont {Li}}, \ and\ \bibinfo {author} {\bibfnamefont {S.~Z.}\ \bibnamefont {Li}},\ }\bibfield  {title} {\enquote {\bibinfo {title} {Conditional local convolution for spatio-temporal meteorological forecasting},}\ }in\ \href@noop {} {\emph {\bibinfo {booktitle} {Proceedings of the AAAI Conference on Artificial Intelligence}}},\ Vol.~\bibinfo {volume} {36}\ (\bibinfo {year} {2022})\ pp.\ \bibinfo {pages} {7470--7478}\BibitemShut {NoStop}%
\bibitem [{\citenamefont {Seo}\ \emph {et~al.}(2018)\citenamefont {Seo}, \citenamefont {Defferrard}, \citenamefont {Vandergheynst},\ and\ \citenamefont {Bresson}}]{seo2018structured}%
  \BibitemOpen
  \bibfield  {author} {\bibinfo {author} {\bibfnamefont {Y.}~\bibnamefont {Seo}}, \bibinfo {author} {\bibfnamefont {M.}~\bibnamefont {Defferrard}}, \bibinfo {author} {\bibfnamefont {P.}~\bibnamefont {Vandergheynst}}, \ and\ \bibinfo {author} {\bibfnamefont {X.}~\bibnamefont {Bresson}},\ }\bibfield  {title} {\enquote {\bibinfo {title} {Structured sequence modeling with graph convolutional recurrent networks},}\ }in\ \href@noop {} {\emph {\bibinfo {booktitle} {Neural Information Processing: 25th International Conference}}}\ (\bibinfo {year} {2018})\ pp.\ \bibinfo {pages} {362--373}\BibitemShut {NoStop}%
\bibitem [{\citenamefont {Cini}\ \emph {et~al.}(2023)\citenamefont {Cini}, \citenamefont {Marisca}, \citenamefont {Bianchi},\ and\ \citenamefont {Alippi}}]{cini2023scalable}%
  \BibitemOpen
  \bibfield  {author} {\bibinfo {author} {\bibfnamefont {A.}~\bibnamefont {Cini}}, \bibinfo {author} {\bibfnamefont {I.}~\bibnamefont {Marisca}}, \bibinfo {author} {\bibfnamefont {F.~M.}\ \bibnamefont {Bianchi}}, \ and\ \bibinfo {author} {\bibfnamefont {C.}~\bibnamefont {Alippi}},\ }\bibfield  {title} {\enquote {\bibinfo {title} {Scalable spatiotemporal graph neural networks},}\ }in\ \href@noop {} {\emph {\bibinfo {booktitle} {Proceedings of the AAAI conference on Artificial Intelligence}}},\ Vol.~\bibinfo {volume} {37}\ (\bibinfo {year} {2023})\ pp.\ \bibinfo {pages} {7218--7226}\BibitemShut {NoStop}%
\bibitem [{\citenamefont {Yu}, \citenamefont {Yin},\ and\ \citenamefont {Zhu}(2018)}]{yu2017spatio}%
  \BibitemOpen
  \bibfield  {author} {\bibinfo {author} {\bibfnamefont {B.}~\bibnamefont {Yu}}, \bibinfo {author} {\bibfnamefont {H.}~\bibnamefont {Yin}}, \ and\ \bibinfo {author} {\bibfnamefont {Z.}~\bibnamefont {Zhu}},\ }\bibfield  {title} {\enquote {\bibinfo {title} {Spatio-temporal graph convolutional networks: a deep learning framework for traffic forecasting},}\ }in\ \href@noop {} {\emph {\bibinfo {booktitle} {Proceedings of the 27th International Joint Conference on Artificial Intelligence}}}\ (\bibinfo {year} {2018})\ pp.\ \bibinfo {pages} {3634--3640}\BibitemShut {NoStop}%
\bibitem [{\citenamefont {Rasp}\ \emph {et~al.}(2020)\citenamefont {Rasp}, \citenamefont {Dueben}, \citenamefont {Scher}, \citenamefont {Weyn}, \citenamefont {Mouatadid},\ and\ \citenamefont {Thuerey}}]{rasp2020weatherbench}%
  \BibitemOpen
  \bibfield  {author} {\bibinfo {author} {\bibfnamefont {S.}~\bibnamefont {Rasp}}, \bibinfo {author} {\bibfnamefont {P.~D.}\ \bibnamefont {Dueben}}, \bibinfo {author} {\bibfnamefont {S.}~\bibnamefont {Scher}}, \bibinfo {author} {\bibfnamefont {J.~A.}\ \bibnamefont {Weyn}}, \bibinfo {author} {\bibfnamefont {S.}~\bibnamefont {Mouatadid}}, \ and\ \bibinfo {author} {\bibfnamefont {N.}~\bibnamefont {Thuerey}},\ }\bibfield  {title} {\enquote {\bibinfo {title} {Weatherbench: a benchmark data set for data-driven weather forecasting},}\ }\href@noop {} {\bibfield  {journal} {\bibinfo  {journal} {Journal of Advances in Modeling Earth Systems}\ }\textbf {\bibinfo {volume} {12}},\ \bibinfo {pages} {e2020MS002203} (\bibinfo {year} {2020})}\BibitemShut {NoStop}%
\bibitem [{\citenamefont {Oreshkin}\ \emph {et~al.}(2019)\citenamefont {Oreshkin}, \citenamefont {Carpov}, \citenamefont {Chapados},\ and\ \citenamefont {Bengio}}]{oreshkin2019n}%
  \BibitemOpen
  \bibfield  {author} {\bibinfo {author} {\bibfnamefont {B.~N.}\ \bibnamefont {Oreshkin}}, \bibinfo {author} {\bibfnamefont {D.}~\bibnamefont {Carpov}}, \bibinfo {author} {\bibfnamefont {N.}~\bibnamefont {Chapados}}, \ and\ \bibinfo {author} {\bibfnamefont {Y.}~\bibnamefont {Bengio}},\ }\bibfield  {title} {\enquote {\bibinfo {title} {N-beats: Neural basis expansion analysis for interpretable time series forecasting},}\ }in\ \href@noop {} {\emph {\bibinfo {booktitle} {Proceedings of the 7th International Conference on Learning Representations}}}\ (\bibinfo {year} {2019})\BibitemShut {NoStop}%
\bibitem [{\citenamefont {Wu}\ \emph {et~al.}(2021)\citenamefont {Wu}, \citenamefont {Xu}, \citenamefont {Wang},\ and\ \citenamefont {Long}}]{wu2021autoformer}%
  \BibitemOpen
  \bibfield  {author} {\bibinfo {author} {\bibfnamefont {H.}~\bibnamefont {Wu}}, \bibinfo {author} {\bibfnamefont {J.}~\bibnamefont {Xu}}, \bibinfo {author} {\bibfnamefont {J.}~\bibnamefont {Wang}}, \ and\ \bibinfo {author} {\bibfnamefont {M.}~\bibnamefont {Long}},\ }\bibfield  {title} {\enquote {\bibinfo {title} {Autoformer: Decomposition transformers with auto-correlation for long-term series forecasting},}\ }in\ \href@noop {} {\emph {\bibinfo {booktitle} {Advances in Neural Information Processing Systems}}},\ Vol.~\bibinfo {volume} {34}\ (\bibinfo {year} {2021})\ pp.\ \bibinfo {pages} {22419--22430}\BibitemShut {NoStop}%
\end{thebibliography}%

\end{document}